\documentclass[twoside, 12pt, a4-paper]{article}

\usepackage{PRIMEarxiv}
\usepackage{hyperref}       
\usepackage{url}            
\usepackage{booktabs}       
\usepackage{amsfonts}       
\usepackage{nicefrac}       
\usepackage{microtype}      
\usepackage{lipsum}
\usepackage{fancyhdr}       
\usepackage{graphicx}       
\graphicspath{{media/}}     
\usepackage{multicol,lipsum}
\usepackage{enumerate}
\usepackage[table, dvipsnames]{xcolor}
\renewcommand{\headrulewidth}{.4pt}

\usepackage{graphicx}
\usepackage{afterpage}
\usepackage[utf8]{inputenc}
\usepackage{graphicx}
\usepackage{amssymb}
\usepackage{amsmath}
\usepackage{cleveref}
\usepackage{bm}
\usepackage{hyperref}
\usepackage{float, multirow}
\usepackage{xcolor}
\usepackage{setspace}
\usepackage{mathptmx} 
\usepackage{newtxtext,newtxmath}
\usepackage{lscape}
\usepackage{soul} 
\usepackage[title]{appendix}
\usepackage{setspace}
\usepackage{hanging}
\usepackage{xspace}
\usepackage{tikz}
\usetikzlibrary{positioning}
\usetikzlibrary{shapes.geometric, arrows}
\usepackage{calc}
\usetikzlibrary{arrows.meta}
\usepackage{varwidth}
\usepackage[numbers, sort&compress, square]{natbib}
\usepackage{hyperref}
\hypersetup{colorlinks,linkcolor={blue},citecolor={blue},urlcolor={blue}}
\usepackage{tabulary}
\newcolumntype{L}[1]{>{\raggedright\let\newline\\\arraybackslash\hspace{0pt}}m{#1}}
\newcolumntype{C}[1]{>{\centering\let\newline\\\arraybackslash\hspace{0pt}}m{#1}}
\newcolumntype{R}[1]{>{\raggedleft\let\newline\\\arraybackslash\hspace{0pt}}m{#1}}
\usepackage[skip=2pt, font=small]{subcaption}
\makeatletter
\renewcommand{\fnum@figure}{Fig. \thefigure}									 
\makeatother
\usepackage[skip=2pt,font=small]{caption}
 
\usepackage{enumitem}

\newcommand\nox{\texorpdfstring{NO\textsubscript{x}}{NOx}\xspace}

\usepackage{fancyhdr}
\usepackage{lineno}
\usepackage{stackengine}

\newlist{romanitem}{enumerate}{1}
\setlist[romanitem,1]{label=\textbf{Stage \Roman*}:, align=left, leftmargin=*}

\begin{document}
 \title{A Digital Twin for Diesel Engines: Operator-infused Physics-Informed Neural Networks with Transfer Learning for Engine Health Monitoring}
\author{Kamaljyoti Nath \\ Division of Applied Mathematics\\  Brown University \\ United States of America \\ kamaljyoti\_nath@brown.edu \And Varun Kumar \\ School of Engineering \\ Brown University \\ United States of America \\ varun\_kumar2@brown.edu \And Daniel J. Smith \\ Cummins Inc. \\ United States of America \\ daniel.j.smith@cummins.com \And George Em Karniadakis \\ Division of Applied Mathematics \\ Brown University \\ United States of America \\ george\_karniadakis@brown.edu \thanks{\textit{\underline{Corresponding author:}} G. E. Karniadakis (george\_karniadakis@brown.edu)}}
\maketitle
\fancypagestyle{alim}{\fancyhf{}\renewcommand{\headrulewidth}{0pt}\fancyfoot[R]{\today}\fancyfoot[L]{Preprint}}
\thispagestyle{alim}
\vspace{-1cm}
\maketitle

\begin{abstract}
Improving diesel engine efficiency, reducing emissions, and enabling robust health monitoring have been critical research topics in engine modelling. While recent advancements in the use of neural networks for system monitoring have shown promising results, such methods often focus on component-level analysis, lack generalizability, and physical interpretability. In this study, we propose a novel hybrid framework that combines physics-informed neural networks (PINNs) with deep operator networks (DeepONet) to enable accurate and computationally efficient parameter identification in mean-value diesel engine models. Our method leverages physics-based system knowledge in combination with data-driven training of neural networks to enhance model applicability. Incorporating offline-trained DeepONets to predict actuator dynamics significantly lowers the online computation cost when compared to the existing PINN framework. To address the re-training burden typical of PINNs under varying input conditions, we propose two transfer learning (TL) strategies: (i) a multi-stage TL scheme offering better runtime efficiency than full online training of the PINN model and (ii) a few-shot TL scheme that freezes a shared multi-head network body and computes physics-based derivatives required for model training outside the training loop. The second strategy offers a computationally inexpensive and physics-based approach for predicting engine dynamics and parameter identification, offering computational efficiency over the existing PINN framework. Compared to existing health monitoring methods, our framework combines the interpretability of physics-based models with the flexibility of deep learning, offering substantial gains in generalization, accuracy, and deployment efficiency for diesel engine diagnostics.
\end{abstract}


\section{Introduction}
\label{Section:Introduction}
Model-based health monitoring of internal combustion engines has seen significant progress in recent decades \cite{health_monitoring_review, health_monitoring_survey_gao2015, health_monitoring_acoustic, health_monitoring_wavelets, health_monitoring_wear_rate}, motivated by the need to enhance engine performance, reliability, and maintenance efficacy. Engine manufacturers have developed several diagnostic and service tools for operational needs. However, conventional monitoring strategies face persistent challenges, including poor fault isolation, limited prognostic capabilities, a lack of robustness under varying operating conditions, and gradual degradation in diagnostic performance over time. For example, system-level monitors can falsely flag faults due to marginally degraded components acting in combination, leading to expensive trial-and-error approaches for fault identification. To address these limitations, manufacturers have explored component-level monitoring with tight thresholds. This introduces a high rate of false positives and requires the development of high-fidelity models for individual components, resulting in a significant increase in monetary investment and cycle time. Consequently, various alternative solutions have been explored, including the use of complex logic algorithms, multiple degradation thresholds, and out-of-mission system tests. These approaches have led to increasingly sophisticated and expensive diagnostic infrastructures, burdening system design, tuning, and validation processes.

Recent advances in sensing and control technologies have enabled real-time collection of high-fidelity data from the engine's sub-systems. This has accelerated research interest in the development of tools and methods for engine modelling and diagnostics. Traditional methods such as Extended Kalman Filter (EKF) \cite{EKF_based_sensor} have been proposed for fault signal correction and diagnosis in the engine's air intake system. Although widely used, such methods are limited in their ability to capture strong non-linear, transient behaviour observed under real-world operating conditions. To overcome such challenges, numerous studies have explored the use of neural networks for dynamic modelling, fault detection, and virtual sensing in diesel engines. For instance, \citeauthor{Biao_2009} \cite{Biao_2009} utilized an auto-regressive neural network framework using engine load, fuel injection, and feedback rotational speed to predict rotational speed and power generation of a sixteen-cylinder locomotive engine with good accuracy. \citeauthor{Tosun_2016} \cite{Tosun_2016} compared linear regression and neural network for diesel engines fuelled with biodiesel-alcohol mixture to predict engine torque, CO and \nox, illustrating superior results with neural networks. \citeauthor{thompson2000neural} \cite{thompson2000neural} proposed a neural network framework for heavy-duty diesel engine emission and performance prediction using a data-driven approach. They used a partial recurrent network architecture utilizing instantaneous engine parameter values and 5-10 sec history window of same data as inputs for predicting engine emission and torque prediction within 5\% error on field data. Similarly, \citeauthor{wang2011_model} \cite{wang2011_model} proposed a neural network method for a virtual \nox sensor aimed at replacing the physical sensors. Their approach utilizes a physical model for generating signals such as EGR flow rate, coolant temperature and others that are then used as an input to a neural network model that predicts \nox levels at transient operating conditions. \citeauthor{shamekhi2015_dieselmod} \cite{shamekhi2015_dieselmod} proposed Neuro-Mean Value Model that improved the mean value diesel engine model by embedding small, pre-trained neural networks in engine sub-systems.

Building on the advances in applying neural networks for modelling dynamic engine states, efforts have also been made to utilize deep learning methods for fault detection. In an early work, \citeauthor{neuralFDI_1993} \cite{neuralFDI_1993} proposed a single-layer neural network for predicting six output conditions representing the engine's state of health. This framework was trained using simulated data for healthy and faulty engines, where the fault was simulated with a leak in the exhaust manifold to analyze turbocharger performance degradation. In more recent works, \citeauthor{CERVANTES_multi_fault} \cite{CERVANTES_multi_fault} developed a multi-neural network scheme that uses five parallel networks that simultaneously estimate the response from three critical sensors and generate a residual signal by comparing with individual sensor signals during engine operation. The disparity between the physical sensor and neural network estimation was used to trigger a fault signal when a certain threshold is reached. \citeauthor{active_fault_tolerance} \cite{active_fault_tolerance} embedded neural observers within the air-fuel ratio control system for analytical redundancy. In the event of a faulty sensor detected by the engine control unit, the analytical signal generated by the neural observer replaces the physical sensor signal to ensure adequate air-fuel ratio is maintained. \citeauthor{cnn_fault_diagnosis_Yang} \cite{cnn_fault_diagnosis_Yang} proposed a fault detection framework using 1D CNN layers to classify three different fault conditions based on the engine's vibration signature. The fault conditions were simulated on an engine test bed and the vibration signal was measured using dedicated accelerometers. \citeauthor{autoencoder_fault} \cite{autoencoder_fault} utilized an auto-encoder setup to identify faulty signals based on the magnitude of reconstruction error between healthy and faulty input signals.

The idea of developing Digital Twins (DT) coupled with deep learning for modelling complex engine systems has also gained popularity. \citeauthor{digital_twin_marine_engine} \cite{digital_twin_marine_engine} proposed a comprehensive DT framework combining a thermodynamic model with operational engine data for component health assessment. This framework successfully identified underperforming cylinders through deviations in key indicators such as Brake Specific Fuel Consumption (BSFC), Indicated Mean Effective Pressure (IMEP), and Exhaust Gas Temperature (EGT). However, their approach relies heavily on first-principles modelling and calibration using controlled test data. \citeauthor{digital_twin_vibration} \cite{digital_twin_vibration} proposed a DT-based auxilliary model with adaptive sparse attention networks to identify valve clearance faults by dynamically filtering redundant features and identifying salient segments of engine vibration data. \citeauthor{digital_twin_fault_mitigation} \cite{digital_twin_fault_mitigation} extended the DT paradigm by using it for fault identification and active control strategy in their hybrid engine model that incorporates neural networks with conventional engine models. This framework demonstrated the ability to identify faults in injector and manifold leakage, and successfully modify engine parameters to help recover engine performance.

Despite these advances in engine modelling and fault diagnosis, some key limitations persist in literature. The majority of the approaches are purely data-driven which limits interpretability and physical consistency when extrapolating across operating regimes other than what exists in the training set. Additionally, the accuracy of such data-driven approaches rely on the quality and quantity of data that can be collected. To address this, hybrid modelling approaches have emerged with the objective of combining the capability of neural networks with physics-based models. For example, \citeauthor{pulpeiro2022_dieselmod} \cite{pulpeiro2022_dieselmod} proposed a hybrid method combining physics and data to model gas exchange process within diesel engine sub-systems. Although promising, their work is focused on gas flow dynamics of individual components and does not address the dynamics at the system level. In a recent work, \citeauthor{hybrid_CNN_fault_diagnosis_Singh} \cite{hybrid_CNN_fault_diagnosis_Singh} utilize a combination of neural network surrogates and physics-based mean value engine model to predict the fault category of diesel engine based. Specifically, they use an auto-encoder to compress an 89-dimensional input signal collected from an engine test bed, combine it with six simulated signals generated by their physics-based model, and use it as an input to a CNN network that classifies the type of engine fault. This method, however depends on identifying deviations between healthy and faulty signals over a period of time which can lead to false prognosis.

A review of existing works, including those using neural networks, indicates a focus on state detection (eg. fault/no-fault classification of known faults) rather than parameter estimation to identify and quantify deviant behaviour. Additionally, typical neural-network based approaches rely purely on empirical data and often generalize poorly outside the training regime. Also, we note that most existing methods focus on component-level modelling and state estimation, instead of complete engine system due to significantly greater complexity associated with such system-wide integration.  In our view, monitoring parameter drifts allows estimation of underlying system-parameters based on known physical relationships between components, thus offering a more precise root cause analysis. Also, incorporating parameter identification in health monitoring system helps improve generalization and reduce ambiguity, since these methods are grounded in physics and are more likely to disentangle overlapping effects. In light of these research gaps, this study aims to streamline the development of health monitoring systems, improve fault isolation, system health prognostics, and predictive maintenance capabilities. Our focus is on advancing system-level monitoring solutions that utilize all available data to provide high-fidelity monitoring of system performance while also delivering robust fault isolation and prognostic capabilities.

\par As a step towards parameter-based health monitoring, we propose an integrated physics and data-based approach to identify unknown parameters of engine sub-systems by using a framework based on physics-informed neural networks (PINNs). In particular, PINNs \cite{Raissi_2019} leverage the known physical laws, typically represented by differential equations, to train neural network parameters for predicting solutions to these equations or predicting unknown parameters of these equations from known data.  While PINNs have demonstrated success across a variety of applications \cite{Chen_2020_neno_optics, Depina_2022_unsaturated, Majid_2022, Vahab_2023, Daneker_2023, Castro_2023, mariappan_2023}, their potential for fast prediction in physical systems remains an open research challenge. In a previous study by \citeauthor{Nath_2023} \cite{Nath_2023}, it was demonstrated how PINNs can be used for estimating unknown parameters and gas flow dynamics in a diesel engine model described in \cite{Wahlstrom}. It was shown that the proposed method could predict simultaneously both unknown parameters and dynamics accurately. However, a limitation of this work was the high computational cost for parameter estimation, primarily stemming from the need to re-train the PINN model every time inference was required on a new data signal. This limited the applicability of this framework in field systems that require rapid prediction capability. Thus, our research premise is as follows: given a set of measurements and a physics-based model, estimate the critical parameters for sub-systems that can help in engine health monitoring. In this study, first, we propose a hybrid method combining PINNs with neural operator learning for accelerating parameter estimation. While PINNs are employed to estimate the state variable and unknown parameters, independent states are approximated using a neural operator. Further, we propose two transfer learning (TL) strategies to reduce the computational cost for application in the real-world.

\par We consider the Deep Operator Network (DeepONet) \cite{Lu_2021_deepOnet} as our choice of the neural operator and build on the work of \cite{Kumar_2023}, who used DeepONet for predicting the gas flow dynamics described by \cite{Wahlstrom}. This DeepONet model proposed by \cite{Kumar_2023} is not designed for estimating the parameters of critical sub-systems but provides rapid estimates for gas flow dynamics. Hence, our objective is to combine the fast prediction capability of the operator network with the hybrid PINN framework for accelerating parameter estimation on diesel engines. In order to further reduce the computational cost, we present two transfer learning (TL) approaches: a) multi-stage transfer learning and b) a few-shot transfer learning using a larger multi-head pre-trained neural network. In a  previous study \cite{Nath_2023}, pre-trained neural networks were used for approximating empirical relationships, and the results showed good matches with the ground truths. One of the major advantages of using a pre-trained neural network over least square-based regression is the fact that neural networks can inherently learn the non-linear mapping between inputs and outputs without assuming the type of relationship. Tosun et al. \cite{Tosun_2016} also observed that a neural network approximation is better than the least squares method. However, one of the major concerns in both neural network and least squares method approximation is how to incorporate the engine's systematic (e.g., manufacturing) variability in empirical formulae. In other words, the engine unit used to obtain data for training the pre-trained network may vary marginally from the actual engines used in the field. In this study, we consider dropout in the pre-trained networks to account for such systematic variability.

\par In summary, we propose a hybrid deep operator-infused Physics-Informed Neural Network (PINN) method to reduce the computation time required for estimating unknown parameters on a diesel engine. To further reduce the computation cost, we present two TL approaches, a multi-stage TL and a few-shot TL method. The applicability and accuracy of the proposed methods are shown by considering specific examples for each approach. For this work, we generate training data utilizing the Simulink file \cite{Simulink_file} provided in \cite{Wahlstrom}. The input signals for generating our simulated data are obtained from field measurements on engine test beds. Additionally, we incorporate appropriate Gaussian noise in the simulated data for the inverse problem to mimic real-world field data and the expected aleatoric uncertainty. Detailed problem statements and different cases considered are discussed in section \ref{Subsection:Problem setup} along with a brief discussion on the engine model considered in this study.

\subsection{Key research gaps and our contributions}
We summarize below the key research gaps we identified and our contributions to address these gaps.
\begin{itemize}[leftmargin=*]
    \item \textbf{Research gap}: There is a need to integrate physics-based models with existing data-driven neural network methods for health monitoring in diesel engines. Current neural network methods rely on residual-signal monitoring, making them ambiguous and less objective.\\
    \emph{Contribution:} We propose an operator-infused PINN framework for rapid estimation of unknown system parameters and dynamics in diesel engines. The operator networks approximate the independent state variables using data-driven training while PINN is used to model the gas filling dynamics and estimate the unknown parameters using known physics from the mean value model. This approach to parameter identification through inverse analysis reduces ambiguity in fault detection, thereby enabling faster decision making.
    
    \item \textbf{Research gap:} Current methods rely on state estimates of sub-systems for health monitoring and fault detection but do not address the engine system holistically.\\
    \emph{Contribution:} We propose a system-of-systems approach for modelling diesel engines through neural networks by substituting each sub-system of a diesel engine model with data and physics-based neural networks. Each network works in tandem to replicate a complete mean value model of the diesel engine.
    
    \item  \textbf{Research gap:} Real-world application requires methods that can estimate engine dynamics efficiently across different operating conditions. Physics-based methods such as \cite{Nath_2023} can be extended across different operating scenarios but fail to meet efficiency requirements for field applications.\\
    \emph{Contribution:} We develop an efficient parameter estimation method suited for deployment on field systems, which is grounded in physics based knowledge. The robustness of our method is tested by replicating real-world noisy conditions in simulated data. We propose two transfer learning strategies to meet the computational requirements of field systems. These frameworks utilize transfer learning combined with pre-trained operator networks to meet computational and robustness requirements for field systems. Additionally, we employ a multi-head training approach utilizing a large corpus of prior training data. In this method, only a new output layer is trained for a new scenario and the derivatives for the physics loss is calculated only once outside the training loop, further reducing computational complexity. The multi-head model bears close association to a Foundation model \cite{Poseidon} since it can be reused for different scenarios or engine configurations.

\end{itemize}

\subsection{Problem setup}
\label{Subsection:Problem setup}
\begin{figure}[H]
	\centering
	\includegraphics[width=0.5\textwidth]{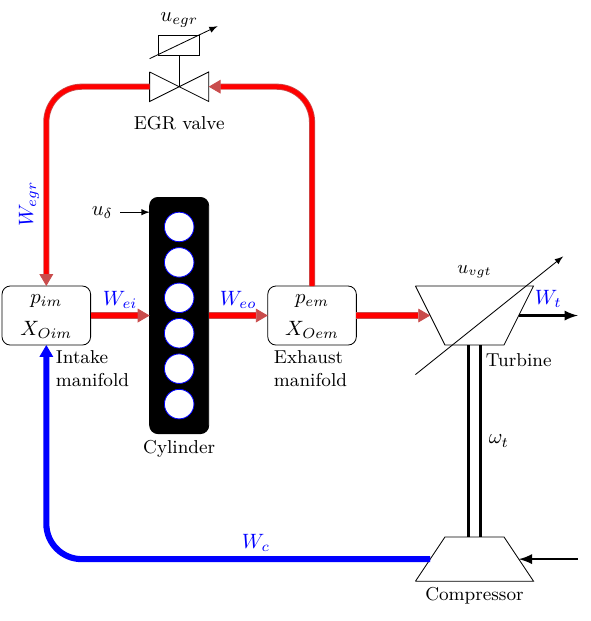}
	\caption{\textbf{A schematic diagram of the mean value diesel engine} with a variable-geometry turbocharger and exhaust gas recirculation \cite{Wahlstrom}. The main components of the engine are the intake manifold, the exhaust manifold, the cylinder, the EGR valve system, the compressor, and the turbine. The control input vector is $\bm{u} = \{u_\delta, u_{egr}, u_{vgt}\}$, and the engine speed is $n_e$. The gas flow rates through the different subsystems are denoted by $W_{i}$'s (in blue) and represent the gas flow dynamics. The blue flow path represents clean airflow while the red path represents air-combustion mixture. (Source: Figure adapted from \cite{Wahlstrom})}.
	\label{Figure:Schematic diagram of Engine}
\end{figure}

A mean value model for the gas flow dynamics of a diesel engine was proposed by \cite{Wahlstrom}, and we adopt this model in our present study. The main components of the engine considered in this model are the intake manifold, the exhaust manifold, the cylinder, the EGR valve system, the compressor, and the turbine. The gas flow dynamics through various components of the engine is described by the state variables of its component and the dependent variables. Specifically, the model we consider \cite{Wahlstrom} has eight states 
\begin{equation}
	\bm{x} = \{p_{im}, p_{em},X_{Oim},X_{Oem},\omega_t, \tilde{u}_{egr1},\tilde{u}_{egr2}, \tilde{u}_{vgt}\}
	\label{Eq:States}
\end{equation}
where $p_{im}$ and $p_{em}$ are the intake and exhaust manifold pressure respectively, $X_{Oim}$ and $X_{Oem}$ are the oxygen mass fractions in the intake and exhaust manifold respectively, $\omega_t$ is the turbo speed. $\tilde{u}_{egr1}$ and  $\tilde{u}_{egr2}$ are the two states for the EGR actuator dynamics, and $\tilde{u}_{vgt}$ represents the VGT actuator dynamics.
The mean value engine model is then expressed as
\begin{equation}
	\dot{\bm{x}} = f(\bm{x}, \bm{u}, n_e),
\end{equation}
where $\bm{u} = \{ u_\delta, u_{egr}, u_{vgt}\}$ is the control vector and $n_e$ is the engine speed; $u_\delta$ is the mass of injected fuel, $u_{egr}$ is the EGR actuator position and $u_{vgt}$ is the VGT valve position. A schematic for the engine model is shown in Fig. \ref{Figure:Schematic diagram of Engine}.

\par In this study, the states describing the oxygen mass fraction of the intake and exhaust manifold, i.e., $X_{Oim}$ and $X_{Oem}$, are not considered, since the rest of the states do not depend on these two states. Appendix section \ref{Appendix:Engine model} provides the governing equations that describe the complete engine model. Interested readers may also refer to \cite{Wahlstrom} and \cite{Nath_2023} for more details. The governing equations for the six state variables considered in this work are as follows:
\begin{equation}
	\dfrac{d}{dt} p_{im} = \dfrac{R_a T_{im}}{V_{im}}(W_c+W_{egr} - W_{ei}),
	\label{Eq:p_im}
\end{equation}
\begin{equation}
	\dfrac{d}{dt} p_{em} = \dfrac{R_e T_{em}}{V_{em}}(W_{eo} - W_t - W_{egr}),
	\label{Eq:p_em}
\end{equation}
\begin{equation}
	\dfrac{d}{d t}\omega_t = \dfrac{P_t\eta_m - P_c}{J_t\omega_t},
	\label{Eq:omega_t}
\end{equation}
\begin{align}
	\dfrac{d\Tilde{u}_{egr1}}{dt} & = \dfrac{1}{\tau_{egr1}}\left[u_{egr}(t-\tau_{degr}) - \Tilde{u}_{egr1}\right],
	\label{Eq:u_egr_1}\\
	\dfrac{d\Tilde{u}_{egr2}}{dt} & = \dfrac{1}{\tau_{egr2}}\left[u_{egr}(t-\tau_{degr}) - \Tilde{u}_{egr2}\right],
	\label{Eq:u_egr_2} \\
	\dfrac{d\Tilde{u}_{vgt}}{dt} & = \dfrac{1}{\tau_{vgt}}\left[u_{vgt}(t-\tau_{dvgt}) - \Tilde{u}_{vgt}\right].
	\label{Eq:u_vgt}
\end{align}
The calculation of $T_{em}$ in Eq. \eqref{Eq:p_em} (see Eq. \eqref{Eq:Append:T_em}) involves two additional equation Eqs. \eqref{Eq:T_1} and \eqref{Eq:x_r}. Here, $T_1$ is the temperature when the inlet valve closes after the intake stroke and mixing, and $x_r$ is the residual gas fraction:
\begin{equation}
	T_1 = x_rT_e + (1-x_r)T_{im},
	\label{Eq:T_1}
\end{equation}
\begin{equation}
	x_r = \dfrac{\Pi_e^{1/\gamma_a}x_p^{-1/\gamma_a}}{r_c x_v}.
	\label{Eq:x_r}
\end{equation}

\par In a real system, measurements corresponding to output states $p_{im}$, $p_{em}$, $\omega_t$, $W_{egr}$ and $T_{em}$ are made along with the control inputs ($\bm{u}$) and engine speed $n_e$ through a network of integrated onboard sensors. The intake ($p_{im}$) and exhaust ($p_{em}$) manifold pressures are measured using pressure sensors. The turbocharger shaft speed ($\omega_{t}$) is measured using a rotational speed sensor, while the EGR valve flow rate ($W_{egr}$) is measured using a flow venturi tube. $T_{em}$ may be measured using a thermocouple depending on the engine configuration. Some critical parameters of the diesel engine that are difficult to estimate on a running engine include the maximum effective area of the EGR valve ($A_{egrmax}$), compensation factor for non-ideal cycles ($\eta_{sc}$), the total heat transfer coefficient of the exhaust pipes ($h_{tot}$), and the maximum area of VGT ($A_{vgtmax}$). By estimating the variation in values of these parameters during engine operation, one can assess the health of the particular sub-system and identify potential sources of performance issues.

\par The objectives of the present study are twofold: (1) to predict the gas flow dynamics of all relevant engine states using a physics-informed deep learning framework, and (2) to identify critical unknown parameters in the system, given field measurements of $p_{im}$, $p_{em}$, $\omega_t$, $W_{egr}$, and $T_{em}$ and system models represented by equation \eqref{Eq:p_im}-\eqref{Eq:x_r}. We refer to this as the inverse parameter identification problem in this work. As discussed earlier, we aim to formulate a computationally efficient method suitable for use as engine health monitoring system. The training data required for this study is generated using the Simulink file \cite{Simulink_file} accompanied with \cite{Wahlstrom}. This Simulink file solves the differential equations discussed above with input vector $\bm{u}$ and engine speed $n_e$. Although we consider simulated data for training our neural network model, the input data (i.e. $\bm{u}$ and $n_e$) used to generate the simulation results are collected from actual engine test bed, thus representing real-world conditions. Additionally, we assess the effect of aleatoric uncertainty on our model's performance using Gaussian noise. Thus, we consider two cases for our study:
\begin{enumerate}
    \item Clean data: simulated noise-free data, 
    \item Noisy data: simulated data where appropriate Gaussian noise is added to $p_{im}$, $p_{em}$, $\omega_t$, $W_{egr}$, and $T_{em}$ to represent noisy sensor measurements expected in the field.
\end{enumerate}

Since the input vector $\bm{u}$ and engine speed $n_e$ are collected from actual engine running conditions, they are assumed to be inherently noisy and no further noise is added to the input signals.

\section{Methodology}
\label{Section:Methodology}
\subsection{Overview}
\label{Subsection:Basic philosophy}
We formulate a hybrid model combining DeepONet and PINN for the inverse problem described in the previous section. The method may be subdivided into four parts.

\begin{enumerate}[leftmargin=*]
    \item \textbf{Empirical relationships:} We consider various empirical relationships between variables in the engine model. In  prior study, \cite{Nath_2023} demonstrated how these empirical formulae can be approximated using deep neural networks (DNN). The empirical formulae are engine-specific, and DNNs can provide a generalized representation when the exact relationships are unknown by training the DNN with data from the specific engine. In addition, to account for the systematic (e.g. manufacturing) variations of the engine, we introduce stochastic variations into the empirical relationship by using dropouts in DNN.   
    \item \textbf{Independent States Approximation:} We note that the states describing the EGR and VGT dynamics are independent of the rest of the states, with no unknown parameters in these dynamics. We approximated these states using two DeepONets, one for the two EGR states ($\tilde{u}_{egr1}$ and $\tilde{u}_{egr2}$) and one for the VGT state ($\tilde{u}_{vgt}$) as discussed in section \ref{Subsection:DeepONet}. These DeepONets are pre-trained using simulation data and used as surrogates to predict the actuator states.    
    \item \textbf{Hybrid method:} We propose a hybrid method combining the pre-trained DNN, the pre-trained DeepONet and PINN model, as detailed in section \ref{Subsection:PINN for engine problem}. We also use self-adaptive weights \cite{McClenny_2020} for improved convergence in all examples of the proposed methods.
    \item \textbf{Transfer Learning Strategies:} Lastly, to reduce the computation cost (in terms of the number of epochs and trainable parameters), we consider two TL strategies: 
    \begin{enumerate}
        \item a hybrid PINN multi-stage transfer learning method, discussed in section \ref{Subsection:Transfer learning}, 
        \item a hybrid PINN few-shot transfer learning method, discussed in section \ref{Subsection:Few shot PINN}.
    \end{enumerate}
\end{enumerate}

Furthermore, as noted above, we use pre-trained networks for empirical formulae (DNNs) and for EGR and VGT states (DeepONets). These networks are trained using simulated data (we label this as laboratory data for nomenclature simplicity), which corresponds to the design value of the unknown parameters. However, in field, the values of the unknown parameters may change due to natural systematic changes over time. Thus, we test our proposed hybrid PINN method on a set of parameters different from the design parameters used for generating laboratory data for training our empirical networks and DeepONets. This helps us validate whether our hybrid PINN model can accurately identify parameters independently of the predictions from the pre-trained networks, thereby allowing us to analyze any potential bias originating from this hybrid approach. The different values of the unknown parameters considered for generating laboratory and field data are shown in Table \ref{Table:true value unknown parameters}.

\begin{table}[!h]
\centering
\caption{\textbf{True value of estimated parameters} considered for generating laboratory and field data conditions. The ``Laboratory data" is considered for training the DNNs for empirical formulae, DeepONets for EGR and VGT actuator dynamics, and training the body for few-shot TL. ``Field data" represents data used for training the hybrid PINN framework for parameter estimation, mutli-stage TL and few-shot TL. For confidentiality, we represent the true values for field parameter values as 1, and the laboratory data are scaled with respect to field data. The ``Mask function" ensures the unknown parameters are always positive, and ``Scaling" factor enables the conversion of parameters back to the physical domain.}
\label{Table:true value unknown parameters}
\begin{tabular}{lcccc} \hline
                    & $A_{egrmax}$ & $A_{vgtmax}$ & $\eta_{sc}$ & $h_{tot}$ \\ \hline
    Laboratory data & $1.333$ & $1.1663$ & $0.9179$ & $1$ \\ \hline
    Field data & $1$ & $1$ & $1$ & $1$ \\ \hline
    \multicolumn{5}{c}{Mask and scaling considered for PINN model} \\ \hline
    Mask function & $\exp(.)$ & $\exp(.)$ & $\text{softplus(.)}$ & $\exp(.)$  \\ \hline
Scaling & $\times10^{-4}$ & $\times10^{-4}$ & $\times1$ & $\times10$ \\ \hline
\end{tabular}
\end{table}
\subsection{Neural network surrogates for empirical formulae}
\label{Subsection:Neural network surrogates for empirical formulae}

The empirical formulae such as the polynomial functions for the volumetric efficiency ($\eta_{vol}$), effective area ratio function for EGR valve ($f_{egr}$), turbine mechanical efficiency ($\eta_{tm}$), effective area ratio function for VGT ($f_{vgt}$), choking function (for VGT) ($f_{\Pi_t}$), compressor efficiency ($\eta_c$), and volumetric flow coefficient (for the compressor) ($\Phi_c$) proposed in the mean value engine model in \cite{Wahlstrom} are approximated using neural networks in our study. Dropouts are used in the DNN models to account for systematic variations. In Table \ref{Table:Surrogate empirical formulae}, we present the six DNNs that approximate the empirical formulae along with their input and output variables.
\begin{table}[H]
\centering
\caption{\textbf{Neural network surrogates for empirical formulae} $\mathcal{N}_i^{(P)}(\bm{x}; \bm{\theta}_i^P)$, $i= 1,\dots, 6$ denotes the surrogate for the $i^{th}$ DNN parameterized by $\theta_i^P$ with input $\bm{x}$. All these networks have one output each.}
\label{Table:Surrogate empirical formulae}
\footnotesize
\begin{tabular}{C{1.5cm}|c|c|c|c|c|c} \hline
Surrogate & $\mathcal{N}_1^{(P)}(\bm{x}; \bm{\theta}_1^P)$ & $\mathcal{N}_2^{(P)}(\bm{x}; \bm{\theta}_2^P)$ & $\mathcal{N}_3^{(P)}(\bm{x}; \bm{\theta}_3^P)$ & $\mathcal{N}_4^{(P)}(\bm{x}; \bm{\theta}_4^P)$ & $\mathcal{N}_5^{(P)}(\bm{x}; \bm{\theta}_5^P)$ & $\mathcal{N}_6^{(P)}(\bm{x}; \bm{\theta}_6^P)$ \\ \hline
Input ($\bm{x}$) & $\{p_{im}, n_e\}$ & $\tilde{u}_{egr}$ & $\{\Tilde{u}_{vgt}, \Pi_t\}$ & $\{ \omega_t, T_{em}, \Pi_t \}$ & $\{W_c, \Pi_c\}$ & $\{T_{amb}, \Pi_c, \omega_t\}$\\ \hline
Output & $\eta_{vol}$ & $f_{egr}$ & $F_{vgt, \Pi_t}$ & $\eta_{tm}$ & $\eta_{c}$ & $\Phi_c$  \\ \hline
Output restriction$^\ddag$ & & $S(f_{egr})$ & $1.1\times S(F_{vgt, \Pi_t})$ & $\text{min}(0.818, \eta_{tm})$ & $\text{max}(0.2,S(\eta_c))$ & $S(\Phi_c)$ \\ \hline
Equation$^\dagger$ & \eqref{Eq:Append:eta_vol} & \eqref{Eq:Append:f_egr} & \eqref{Eq:Appendix:F_VGT_PI_T_Cal} & \eqref{Eq:Append:eta_tm} & \eqref{Eq:Append:eta c} & \eqref{Eq:Append:3}\\ \hline
\multicolumn{7}{l}{$^\ddag$ $S\longrightarrow$ sigmoid function}\\ \hline
\multicolumn{7}{l}{$^\dagger$ The empirical equations are discussed in section \ref{Appendix:Engine model}} \\ \hline
\end{tabular}
\end{table}

\par The DNNs for the empirical formulae are trained using a large laboratory dataset that captures all possible operating scenarios for the engine. We consider a mean square loss function for the training of these networks,
\begin{subequations}
\begin{align}
    \mathcal{L}_i(\bm{\theta}_i^P) & = \dfrac{1}{n_i}\sum_{j=1}^{n_i} \left[y_i^{(j)} - \hat{y}_i^{(j)}\right]^2,  \;\;\;\;\; i=1,2,\dots,6\\
    & =  \dfrac{1}{n_i}\sum_{j=1}^{n_i}\left[y_i^{(j)} - \mathcal{N}_i^{(P)}(\bm{x}_i;\bm{\theta}_i^P)^{(j)}\right]^2
\end{align}
\end{subequations}
where the subscript $i=1,2,\dots,6$ indicate different neural networks shown in Table \ref{Table:Surrogate empirical formulae}, $n_i$ represents the number of labelled training data for $i$\textsuperscript{th} neural network. $\bm{x}_i$ denotes the input corresponding to the $i$\textsuperscript{th} network, $\hat{y}_i$ and $y_i$ represent the estimated output and the corresponding labelled values, respectively for the $i$\textsuperscript{th} network. The laboratory data used for training each neural network have been discussed in previous work \cite{Nath_2023} along with detailed calculations for each labelled data. We also consider a 3\% Gaussian noise in the simulated data of the output variables during training to account for the field measurement conditions. These networks are trained using the Adam optimizer. After training, we integrate these DNNs into our hybrid PINN framework to replace the empirical models. These pre-trained neural networks are shown with the output $g(\hat{y})$ in Fig. \ref{Figure:PINN_DeepONet} and Fig. \ref{fig:few_shot_schematic} in sections \ref{Subsection:PINN for engine problem} and \ref{Subsection:Few shot PINN}.

\par We apply a 1\% dropout for the hidden layers of these networks. Notably, dropout is employed not only during the training of these networks but also during the PINN optimization process for the inverse problem and prediction. This helps us to analyze the potential variation in PINN estimates expected due to systematic (e.g., manufacturing) variability for real systems. We also added a 3\% Gaussian noise to the output state variables during training to replicate the expected noise in laboratory measurements.

\subsection{DeepONets for EGR and VGT states}
\label{Subsection:DeepONet} 
Based on the universal approximation theorem of operators \cite{Chen_1995_Universal}, Lu et al. \cite{Lu_2021_deepOnet} proposed the Deep Operator Network (DeepONet), which maps an infinite-dimensional function ($u\in \bm{U}$) to another infinite dimensional output function ($v\in \bm{V}$) in the Banach space. The conventional DeepONet consists of two DNNs, known as branch and trunk networks. The trunk network takes the spatio-temporal information (e.g. time $t$, domain points ${x,y,z}$) as input and provides the bases function as output. The branch network takes the input function ($u\in \bm{U}$) at fixed points and approximates the coefficients for each basis function. Importantly, the DeepONet architecture does not impose restrictions on the type of networks used for trunk and branch networks. The ability of DeepONet to learn complex mapping between input and output functional spaces makes it suitable for various applications in scientific machine learning. Mathematically, DeepONet learns an operator that maps $\mathcal{G}_{\bm{\theta}}:\bm{U}\rightarrow\bm{V}$, and the output is given as 
\begin{equation}
    v(t) \approx \mathcal{G}_\theta(u)(t) = \sum_{i=1}^p br(u)_i tr(t)_i,
\end{equation}
where $br(u)_i$ and $tr(t)_i$ are the outputs of the branch and the trunk nets, respectively; $p$ is the number of neurons at the last layer of each of the branch and trunk networks, which corresponds to the number of base functions used in approximation of the spatio-temporal points. $u_1, u_2, \dots u_n$ are the input function sampled at $n$ fixed points. DeepONet has been applied to various engineering problems in its standard form or with appropriate modifications in its network and input structures. A comprehensive review and comparison of several variants can be found in \cite{Lu_2022_Fair_comparision}. As discussed in section \ref{Section:Introduction}, Kumar et al. \cite{Kumar_2023} utilized DeepONet for predicting the gas flow dynamics of a diesel engine, albeit limited to accurate predictions of time sequences of up to five seconds. To extend the prediction length, the authors predict results every five seconds and stitch them to obtain long-duration results. In this case, the initial conditions are required every five seconds. Alternatively, the authors consider the predicted output of the DeepONet as the initial condition for the next time sequence. However, the authors in \cite{Kumar_2023} observed that this approach results in higher error accumulation over time.

\par We note here that the engine is a continuous dynamical system, and this makes it challenging to identify proper operating sub-spaces for input-output mapping for the operator network learning. Hence, it is important to study the driving behaviour (engine running condition) over large duration and sample sizes. An appropriate window size (time duration) is required in which correct input-output function spaces can be assumed while preserving the dynamic behaviour of the system. Here, we consider a 60-sec window and approximate the independent state variables of the engine ($\tilde{u}_{egr1}$, $\tilde{u}_{egr2}$, $\tilde{u}_{vgt}$) using Causality-DeepONet \cite{Liu_2022_Causality}, as proposed by Liu et al. \cite{Liu_2022_Causality}. Causality-DeepONet integrates causality and convolution of the input function in its formulation, which enhances accuracy in approximating oscillatory dynamic problems with a smaller training sample set. We find that this framework is particularly useful in capturing the rapidly changing dynamics of a diesel engine signal. 
\par In this study, we use Causality-DeepONets to approximate the dynamics of the EGR actuator dynamics, i.e., $\tilde{u}_{egr1}$ (Eq. \eqref{Eq:u_egr_1}) and $\tilde{u}_{egr2}$ (Eq. \eqref{Eq:u_egr_2}) and the VGT actuator dynamics $\tilde{u}_{vgt}$ (Eq. \eqref{Eq:u_vgt}). Details of the two DeepONets are shown in Table \ref{Table:DeepONet surrogates}. From hereon, we refer to Causality-DeepONet simply as DeepONet unless stated otherwise. These DeepONets approximate solutions to non-homogeneous initial value problems by learning the operator mapping between input function ($u$) and the initial condition ($\tilde{u}_0$) to the output ($\tilde{u}(t))$ using individual branch networks for the input function and initial condition, as described in Table \ref{Table:DeepONet surrogates}. We note that the equations governing EGR and VGT actuator dynamics are first-order differential equations that can be computed efficiently using fast numerical methods. However, discrepancies between the mathematical model and experimental data were observed in \cite{Wahlstrom}. By employing DeepONets trained on historical data, we can obtain a better representation of real system dynamics as compared to a numerical method that relies on physical equations. Hence, our approach of using DeepONet surrogates for predicting actuator dynamics is likely to show better correlation when used on real system. We can write the operator mapping from input function ($u$) and initial condition ($\tilde{u}_0$) to the dynamics of the output ($\tilde{u}(t)$) as 
\begin{equation}
    \mathcal{G}_{\bm{\theta}} : u, \tilde{u}_0\rightarrow \tilde{u}(u, \tilde{u}_0)(t)
\end{equation}
\par We have discussed the DeepONet architecture, training, and numerical results in Appendix \ref{Subsection:Numerical:DeepONet}

\begin{table}[H]
\centering
\caption{\textbf{DeepONet surrogates} considered for approximation of EGR and VGT states. $\mathcal{D}_i(t,\bm{x};\bm{\theta}_{DP,i}), \, i = 1,2$ denotes the DeepONet surrogate for the $i^{th}$ DeepONet parameterized by $\bm{\theta}_{DP,i}$. $\mathcal{D}_1(t,\bm{x};\bm{\theta}_{DP,1})$ approximates the two EGR states while $\mathcal{D}_2(t,\bm{x};\bm{\theta}_{DP,2})$ approximates the VGT state. The input to the trunk is time $t$ while $\bm{x}$ is the input to the branch. A total of four branches are used for  $\mathcal{D}_1(t,\bm{x};\bm{\theta}_{DP,1})$ with two branches each taking $u_{degr}$ as input and the other two take initial condition $\{\tilde{u}_{egr1, 0}$, $\tilde{u}_{egr2, 0}\}$ as input. }
\label{Table:DeepONet surrogates}
\begin{tabular}{L{3cm}|C{5cm}|C{2.5cm}}
\hline
 DeepONet  & $\mathcal{D}_1(t,\bm{x};\bm{\theta}_{DP1})$ & $\mathcal{D}_2(t,\bm{x};\bm{\theta}_{DP2})$ \\ \hline
 No. of branch & 4 & 2 \\ \hline
 Branch Input ($\bm{x})$ & $\{u_{degr}$, $\tilde{u}_{egr1,0}\}$, $\{u_{degr}$, $\tilde{u}_{egr2,0}\}$ & $u_{dvgt}$, $\tilde{u}_{vgt,0}$ \\
 \hline
 Output Variable    & $\{\tilde{u}_{egr1}$, $\tilde{u}_{egr2}\}$ & $\{\tilde{u}_{vgt}\}$ \\ \hline
 Eq. & \eqref{Eq:Append:u_egr_1} and \eqref{Eq:Append:u_egr_2} & \eqref{Eq:Append:u_vgt} \\ \hline
 \multicolumn{3}{l}{For equations, refer Appendix.} \\ \hline
 \multicolumn{3}{l}{
      For details on DeepONet architecture, training, and results, refer Appendix \ref{Subsection:Numerical:DeepONet}
      } \\ \hline
\end{tabular}
\end{table}

\subsection{A hybrid model using DeepONet and PINN for the engine problem}
\label{Subsection:PINN for engine problem}
In the previous section (\ref{Subsection:DeepONet}), we discussed how the three states corresponding to EGR and VGT can be approximated using DeepONet surrogates. In section \ref{Subsection:Neural network surrogates for empirical formulae}, we introduced how we use DNNs to estimate empirical relationships in the engine model. In this section, we discuss in detail how we construct a hybrid model combining DeepONet, empirical DNNs, and PINN to solve the inverse problem of model-driven parameter identification. We propose a hybrid PINN model combining DeepONet and DNN to predict the gas flow dynamics and estimate unknown parameters. Preliminary background on use of PINNs to approximate the state variables of an ODE is presented in Appendix \ref{Appendix:PINN and DeepONet}

\par Fig. \ref{Figure:PINN_DeepONet} illustrates a schematic of the proposed hybrid model comprised of PINN and DeepONet. The first part of hybrid PINN model comprises a fully connected neural network and approximates the solution to a specific differential equation. The second part is the physics-informed segment, in which we encode the corresponding differential equation using automatic differentiation \cite{Baydin_2018}. We denote the unknowns in the equation as $\bm{\Lambda}$. If the unknowns are constant, these are treated as trainable scalar variables. If the unknowns are fields, these are approximated using DNN(s). The empirical formulas are approximated using DNN as discussed in section \ref{Subsection:Neural network surrogates for empirical formulae}. Additionally, the independent states related to actuator signals are approximated using DeepONets. These states (or the function of these states) may be incorporated into the loss function for PINN training. The loss function ($\mathcal{L}(\bm{\theta}; \bm{\mathcal{\varLambda}})$ for solving the inverse problems consists of two components: the physics loss ($\mathcal{L}_\text{phy}(\bm{\theta}; \bm{\mathcal{\varLambda}})$, and the data loss ($\mathcal{L}_\text{data}(\bm{\theta}; \bm{\mathcal{\varLambda}})$) 
\begin{equation}
    \mathcal{L}(\bm{\theta}; \bm{\mathcal{\varLambda}}) = \lambda_1 \mathcal{L}_\text{phy}(\bm{\theta}; \bm{\mathcal{\varLambda}}) + \lambda_2\mathcal{L}_\text{data}(\bm{\theta}; \bm{\mathcal{\varLambda}})
    \label{Eq:loss_inverse}
\end{equation}
where $\bm{\theta}$ denotes the set of parameters of the neural network, i.e. the weights, $\bm{W}$ and the biases, $\bm{b}$; $\bm{\mathcal{\varLambda}}$ are the unknown scalar parameters to be estimated. The parameters $\lambda_1$ and $\lambda_2$ are the weights which control the convergence of each loss component. These $\lambda$ values may be constant or self-adaptive \cite{McClenny_2020} depending on the problem setup and solution method. The data loss may be considered as the mean squared error (MSE) loss of the known labelled data, given by: 
\begin{equation}
    \mathcal{L}_\text{data}(\bm{\theta}; \bm{\mathcal{\varLambda}}) = \frac{1}{M}\sum^M_{i=1} \left[\hat{y}(t_i; \bm{\theta}) - y(t_i)\right]^2,
\end{equation}
where $M$ represents the number of measured points, $\hat{y}(t_i; \bm{\theta})$ and $y(t_i)$ denote the prediction of neural network and ground truth data, respectively, at time $t_i$. For multiple quantities, the data loss may be considered as the summation of individual losses for each quantity. The physics loss may be written as
\begin{align}
    \mathcal{L}_\text{phy}(\bm{\theta}; \bm{\mathcal{\varLambda}}) = \frac{1}{N}\sum^N_{i=1} \left[r(t_i; \bm{\theta}; \bm{\mathcal{\varLambda}})\right]^2,
\end{align}
where $N$ is the number of collocation points chosen for solving the equation and $r(t_i; \bm{\theta}; \bm{\mathcal{\varLambda}})$ represents the residual of the corresponding differential equation. The residual $r(t_i; \bm{\theta}; \bm{\mathcal{\varLambda}})$ should ideally be zero throughout the entire domain (or time duration). The objective, therefore, is to obtain the optimal parameters ($\bm{\theta}$) of the neural network as well as the unknown parameters ($\bm{\mathcal{\varLambda}}$) in the system. We minimize the loss function Eq. \eqref{Eq:loss_inverse} to obtain the optimal values of these parameters. Furthermore, we also incorporate self-adaptive weights \cite{McClenny_2020} in the loss. We minimize the loss function with respect to the parameters ($\bm{\theta}$ and $ \bm{\varLambda}$) and maximize with respect to the self-adaptive weights ($\lambda$). Thus, the optimization process may be written as,
\begin{equation}
    \min\limits_{\bm{\theta}, \bm{\varLambda}}\;\max\limits_{\lambda} \mathcal{L}(\bm{\theta},\varLambda)
\end{equation}
\par To integrate the complete hybrid model, we combine the DNN models used for approximating the empirical formulae as discussed in section \ref{Subsection:Neural network surrogates for empirical formulae} and the two DeepONet models for approximating actuator states $\tilde{u}_{egr1}$, $\tilde{u}_{egr2}$ and $\tilde{u}_{vgt}$ with the PINN DNN model, as shown in Fig. \ref{Figure:PINN_DeepONet}. We fix the parameters of DeepONets and DNNs (for the empirical formulae) when training the hybrid PINN model. Eq. \eqref{Eq:loss_inverse}, demonstrates how PINNs can be used for solving one differential equation. For systems involving multiple equations, one can either use one DNN with multiple outputs or multiple DNNs each with one output. The loss function in Eq. \eqref{Eq:loss_inverse} can be extended to include the loss function for all the equations and data by summing up individual loss functions for each equation and data. This hybrid approach leverages pre-trained networks and a PINN network that can be trained on live data for efficient and accurate estimation of engine parameters and state variables. 
\begin{figure}[!t]
    \centering
    \includegraphics[width=0.95\textwidth]{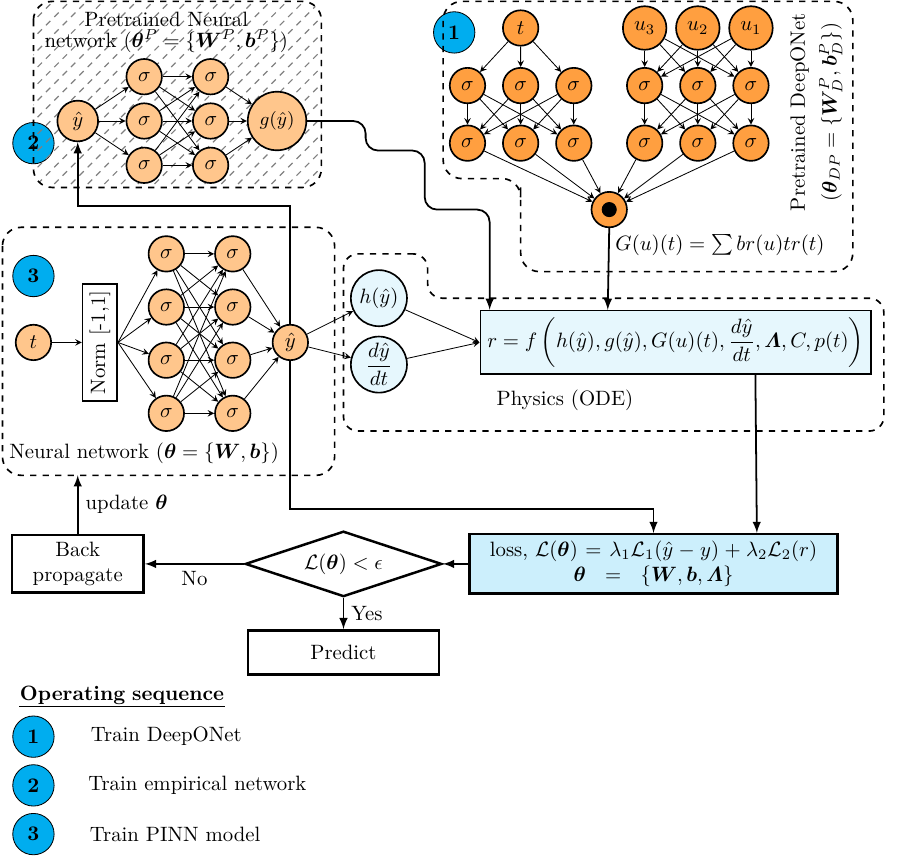}
    \caption{\textbf{Schematic of a hybrid model of DeepONet and physics-informed neural networks (PINNs) for inverse problems.} In the left part of the figure, a PINN approximates the solution $y$ to a differential equation with time $t$ as input. The top left section (enclosed within black hatched lines) represents another DNN which takes as input $\hat{y}$ (and potentially other inputs such as ambient condition) and outputs a function $g(\hat{y})$. This network is pre-trained with labelled laboratory data. On the top right section of the figure (enclosed in black dashed line), a DeepONet model takes inputs of time $t$ and an input function $u$ to approximate a state $G(u)(t)$. This network is also pre-trained with labelled laboratory data. The physics/residual loss is shown in the bottom right part of the figure (enclosed within the blue dashed line). The PINN approximates the solution to differential equations. The total loss $\mathcal{L}(\bm{\theta})$ includes the loss of equation as well as the data. Weights $\lambda_1$ and $\lambda_2$ are applied to the data and physics loss respectively, and may be fixed or adaptive depending on the problem and solution method. $\bm{\theta} = \{\bm{W}, \bm{b}, \bm{\Lambda}\}$ represents the parameters of the PINN, where $\bm{W}$ and $\bm{b}$ are the weights and biases, and $\bm{\Lambda}$ are the unknown parameters of the ODE; $\sigma$ denotes the activation function used for the PINN model, and $r$ is the residual for the equation. Further, $\bm{\theta}^P = \{\bm{W}^P, \bm{b}^P\}$ denotes the parameters of the pre-trained neural network $g(y)$. Similarly, $\bm{\theta}_{DP} = \{\bm{W}_{DP}, \bm{b}_{DP}\}$ represents the parameters of the pre-trained DeepONet $G(u)(t)$.}
    \label{Figure:PINN_DeepONet}
\end{figure}

\par In section \ref{Subsection:Problem setup}, we discussed the inverse problem of interest for this study. Our objectives are to: (1) learn the dynamics of the six gas flow dynamics state estimates, three of which are approximated using two DeepONets, and (2) infer the unknown parameters in the system, given field measurements of system variables $\{p_{im}, p_{em}, \omega_t, W_{egr}, T_{em}\}$ as well as Eqs. \eqref{Eq:p_im}-\eqref{Eq:x_r}, using PINNs. To achieve these objectives, we utilize four DNNs as surrogates for the solutions to different equations, as shown in Table \ref{Table:FNN for PINN}. We encode the physics loss corresponding to Eqs \eqref{Eq:p_im}-\eqref{Eq:x_r} (EGR, VGT actuator equations are not included as these are approximated using DeepONets) using automatic differentiation and define the physics-informed loss as:
\begin{equation}
    \begin{split}
    \mathcal{L}(& \bm{\theta}, \bm{\mathcal{\varLambda}}, \bm{\lambda}_{p_{im}}, \bm{\lambda}_{p_{em}}, \bm{\lambda}_{\omega_t}, \bm{\lambda}_{W_{egr}}, \lambda_{T_{em}}, \lambda_{T_1}) \\ = & \mathcal{L}_{p_{im}} + \mathcal{L}_{p_{em}} + \mathcal{L}_{\omega_{t}} + 100\times \mathcal{L}_{x_{r}} + \lambda_{T_1}\times\mathcal{L}_{T_{1}} + \\
    & \mathcal{L}^{ini}_{p_{im}} + \mathcal{L}^{ini}_{p_{em}} + \mathcal{L}^{ini}_{\omega_{t}} + 100\times \mathcal{L}^{ini}_{x_{r}} + 100\times \mathcal{L}^{ini}_{T_{1}} + \\
    & \mathcal{L}^{data}_{p_{im}}(\bm{\lambda}_{p_{im}}) + \mathcal{L}^{data}_{p_{em}}(\bm{\lambda}_{p_{em}}) +     \mathcal{L}^{data}_{\omega_t}(\bm{\lambda}_{\omega_t}) + \\ & \mathcal{L}^{data}_{W_{egr}}(\bm{\lambda}_{W_{egr}}) + \lambda_{Tm}\mathcal{L}^{data}_{T_{em}},
    \end{split}
\label{Eq:Loss total loss}
\end{equation}
where, $\mathcal{L}_X$, $X=\{p_{im}, p_{em}, \omega_t, x_r, T_1\}$ are the physics loss, $\mathcal{L}^{ini}_X$, $X=\{p_{im}, p_{em}, \omega_t, x_r, T_1\}$ are the initial condition loss and $\mathcal{L}^{data}_X$, $X=\{p_{im}, p_{em}, \omega_t, W_{egr}, T{em}\}$ are the data loss, $\lambda$s are the self adaptive weights. We discussed a detailed flowchart for the calculation of physics-informed losses for the hybrid PINN model in section \ref{Section:Appendix:Flowchart}, and the detailed loss function in the Appendix \ref{Suppl:Detail loss function}.
\begin{table}[!h]
    \centering
    \caption{\textbf{Neural network for PINN:} Neural network surrogates considered to approximate the state variables and $x_r$ and $T_1$ for solving the inverse problems. $\mathcal{N}_i(t; \bm{\theta}_i), i = 1, ..., 4$ denotes the surrogate for the $i^{th}$ DNN parameterized by $\bm{\theta}_i = \{\bm{W}, \bm{b}\}_i$. The input to these DNN is time $t$. DNN $\mathcal{N}_1(t; \bm{\theta}_1)$ has two outputs for $\{p_{im}$ and $p_{em}\}$; the remaining DNNs have only one output. The multihead networks are pre-trained offline. During multi-stage transfer learning the number of parameters changes with epoch as discussed in the } 
    \label{Table:FNN for PINN} 
    \resizebox{\textwidth}{!}{%
    \begin{tabular}{L{3cm} |C{3.85cm}|C{3.75cm}|C{3.5cm}|C{3cm}}
    \hline
    Surrogate  & $\mathcal{N}_{1}(t; \bm{\theta}_1)$ &  $\mathcal{N}_{2}(t; \bm{\theta}_2)$ & $\mathcal{N}_{3}(t; \bm{\theta}_3)$ & $\mathcal{N}_{4}(t; \bm{\theta}_5)$ \\ \hline
    Output Variables   & ${p}_{im}$, ${p}_{em}$ & ${x}_{r}$ & ${T}_{1}$ & ${\omega}_{t}$ \\ \hline
    Hidden layer activation   & Tanh & Tanh & Tanh & Tanh \\ \hline
    Output transformation $^\dagger$ & $S_p(p_{em})$, $S_p(p_{im})$ & $S(x_r)$ &  $S_p(T_1) + 230./300.0$ & $S_p(\omega_t$)\\ \hline
    Output scaling & $\times10^5$ & $\times1$ & $\times300$ & $\times5.0\times10^3$ \\ \hline
    Equation & \eqref{Eq:p_im} and \eqref{Eq:p_em} & \eqref{Eq:x_r} & \eqref{Eq:T_1} & \eqref{Eq:omega_t}\\ \hline
    
    \multicolumn{5}{c}{Network size} \\ \hline
    Multi-stage TL & [1, 15, 15, 15, 2] & [1, 15, 15, 15, 1] & [1, 15, 15, 15, 1] & [1, 15, 15, 1] \\ \hline
    Multihead network & $[1,250,250,250,125,480]$ & $[1,170, 170, 170, 50, 240]$ & $[1, 50,50, 20, 240]$ & $[1, 120,120,70, 240]$\\ \hline
    During few-shot TL & $[125,1]$ and $[125,1]$ & $[50,1$] & $[20,1]$ & $[70,1]$\\ \hline
    \multicolumn{5}{l}{$^\dagger$ \hspace{0.2cm} $S_p\rightarrow$ Softplus(), \hspace{0.2cm} $S\rightarrow$ sigmoid()} \\ \hline
\end{tabular}}
\end{table}

\par The optimal parameters of the neural networks for the hybrid PINN model (ref. Table \ref{Table:FNN for PINN}), along with unknown parameters (ref. Table \ref{Table:true value unknown parameters}) and the self-adaptive weights ($\lambda$), are obtained by optimizing the loss function from Eq. \eqref{Eq:Loss total loss} using the Adam \cite{Kingma_2014adam} optimizer. The optimization process is divided into two phases. In the first phase, we optimize the PINN network parameters, unknown parameters, and self-adaptive weights for a certain number of epochs. At the end of this phase, we freeze the values of the self-adaptive weights. Thereafter, we optimize only the parameters of the hybrid PINN model and the unknown parameters using Adam, keeping the self-adaptive weight constant with the values obtained at the end of the first phase. As discussed earlier, the parameters of the pre-trained neural networks for the empirical formula and the DeepONets are kept frozen during the hybrid PINN optimization process (both phases).

\subsection{Multi-stage Transfer learning}
\label{Subsection:Transfer learning}
In the preceding sections, we have outlined a method for predicting unknown parameters and gas flow dynamics of a diesel engine. In this method, the parameters of the networks, unknown parameters and the self-adaptive weights are initialized randomly and trained to obtain the optimal parameters. Due to the stochastic initialization of the network parameters, unknown parameters and self-adaptive weights, significant computation time is required to converge to an accurate estimate. Furthermore, any changes in the control input vector $\bm{u} = \{u_{egr}, u_{vgt}, u_\delta\}$, engine speed $n_e$ or ambient conditions necessitates re-training the entire PINN model.
\par
\begin{figure}[t]
    \centering
    \includegraphics[width = 0.85\textwidth]{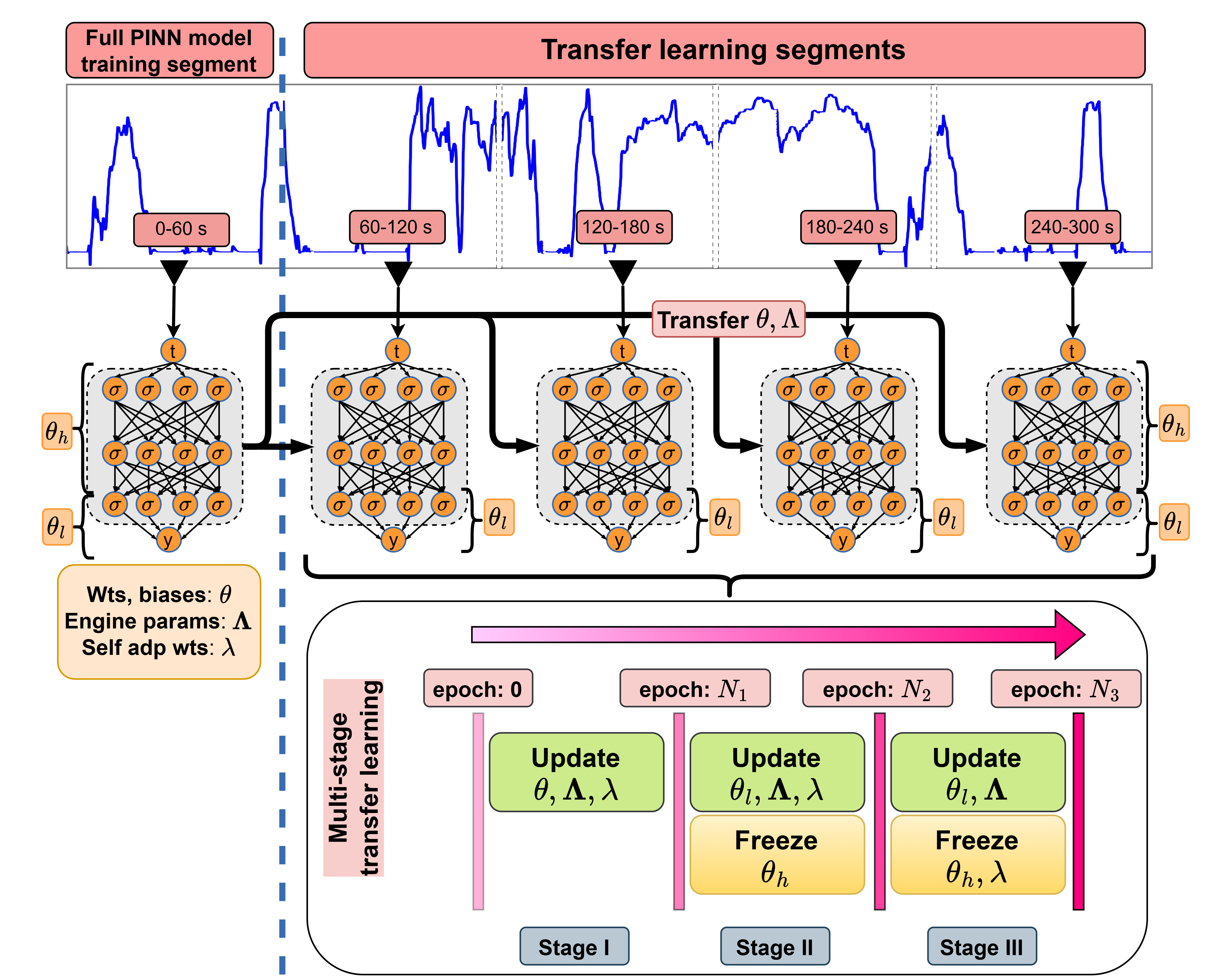}
    \caption{\textbf{Schematic for Multi-stage TL} showing the network training protocol for different segments of time. In this approach, the hybrid PINN model is trained for the first segment of time ($0\sim60$ sec). Thereafter, for the subsequent time segments ($60\sim300$ sec in each one-minute segment), we transfer the network parameters ($\bm{\theta}$) and engine parameter estimated ($\bm{\Lambda}$) from the hybrid PINN model trained on the first time segment. Each segment uses a different control vector $\{u_\delta, u_{egr}, u_{vgt}$\}, and engine speed $n_{e}$. The adaptive weights ($\bm{\lambda}$) are initialized at the start of training at epoch=0. The network weights and biases of all the layers ($\bm{\theta}$), self-adaptive weights ($\bm{\lambda}$), and unknown parameters ($\bm{\Lambda}$) are updated in Stage I training phase. In Stage II, only weights and biases of the last layer of the PINN model ($\bm{\theta_{l}}$) are trained while freezing all the hidden layers ($\bm{\theta_{h}}$). Additionally, $\bm{\lambda}$ and $\bm{\Lambda}$ continue to be updated in this phase. In Stage III, we continue to update $\bm{\theta_{l}}$ and $\bm{\Lambda}$ while the self-adaptive weights $\bm{\lambda}$ are fixed to the values obtained at the end of Stage II. In total, for predicting across a 5-minute time window, we train five models, with the model trained for segment 1 (0-60 sec) serving as the reference model from which the remaining models acquire knowledge during the TL phase. } 
    \label{Figure:Transfer learning schematic}
\end{figure}

 To mitigate the challenge related to computational time, we implement a multi-stage TL paradigm to train the hybrid PINN model individually for each one-minute time segment as shown in Fig. \ref{Figure:Transfer learning schematic}. Transfer learning \cite{trnf_learn_caruana1997, trnf_learn_pratt1992discriminability, Goswami_2020_Trnsfer} utilizes knowledge gained from training a model on one dataset (typically referred to as source data) to enable knowledge transfer to train on a different but similar dataset (known as target data). We demonstrate our methodology using a five-minute long data signal divided into five one-minute segments. First, the complete hybrid PINN model is trained with the given control vector $\{u_\delta, u_{egr}, u_{vgt}$\}, and engine speed $n_{e}$ for time segment $0 - 60$ sec. The weights and biases of the network, denoted by $\bm{\theta}$, are initialized randomly. The unknown parameters for the engine,  $\bm{\Lambda}:= \{A_{egrmax}, A_{vgtmax}, \eta_{sc}, h_{tot}\}, \bm{\Lambda}\in \mathbb R^+$ are initialized with a random positive value (discussed in detail in section \ref{Subsection:Results for PINN inverse problem}). Similarly, the self-adaptive weights $\bm{\lambda}$ are also initialized randomly at the beginning of the training. Thereafter, the PINN network is trained as per the process outlined in section \ref{Subsection:PINN for engine problem}, and the network parameters and unknown parameters $\bm{\Lambda}$ are saved. In the TL stage, these saved parameter values are used for the initialization of network parameters (weights and biases) and unknown parameters. In Fig. \ref{Figure:Transfer learning schematic}, we illustrate and describe in detail the different stages of the proposed TL scheme.

\subsection{Few-shot transfer learning}
\label{Subsection:Few shot PINN}
In the previous sections, first, we developed a hybrid DeepONet and PINN model and proposed a multistage TL for the inverse problem of engine parameter estimation discussed in section \ref{Subsection:Problem setup}. In this proposed method, the model needs to be trained for the first time segment starting with random initialization of the weights and biases. To calculate the physics loss, the derivative for the output of the neural network with respect to the input needs to be calculated at each epoch. Although TL improves the computation time, the computational time required by the proposed framework is still high for field applications. One approach to improve the computational efficiency of this model is by reducing (or eliminating) the gradient calculation involved in PINN training. This may be achieved by considering the derivative to be a linear combination of some base vector. For example, in function connection \cite{Schiassi_2021}, multi-head network \cite{desai_2022_oneshot} or reservoir computing \cite{2021_Alberto}, we generally do not train the hidden layer and approximate the derivative of the output as a linear combination of the derivative of the hidden layer. Desai et al. \cite{desai_2022_oneshot} incorporated one-shot inference on new tasks by training the PINN network on a training task set and then freezing all except the last layer that is trained for the new task. This approach enables one to obtain the network output as a linear combination of the output of the frozen layer with the last layer. This reduces the need for automatic differentiation to evaluate the gradients of outputs in each epoch of training with respect to inputs, which is computationally expensive. The derivative is calculated as the linear combination of the derivative of the hidden layers. We adopt this TL paradigm here, called few-shot transfer learning.

\par
\begin{figure}[!t]
    \centering
    \includegraphics[width = 0.9\textwidth]{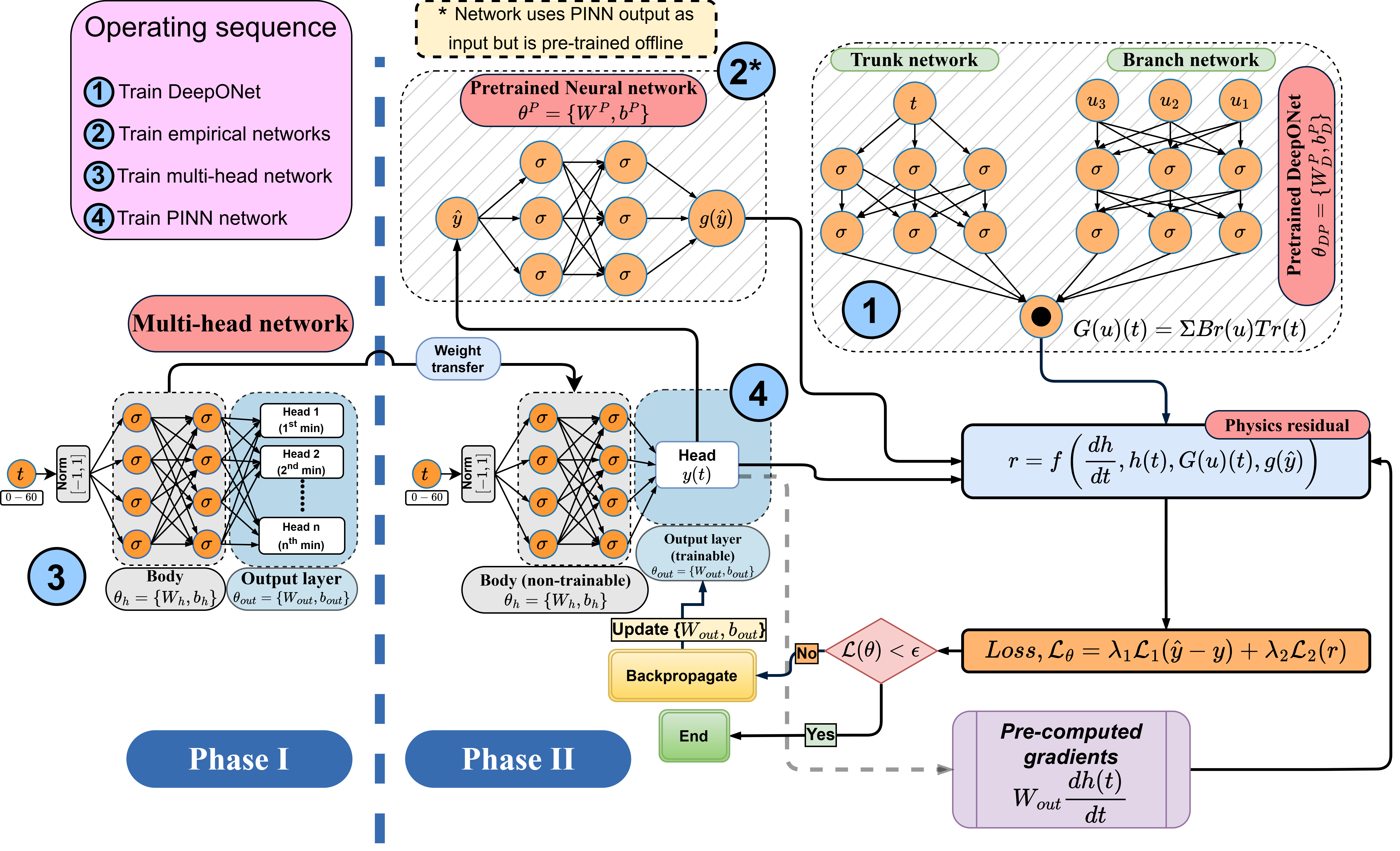}
    \caption{\textbf{Schematic diagram for the few-shot TL architecture with multi-head network}. The left side of the plot shows Phase I of the few-shot TL approach. In this phase, we train the multi-head networks using laboratory data where each head approximates one minute of response (corresponding to a different time zone). Input to this network is time $0 - 60$ sec, normalized between $[-1,1]$. Phase I training is a data-driven offline training. On the right side of the plot, Phase II, which involves TL, is shown. In this phase, we transfer the hidden layer and derivative of the output of the last hidden layer from the network trained in Phase I. We randomly initialize the weights and biases of the last layer along with the unknown parameters. The output and the derivative of the output are predicted using Eqs. \eqref{Eq:Few shot output} and \eqref{Eq:Few shot derivative}, respectively. We formulate the hybrid PINN model by combining the PINN network with DeepONet and empirical DNN. The total loss ($\mathcal{L}_{\bm{\theta}}$) is calculated as the weighted sum of the physics and data loss. The weights $\lambda_1$ and $\lambda_2$ may be constant or adaptive depending on the problem setup and solution method. The trainable parameters are the weights and bias of the output layer of the PINN network ($\bm{\theta}_{out} = \{\bm{W}_{out}, b_{out}\}$), the unknown parameters of the equation, the self-adaptive (in case of self-adaptive) weights. The weight and biases of the hidden layer of the PINN network, the parameters of the pre-trained network for empirical formula, and the parameters of the DeepONets remain fixed during the optimization process.}    
    \label{fig:few_shot_schematic}
\end{figure}

 Fig. \ref{fig:few_shot_schematic} illustrates a schematic diagram of a few-shot TL combined with pre-trained DNN and DeepONet. Here, we note that the objectives of the pre-trained DNN and DeepONet are the same as discussed in the previous section. Hybrid PINN, using few-shot TL, may be divided into two phases. The first phase (Phase I) is an offline training phase with multiple heads for unknown variables. In this phase, we train the network offline with multiple heads, each head approximating one minute (corresponding to a different time zone) of the same output variable. The input to the network is time $0 - 60$ sec, normalized between $[-1,1]$. The first phase of offline training aims to obtain good universal basis functions (hidden layers), which can be combined to obtain appropriate output in Phase II for solving the inverse problem using PINN. In the first phase, the network is trained in a data-driven setting in an offline mode. Once the network is trained, the hidden layers (universal basis function) are transferred to the hybrid PINN model to estimate the variables in Phase II.
\par In Phase II, we transfer the hidden layer of the multi-head network trained in the first phase. We randomly initialize the last layer ($\bm{W}_{out}, b_{out}$) and set up the hybrid PINN inverse problem for one-minute duration independently. In this phase, only the last layer of the network and the unknown parameters of the equation and the self-adaptive weights are optimized. The estimated output is obtained as
\begin{equation}
    y(t) = \bm{h}(t) \bm{W}_{out} + b_{out}, 
    \label{Eq:Few shot output}
\end{equation}
where $\bm{h}(t)$ is the output of the last hidden layer, $\bm{W}_{out}$ and $b_{out}$ are the weights and biases of the output layer. The derivative of the output with respect to the input variable is
\begin{equation}
    \dfrac{dy(t)}{dt} = \dfrac{d\bm{h}(t)}{dt}\bm{W}_{out}.
    \label{Eq:Few shot derivative}
\end{equation}
Since in Phase II, the hidden layers are fixed (not trained), the derivative $d\bm{h}(t)/dt$ does not change and can be kept outside the training loop for efficiency. Thus, the derivatives need to be calculated only once outside the training loop and we approximate the derivative of the output as a linear combination of the derivative of the hidden layer with the same combination matrix ($\bm{W}_{out}$). In this case, we trained only the last layer, such that we can keep the derivative outside the training loop using Eq. \ref{Eq:Few shot derivative}, thus reducing the computational cost. The loss function for few shot TL is the same as that of hybrid PINN given by Eq. \eqref{Eq:Loss total loss}. Note here that we utilize a non-linear activation function in the output layer and hence we employ the chain rule for calculating the derivative of output with respect to input. We discuss this in detail in the numerical result section \ref{Section:Numerical results: Few-shot transfer learning}.

\subsection{Data generation}
\label{Subsection:Data generation}
As discussed earlier, we utilize two types of networks: (1) pre-trained DNN for the empirical formula and DeepONets (pre-trained) for the independent states ($\tilde{u}_{egr1}$, $\tilde{u}_{egr2}$ and $\tilde{u}_{vgt}$), (2) PINN networks used for online learning. Thus, two distinct types of data is required. We label the data used for training the empirical formulae networks and DeepONets for EGR and VGT states as laboratory data. The data used for solving the inverse problem through PINNs is labeled as field data, which is used to simulate a real testing scenario. A prior study \cite{Nath_2023} provides an in-depth discussion on the laboratory data required for training each DNN for empirical formulae and other necessary details. In the absence of laboratory and/or field data, we use Simulink \cite{Simulink_file} to generate both datasets presented in Table \ref{Table:true value unknown parameters}. Furthermore, in the case of few-shot TL, we consider the laboratory data to train the multi-head NN to obtain the body of the NN.

\par We consider four sets of input data (control vector and speed) to generate lab and field data and generate data for different ambient temperature and ambient pressure as shown in Table \ref{Table:Pretrained data}. The simulated data are generated with $dt=0.2$ sec (sampling frequency $5 \, Hz$) for training and testing of DNN for empirical formulae and DeepONet except training of $\eta_{tm}$. In the case of $\eta_{tm}$, these are generated at $dt=0.025$ sec. We trained the DNN for $f_{egr}(\mathcal{N}_2^P(:,\bm{\theta}))$ only with data from Case-I shown in Table \ref{Table:Pretrained data}. The data for training of multi-head DNN for few-shot TL are discussed later in the numerical results section. For further details regarding data generation, we refer readers to \cite{Nath_2023}. We consider PINN for each 60 sec time segment independently in our study. The initial conditions for all the state variables including for $x_r$ and $T_1$ are known for each time segment.

\begin{table}[!t]
\centering
    \caption{\textbf{Data considered for training and testing} of pre-trained networks, DeepONets and field data for PINN. The input data Set-I is two hours of input control vector $\{u_\delta, u_{egr}, u_{vgt}\}$ and $n_e$. Set-II is twenty-minute of input control vector $\{u_\delta, u_{egr}, u_{vgt}\}$ and $n_e$. Similarly, Set-III twenty-four minute data marginally modified $u_{egr}$ to get a larger operating range. Similarly, Set-VI is a twenty-four-minute data marginally modified $u_{vgt}$ to get a larger operating range. Case-I to Case-IV are considered for the training of the surrogate neural network for the empirical formulae ($\mathcal{N}_i^{(P)}(:,\bm{\theta}_i)$). Case-I and Case-VI are considered for the training, while Case-V is considered for the testing of these networks. The data for the field data are also considered from Case-V. Case-1 and Case-VI are considered for the training of DeepONet for EGR. Case-1 and Case-VII are considered for the training of DeepONet for VGT. DeepONet for EGR and VGT are tested using Case-V. $T$ and $P$ refer to the temperature and pressure of baseline measurement.}
    \label{Table:Pretrained data}
    \begin{tabular}{L{1.3cm}|C{2cm}C{2cm}C{1.3cm}C{1.5cm}C{4cm}}\hline
    \multirow{2}{*}{Case} & $T_{amb}$ & $p_{amb}\times10^5$  & \multirow{2}{*}{Input} & Sampling   & \multirow{2}{*}{Purpose} \\ 
    &  (K) &  (Pa) & & $dt$ & \\ \hline
    
    Case-I & $T-65$ & $P-0.100$ & Set-I & 1 sec & \multirow{4}{*}{ \begin{tabular}{L{3.4cm}}
         Training pre-trained DNN for empirical formulae
    \end{tabular}} \\
    
    Case-II & $T-65$ & $P+0.211$ & Set-I & 1 sec & \\
    
    Case-III & $T-28$ & $P-0.100$ & Set-I & 1 sec & \\
    
    Case-IV & $T+15$ & $P+0.211$ & Set-I & 1 sec & \\ \hline

    \rowcolor{blue!10!green!10}
    Case-V & $T$ & $P$ & Set-II & 0.2 sec & \begin{tabular}{L{3.4cm}}
    Testing of DNN and DeepONet / Field data for PINN 
    \end{tabular}\\ \hline

    Case-I & \multirow{2}{*}{$T-65$} & \multirow{2}{*}{$P-0.100$} & Set-I & 1 sec & \multirow{2}{*}{
    \begin{tabular}{L{3.4cm}}
    Training $\tilde{u}_{egr1}$ and $\tilde{u}_{egr2}$ 
    \end{tabular}}\\ 
    
    Case-VI &  &  & Set-III & 1 sec &  \\ \hline

    Case-I & \multirow{2}{*}{$T-65$} & \multirow{2}{*}{$P-0.100$} & Set-I & 1 sec & \multirow{2}{*}{
    \begin{tabular}{L{3.4cm}}
    Training $\tilde{u}_{vgt}$
    \end{tabular}}\\ 
    Case-VII &  &  & Set-VI & 1 sec &  \\ \hline
    \end{tabular}

\end{table}

\section{Results: Multi-stage Transfer Learning}
\label{Subsection:Results for PINN inverse problem}
In this section, we present results for the inverse problem solved using the hybrid multi-stage TL approach outlined in section \ref{Subsection:Transfer learning}. We consider two scenarios: i) training with clean data, and ii) training with noisy data. Since we use simulated data for model training, we added Gaussian noise to these simulated outputs to mimic real-world conditions. Based on expected noise from field sensors, we added Gaussian noise values of 3\%, 3\%, 1\%, 5\% and 3\%  to simulated outputs for $p_{im}$, $p_{em}$, $\omega_{t}$, $W_{egr}$ and $T_{em}$ respectively. The input signal required for the Simulink model viz. control input $\bm{u} = \{u_\delta, u_{egr}, u_{vgt}\}$ and engine speed $n_e$, are obtained from field measurements on an engine test bed. As discussed in section \ref{Subsection:Transfer learning}, we train the hybrid PINN model on individual one minute data segments. In the first segment, the complete hybrid PINN model is trained. Complete training of this hybrid PINN model is computationally expensive since it requires learning all network parameters and requires more training epochs. To reduce computational time for subsequent time segments, we transfer the network weights and biases to the multi-stage TL model.

\par We approximate $p_{im}$, $p_{em}$, $\omega_t$, $T_1$, and $x_r$ using neural networks (as shown in Table \ref{Table:FNN for PINN}) and incorporate mask function and scaling for the unknown parameters, as shown in Table \ref{Table:true value unknown parameters}. We note that output transformation, mask function, and scaling are important to ensure the network outputs and unknown parameters lie in $\mathbb{R^+}$. Additionally, we incorporate self-adaptive weights for the data loss term of $\bm{\lambda}_{p_{em}}$, $\bm{\lambda}_{p_{im}}$, $\bm{\lambda}_{\omega_{t}}$ and $\bm{\lambda}_{W_{egr}}$, initialized from a Gaussian random variable with mean $\dfrac{\alpha X_{data}}{\max(X_{data})}$, $X = \{p_{em}, p_{im}, \omega_{t}, W_{egr}\}$ and standard deviation $\sigma=0.1$, where $\alpha$ is a heuristically determined weight multiplier ($10^3$, $10^3$, $10^3$, and $2\times 10^3$ for $\bm{\lambda}_{p_{em}}$, $\bm{\lambda}_{p_{im}}$, $\bm{\lambda}_{\omega_{t}}$ and $\bm{\lambda}_{W_{egr}}$ respectively). The self-adaptive weight $\lambda_{T_{em}}$ and $\lambda_{T_{1}}$ are initialized from a Gaussian random variable with mean $1500$ and standard deviation 0.1. A softplus masking function is applied to the self-adaptive weights to ensure that they remain positive. The unknown parameters for the first 0-60 sec are initialized from a Gaussian random variable with standard deviation $\sigma=0.2$ and mean near the true values of the parameters. For calculating PINN loss defined in Eq. \eqref{Eq:Loss total loss}, 301 residual points are used for individual PINN loss terms. The parameters of the neural network, unknown engine parameters, and self-adaptive weights are optimized using the Adam optimizer \cite{Kingma_2014adam} with piece-wise constant learning rates for each stage with multiple optimizers. Stage I extends from 0-20K epochs, Stage II between 20-30K epochs, and Stage III from 30-35K epochs (ref Fig. \ref{Figure:Transfer learning schematic} for further details). To estimate the epistemic uncertainty, we train the hybrid PINN model separately for each data segment to generate an ensemble of 30 runs. This uncertainty arises from the random initialization of self-adaptive weights $\mathbf{\lambda}$, unknown parameter values $\mathbf{\Lambda}$, and dropouts included in the surrogate networks for estimating empirical formulae discussed in section \ref{Subsection:Neural network surrogates for empirical formulae}. Furthermore, we transfer the network parameters from different source models (trained on segment $0\sim 60$ sec) which also adds to the overall uncertainty. In the following sections, we present results for clean and noisy data conditions. A summary of the mean and standard deviation estimates for the unknown parameters under different test conditions is presented in Table \ref{Table: Predicted unknown, transfer_learning}.

\begin{table}[!t] 
\centering
\caption{\textbf{Predicted unknown parameters (multi-stage Transfer Learning)}. The mean and standard deviation of the predicted unknown parameters with multi-stage transfer learning using clean data and Gaussian noisy data. For clean data, we show results obtained by using 151 and 301 data points while for noisy data, we show results obtained by using 301 data points. With noisy data, more data points are required for accurate parameter and state estimations.}
\label{Table: Predicted unknown, transfer_learning}
\begin{tabular}{c|c|c|c|c} \hline
\rowcolor{lightgray}
\textsc{DURATION (min)} & $A_{egrmax}$ & $A_{vgtmax}$ & $\eta_{sc}$ & $h_{tot}$  \\ \hline

\multicolumn{5}{c}{\textbf{Clean data, 151 data points for each known variables (for each 60 sec.).}} \\ \hline
$0\sim1$ & $1.009 \pm 0.0017$	 & $1.022\pm 0.0003$ & $1.004\pm0.0008$ & 	$1.003\pm0.0035$   \\
$1\sim2$  & $0.922\pm 0.0487$ & 	$1.0185\pm0.0036$ & 	$0.977\pm0.0183$ & 	$0.850\pm0.0967$ \\
$2\sim3$ & 	$0.955\pm 0.0117$ & 	$0.991\pm0.0036$ & 	$1.012\pm0.0017$ & $1.013\pm 0.0041$  \\
$3\sim4$ & 	$1.007\pm 0.0017$	 & $1.019\pm0.0004$	 & $1.003\pm0.0025$ & 	$0.992\pm 0.0072$ \\
$4\sim5$ & $0.944\pm0.0980$	 & $0.999\pm0.0073$	 & $1.001\pm0.0025$ & 	$0.988\pm 0.0103$  \\
\hline \hline
\multicolumn{5}{c}{\textbf{Clean data, 301 data points for each known variables (for each 60 sec.).}} \\ \hline
$0\sim1$ 	&	$1.004\pm0.0010$	&	$1.024\pm0.0003$	&	$1.007\pm0.000$	&	$1.005\pm0.0012$  \\
$1\sim2$	&	$0.971\pm0.0470$	&	$1.012\pm0.0046$	&	$1.000\pm0.0833$	&	$0.980\pm0.0561$ \\
$2\sim3$	&	$0.964\pm0.0123$	&	$0.992\pm0.0011$	&	$1.008\pm0.0025$	&	$1.007\pm0.0056$ \\
$3\sim4$	&	$1.004\pm0.0013$	&	$1.019\pm0.0003$	&	$1.003\pm0.0017$	&	$0.992\pm0.0054$ \\
$4\sim5$	&	$0.900\pm0.1033$	&	$1.006\pm0.0022$	&	$0.998\pm0.0042$	&	$0.984\pm0.0137$ \\
\hline \hline
\multicolumn{5}{c}{\textbf{Noisy data, 301 data points for each known variables (for each 60 sec.).}} \\ 
\multicolumn{5}{c}{ Noise level: $p_{em}=3\%$, $p_{im}=3\%$, $\omega_{t}=1\%$, $W_{egr}=5\%$, $T_{em}=3\%$}
\\[0.1cm] \hline

$0\sim1$ 	&	$0.997\pm0.1103$	&	$1.021\pm0.0061$	&	$0.999\pm0.0125$	&	$0.999 \pm 0.0452$  \\
$1\sim2$	&	$0.952\pm0.0693$	&	$1.011\pm0.0072$	&	$1.007\pm0.0108$	&	$1.021\pm0.0675$ \\
$2\sim3$	&	$0.803\pm0.1047$	&	$0.974\pm0.0148$	&	$0.998\pm0.0125$	&	$0.984\pm0.0734$ \\
$3\sim4$	&	$1.004\pm0.0353$	&	$1.019\pm0.0051$	&	$0.993\pm0.0117$	&	$0.978\pm0.0271$ \\
$4\sim5$ &	$0.793\pm0.1333$	&	$1.002\pm0.0103$	&	$0.997\pm0.0117$	&	$0.985\pm0.0328$ \\
\hline \hline
\rowcolor{Salmon}
\multicolumn{5}{c}{\textbf{True value (normalized)}} \\ \hline
\rowcolor{Salmon} & $1.000$ & $1.000$ & $1.000$ & $1.000$  \\ \hline
\end{tabular}
\end{table}

\subsection{Parameter and gas flow dynamics estimation}
\begin{figure}[!t]
    \centering
    \includegraphics[width=0.9\textwidth]{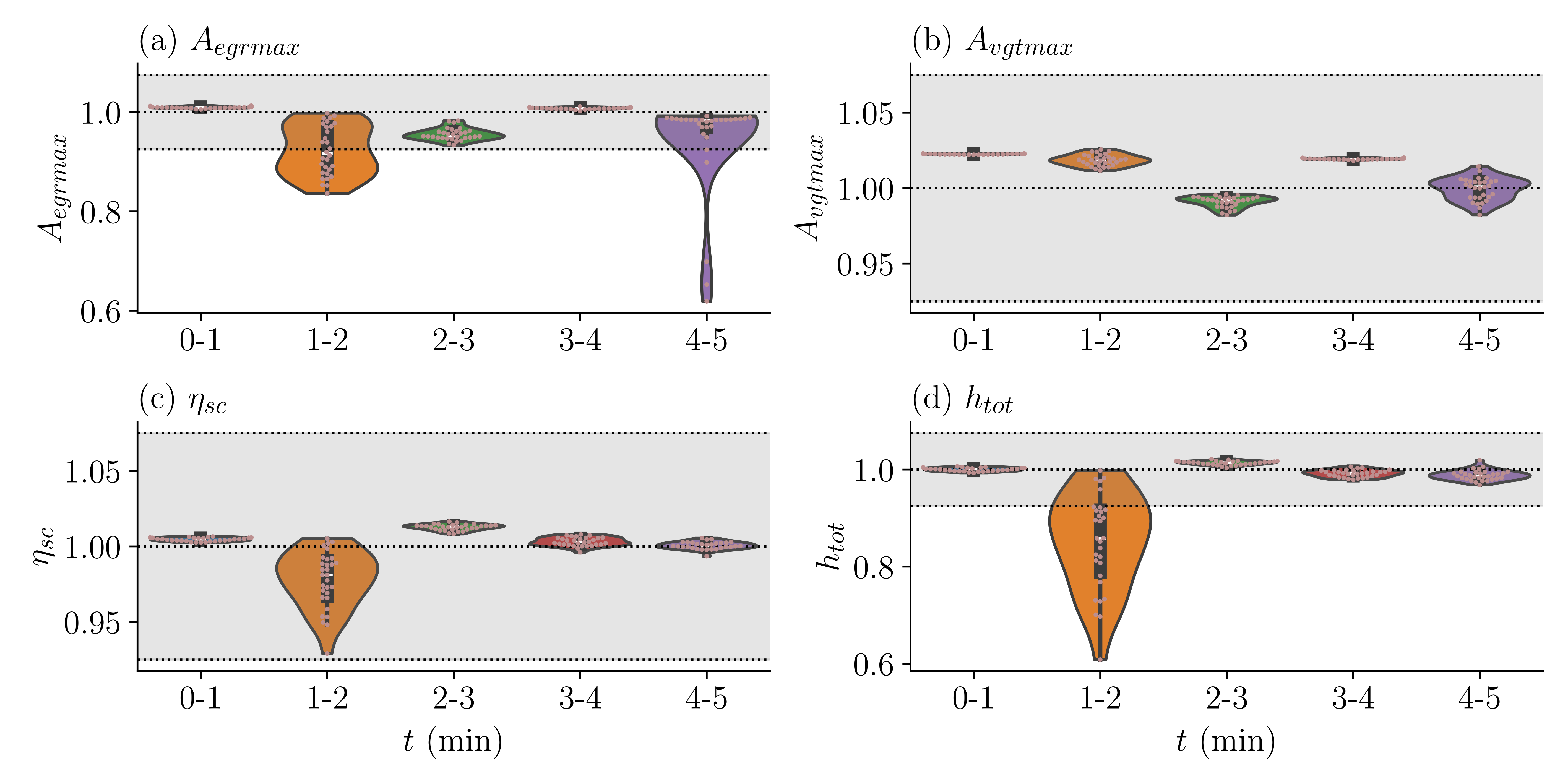}
    \caption{\textbf{Violin plot, multi-stage TL, clean data, 151 data points:} variation in unknown parameter predictions over 30 runs with clean data and 151 data points. 151 data points are considered for each of the known parameters (for each minute) under clean conditions. The violin plots are limited to the maximum and minimum values of the predicted parameters. The dots on the violin plots show a swarm plot indicating each predicted value for the 30 independent runs. The grey band shown indicates ± 7.5\% error band. The dotted line in the centre indicates the true value of the unknown parameter. The top dotted line indicates 1.075 times the true value while the bottom dotted line indicates 0.925 times the true value.} 
    \label{fig:TF_violin_clean151}
\end{figure}

We evaluate the performance of our multi-stage TL model by comparing the state and parameter estimates against known ground truth values by simulating online testing conditions. Due to confidentiality, the dynamics shown in the results are normalized on a scale 0-1 using their min-max values as:
\begin{equation}
    x_{\text{scale}} = \dfrac{x - x_\text{min}}{x_\text{max} - x_\text{min}}
    \label{Eq:Scale for results}
\end{equation}
The variability in parameter estimation is obtained by computing the expected value and variance across thirty independent runs for each segment. The estimate of unknown parameters $\mathbf{\Lambda}$ represent the values obtained at the end of 35,000 epochs. For clean data, we compared the effect of number of data points on parameter estimation by evaluating our model with 151 and 301 points. We observe that for clean data, 151 data points are sufficient for accurate estimation of parameters and gas flow states. See Fig.\ref{fig:TF_violin_clean151} showing the distribution of estimated parameters for 30 independent runs. Higher variation is observed in time segment 1-2 minute for all parameters likely due the engine entering idling condition from a dynamical state which induces bias in PINN training With 301 data points, we observe a sightly reduced variance of parameter estimation across all $\mathbf{\Lambda}$ (see Fig. \ref{fig:TF_violin_clean301}). Training with fewer data improves computational efficiency and hence, is preferred for field applications. This trade-off between requirements of computational time and accuracy needs to be considered for field application. Overall, the mean estimated values are reasonably close to the ground truth values.

\begin{figure}[H]
    \centering
    \includegraphics[width=0.9\textwidth]{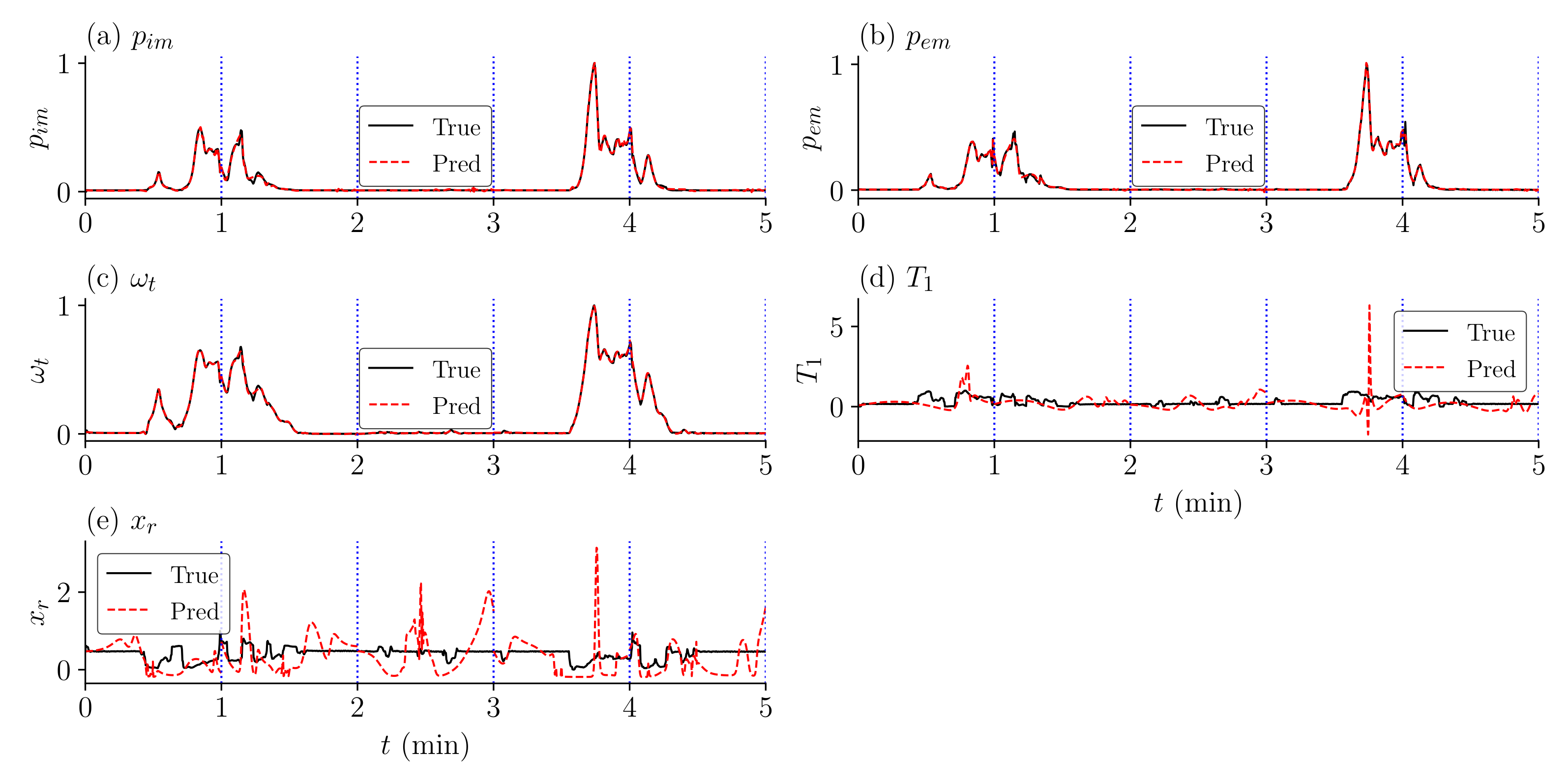}
    \caption{\textbf{Engine states, multi-stage TL, noisy data, 301 data points:} Comparison between hybrid-PINN prediction and ground truth. The model shows reasonable accuracy when predicting engine states considering the rapidly changing engine dynamics across the five minute long test signal.}
    \label{fig:TF_states_noisy}
\end{figure}

\begin{figure}[H]
    \centering
    \includegraphics[width=0.9\textwidth]{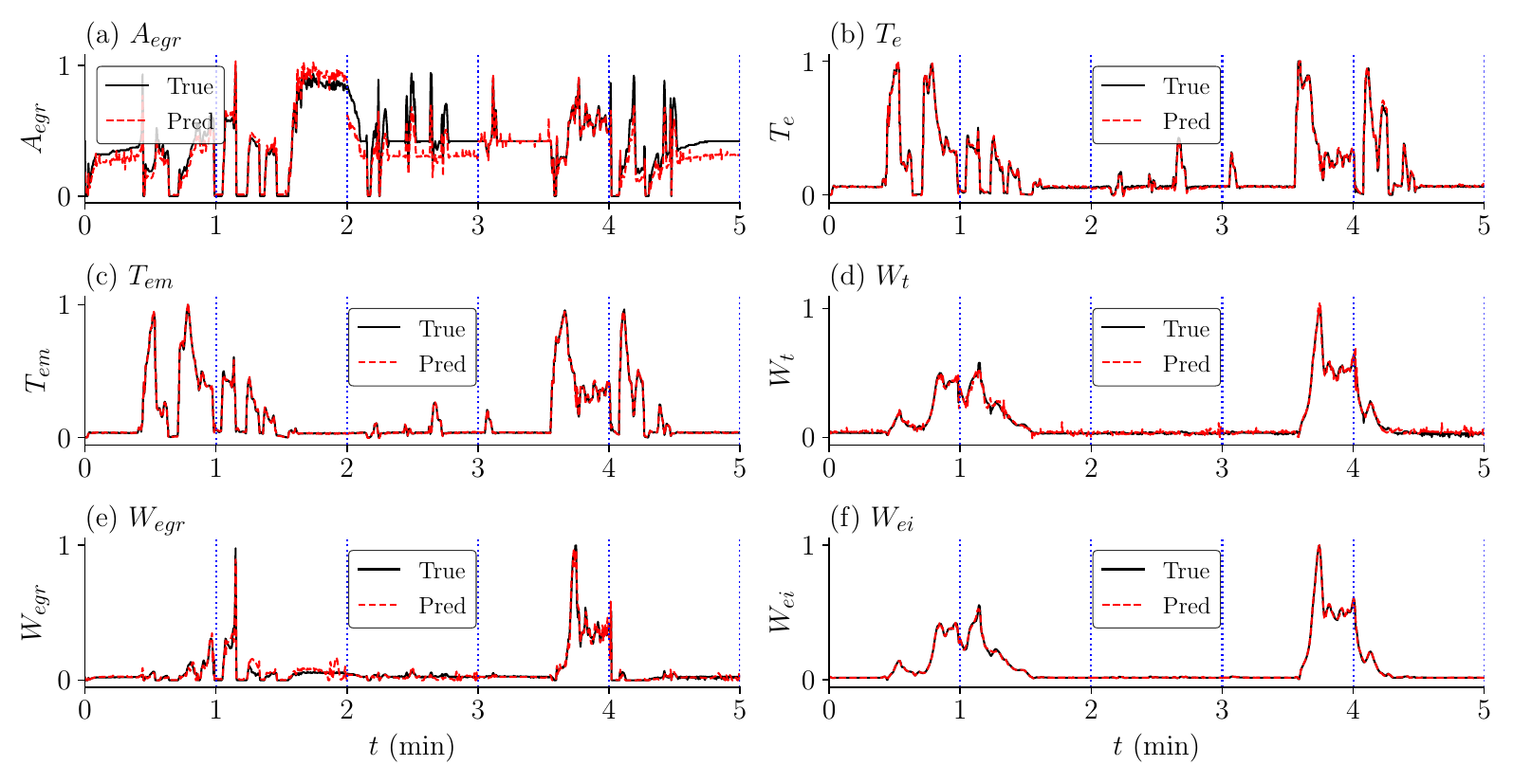}
    \caption{\textbf{Gas flow rates, multi-stage TL, noisy data, 301 data points:} Gas flow rates from multi-stage transfer learning model (data scale normalized due to confidentiality). The plot shows a representative case where 301 noisy data points are used for online training.}
    \label{fig:TF_flows_noisy}
\end{figure}

\begin{figure}[H]
    \centering
    \includegraphics[width=0.9\textwidth]{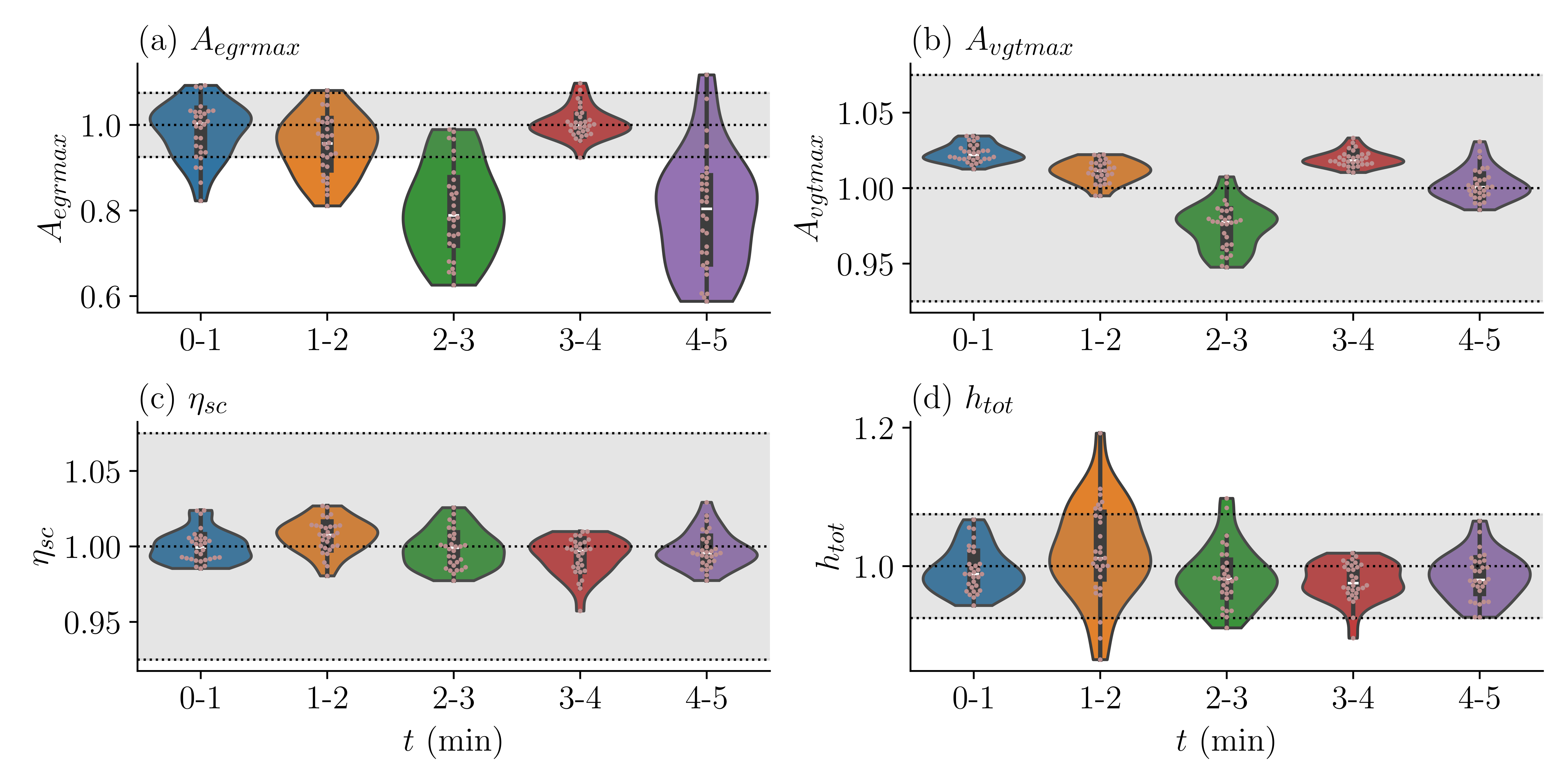}
    \caption{\textbf{Violin plot, multi-stage TL, noisy data, 301 data points:} 301 data points are considered for each of the known parameters (for each minute) under noisy conditions. The violin plots are limited to the maximum and minimum values of the predicted parameters. The results are shown after normalizing the true value of the unknown parameters.}
    \label{fig:TF_violin_NoisyS301}
\end{figure}

To simulate a practical scenario, we assess the performance of our model when trained using noisy data. During our testing, we identified the need to use more data points to enhance the prediction accuracy of the TL model. Consequently, we utilize 301 data points for training the PINN models under noisy conditions. From Fig. \ref{fig:TF_states_noisy}, we observe that predictions from PINN model show a good match with ground truth data for states $p_{im}, p_{em}$, and $\omega_t$ under noisy conditions. However, states $T_1$ and $x_r$ exhibit higher standard deviations from simulated ground truth data. This can be attributed to the relatively smaller influence of these two state variables on gas dynamics. Prediction for these two states depend solely on the gradient of the physics-based loss term which is small leading to higher variation in state estimation due to numerical error. Overall, the prediction from PINN model demonstrates good approximation to ground truth values for all states. We observe from the gas flow rate predictions shown in Fig. \ref{fig:TF_flows_noisy} that the prediction for state $A_{egr}$ from PINN model shows higher deviation from expected ground truth. Since this state estimate is directly related to the estimate of the unknown parameter $A_{egrmax}$, we expect the parameter estimate to exhibit high variability. This can be observed in Fig. \ref{fig:TF_violin_NoisyS301} which illustrates the distribution of unknown parameters $\mathbf{\Lambda}$ over the five time segments with thirty independent runs for each segment. With noise, we see higher variation across all parameter estimates which is expected due to the aleatoric uncertainty. Additional ensemble plots showing the predicted values and expected range of predictions for different QoIs are shown in Appendix section \ref{sec:additional_plots_MTL}.

 

\section{Results: Few-shot transfer learning}
\label{Section:Numerical results: Few-shot transfer learning}
\subsection{Training of multi-head neural networks}
\label{Subsection:Training of multi-head neural networks}
As discussed earlier, few-shot TL involves two Phases: (1) offline training of the neural network with multiple heads and (2) training the head and unknown parameters for the inverse problem. The training in  Phase I is done with 240 heads for each variable; each head is used for estimating a one-minute data segment. The input to the network is $0\sim 60$ sec ($dt=0.2$ sec) normalized to $[-1,1]$. The first 120 heads correspond to the state variable for Case-I dataset, and the second 120 heads correspond to the state variable for Case-IV dataset (ref Table \ref{Table:Pretrained data}). We have considered four NN similar to those in Table \ref{Table:FNN for PINN} with the same transformation and scaling. However, with different sizes and with multiple heads. The training error for each state variable and $x_r$ and $T_1$ are in Table \ref{Table:Multihead_training} (in Appendix \ref{Appendix:Additional figure for few-shot transfer learning}). We trained these networks using the Adam optimizer with an MSE loss function between the labelled and the predicted output. In the case of $N_1(t, \bm{\theta})$, as there are two output state variables, an appropriate weight is considered while summing the two losses. The weighs of the network are initialized from a Gaussian distribution with mean zero and standard deviation $\sigma = \sqrt{2/(\text{dim}_\text{in}+\text{dim}_\text{out})}$ except for the last layer. The last layer is initialized with a standard deviation of $\sigma = \sqrt{10/(\text{dim}_\text{in}+\text{dim}_\text{out})}$. This shows a better accuracy compared to an initialization with a similar standard deviation as the hidden layers. This may be because, as $\text{dim}_\text{out}$ in the last layer has more neurons, the standard deviation becomes smaller compared to the previous layers. Furthermore, to check the applicability of the method for practical application, the training in Phase-I is done with different parameters and the TL (Phase-II) is done with different parameters, as discussed in section \ref{Subsection:Basic philosophy} (ref Table \ref{Table:true value unknown parameters}). We note that in the case of few-shot transfer learning, the PINN network sizes are comparatively larger than the multi-stage network sizes as they approximate multiple heads. However, since only the last layer is trained and gradients are calculated outside the training loop, this framework is computationally more efficient.

\subsection{Parameter and gas flow dynamics estimation}
\label{Seubsection:Few shot: Parameter estimation clean data}
After training,, the next step is to perform PINN to identify the engine parameters (Phase-II). The output layer ($\bm{W}_{out}$, $b_{out}$) of the multi-head networks is replaced with one output and trained for a new control vector ($\bm{u}=\{u_\delta, u_{vgt}, u_{egr}\}$) and engine speed ($n_e$). The weights of the output layer are initialized from a random Gaussian distribution with mean zero and standard deviation $\sigma = \sqrt{2/(\text{dim}_\text{in}+\text{dim}_\text{out})}$. The weights are sorted after initialization, and it shows better and faster convergence. The biases are initialized with zero. Similar to the PINN and TL method, each of the self-adaptive weights for the data loss term of $\bm{\lambda}_{p_{em}}$, $\bm{\lambda}_{p_{im}}$, $\bm{\lambda}_{\omega_{t}}$ and $\bm{\lambda}_{W_{egr}}$ are initialized from a Gaussian random variable with mean $\dfrac{\alpha X_{data}}{\max(X_{data})}$, $X = \{p_{em}, p_{im}, \omega_{t}, W_{egr}\}$ and standard deviation $\sigma=0.1$, where $\alpha$ is a initial weight given ($10^3$, $10^3$, $10^3$ and $2\times 10^3$ for $\bm{\lambda}_{p_{em}}$, $\bm{\lambda}_{p_{im}}$, $\bm{\lambda}_{\omega_{t}}$ and $\bm{\lambda}_{W_{egr}}$ respectively). The self-adaptive weight $\lambda_{T_{em}}$ and $\lambda_{T_{1}}$ are initialized from a Gaussian random variable with mean $20$ and standard deviation 0.1. We consider a softplus mask function to the self-adaptive weights so that the weights are always positive. The unknown parameters are initialized from a Gaussian random variable with standard deviation $\sigma=0.2$ and mean near the true values of the parameter. Furthermore, as these quantities are always positive, we consider mask functions and scaling as shown in Table \ref{Table:true value unknown parameters}. Also note that the trainable parameters of the problem are the weights and bias of the last layer of the neural networks, the unknown parameters of the engine and the self-adaptive weights. These are optimized using eight different Adam optimizers: one for the weights and the biases of the neural networks, one for the unknown parameters, and one for each of the six self-adaptive weights. The loss function is given by the Eq \eqref{Eq:Loss total loss} as discussed in section \ref{Section:Methodology}. We would also like to mention that we only optimized the self-adaptive weights up to $20\times 10^3$ epoch. After that, we fixed the self-adaptive weights with the value at $20\times 10^3$ epochs. We consider total $30\times 10^3$ epochs.  Furthermore, as discussed earlier, in order to reduce the computational cost, we perform the derivative of the output with respect to the input outside the training loop as follows,
\begin{subequations}
\begin{equation}
    y = S_p\left(\bm{h}(t) \bm{W}_{out} +b_{out}\right)
\end{equation}
\begin{equation}
\begin{split}
\dfrac{d y }{dt} = & \dfrac{1}{1+\exp[\bm{h}(t) \bm{W}_{out} +b_{out}]} \times \\ &\exp\left[\bm{h}(t) \bm{W}_{out} +b_{out}\right] \dfrac{d\bm{h}(t)}{dt} \bm{W}_{out}    
\end{split}
\label{Eq:mult-head derivative softplus}
\end{equation}
\end{subequations}
where $S_p$ is softplus function (refer Table \ref{Table:Multihead_training}), $\bm{h}(t)$ is the output of the last hidden layer, $\bm{W}_{out}$ and $b_{out}$ are the weights and bias of the output layer,  $\dfrac{d\bm{h}(t)}{dt}$ is the derivative of the output of the last hidden layer with respect to input $t$. From Eq. \eqref{Eq:mult-head derivative softplus}, the derivative term is equal to $\dfrac{d\bm{h}(t)}{dt}$, and as the hidden layers are not trained, this remains fixed and can be evaluated outside the network training loop. The size of $\dfrac{d\bm{h}(t)}{dt}$ is a matrix with size $n_t\times n_h$, where $n_t$ is the number of input points in time, and $n_h$ is the number of neurons in the last hidden layer. 

\par We consider two different scenarios with clean data without additional noise in the field data (simulated). However, field data may be noisy in practice due to the inherent instrument noise or other reasons. Thus, to simulate a more practical scenario and the applicability of the proposed method, we consider added noise and perform PINN. We added a Gaussian noise of 3\%, 3\%, 1\%, 5\% and 3\% added to the simulated data of $p_{em}$, $p_{im}$, $\omega_{t}$, $W_{egr}$ and $T_{em}$ respectively. We study five minutes of engine operation and sub-divide into five independent problems, each studied independently. For each one minute, we consider 301 residual points with equal $dt=0.2$ sec. For the first case of clean data, we consider $151$ data points ($dt = 0.4$ sec.) for the data loss of $p_{em}$, $p_{em}$, $\omega_t$, $W_{egr}$ and $T_{em}$, and for the second case of clean data we consider $301$ data points. In the case of noisy data, we consider 301 data points. The size of the self adaptive weights for $\lambda_{p_{em}}$, $\lambda_{p_{im}}$, $\lambda_{\omega_{t}}$ and $\lambda_{W_{egr}}$ is equal to the number of data point considered, and the size self-adaptive weight for $\lambda_{T_{em}}$ and $\lambda_{T_{1}}$ is $1\times1$. Before discussing further, we would like to mention that, there are three different types of uncertainties: (i) uncertainty associated with the initialization of the last layer of the neural network and other parameters (e.g. unknown parameters, self-adaptive weights), (ii) uncertainty in the pre-trained network due to the dropout considered, (iii) uncertainty due to the added noise in the field variables (in case of noisy data). In order to check the effect of initialization and randomness on the accuracy, we consider $30$ independent runs for each one-minute data segment.

\par
\begin{figure}[H]
    \centering
    \includegraphics[width=0.9\textwidth]{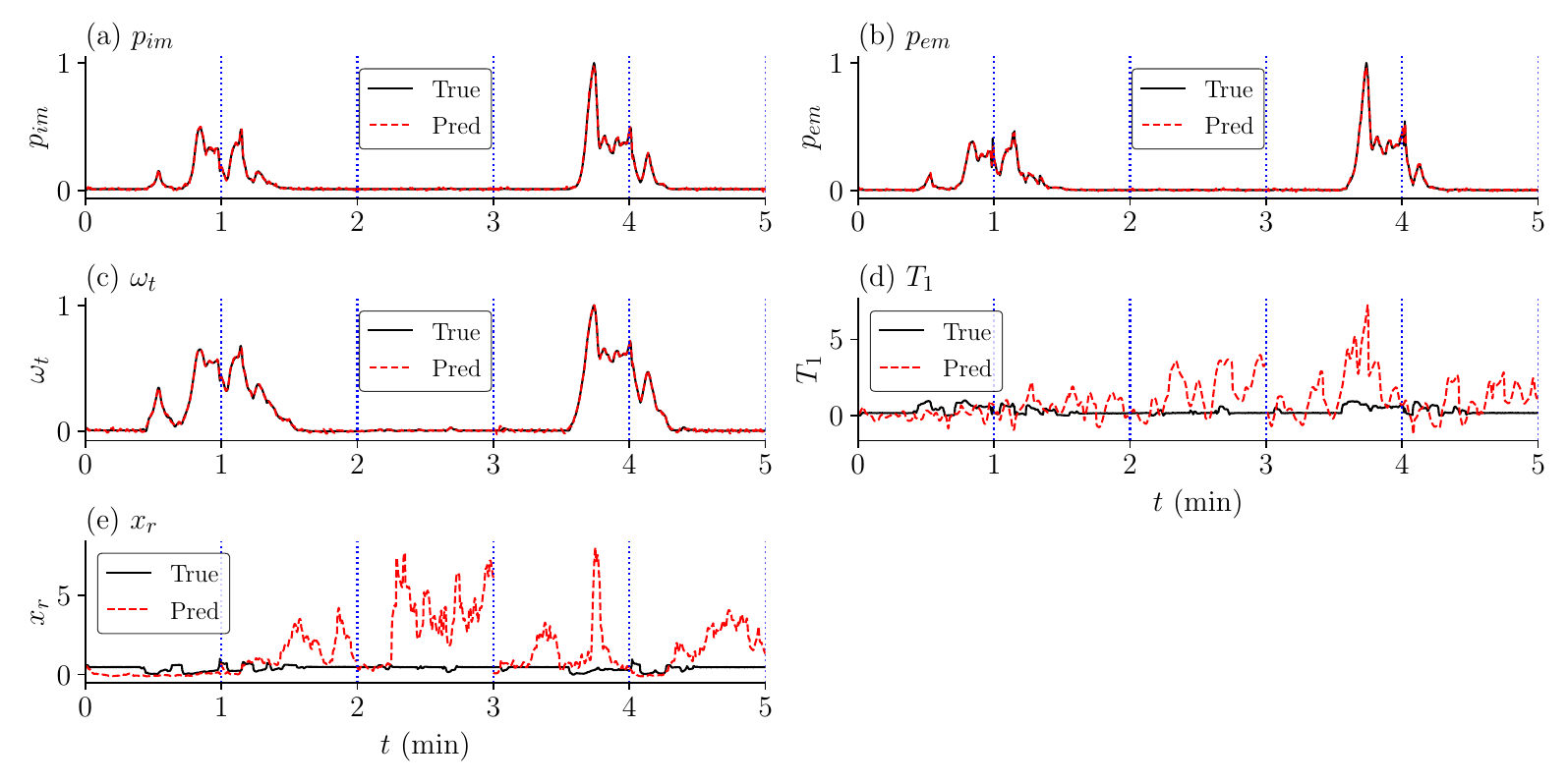}
    \caption{\textbf{Predicted states, Few-shot learning, Noisy data, 301 data points}. One of the predicted realization of the state variables and $x_r$ and $T_1$, for the case where noisy data and 301 data points are considered for each known variable (for each one minute duration). Each 5 minute are trained independently and combined to get the total 5 minute results. The vertical dotted blue lines separate each result. The results are normalized using Eq. \eqref{Eq:Scale for results} with maximum and minimum true values calculated over the 5 minute.}
    \label{Figure: States: Few shot noisy data 301 points}
\end{figure}

\begin{figure}[H]
    \centering
    \includegraphics[width=0.9\textwidth]{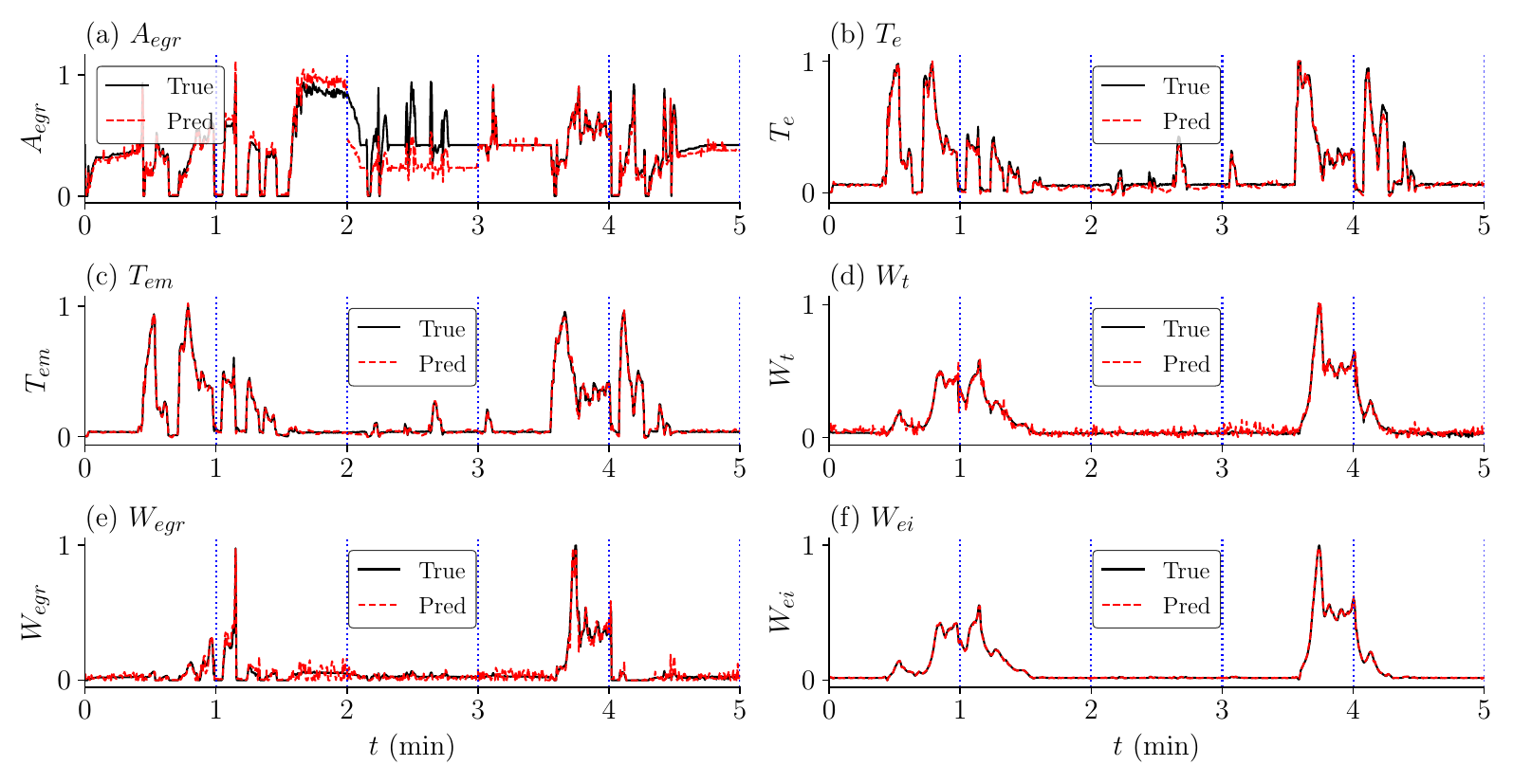}
    \caption{\textbf{Intermediate variables, Few-shot learning, Noisy data, 301 data points}. One of the predicted realizations of intermediate variables of the diesel engine for the case where noisy data and 301 data points are considered for each known variable (for each one-minute duration). The variables $A_{egr}$, $T_{e}$, $T_{em}$ and $W_t$ are the variables which are directly dependent on the unknown parameters $A_{egrmax}$, $\eta_{sc}$, $h_{tot}$ and $A_{vgtmax}$. The variable $T_e$ also depends on the variable $T_1$ and $x_r$. }
    \label{Figure: Other variable: Few shot Noisy data 301 points}
\end{figure}

\begin{figure}[H]
    \centering
    \includegraphics[width=0.9\textwidth]{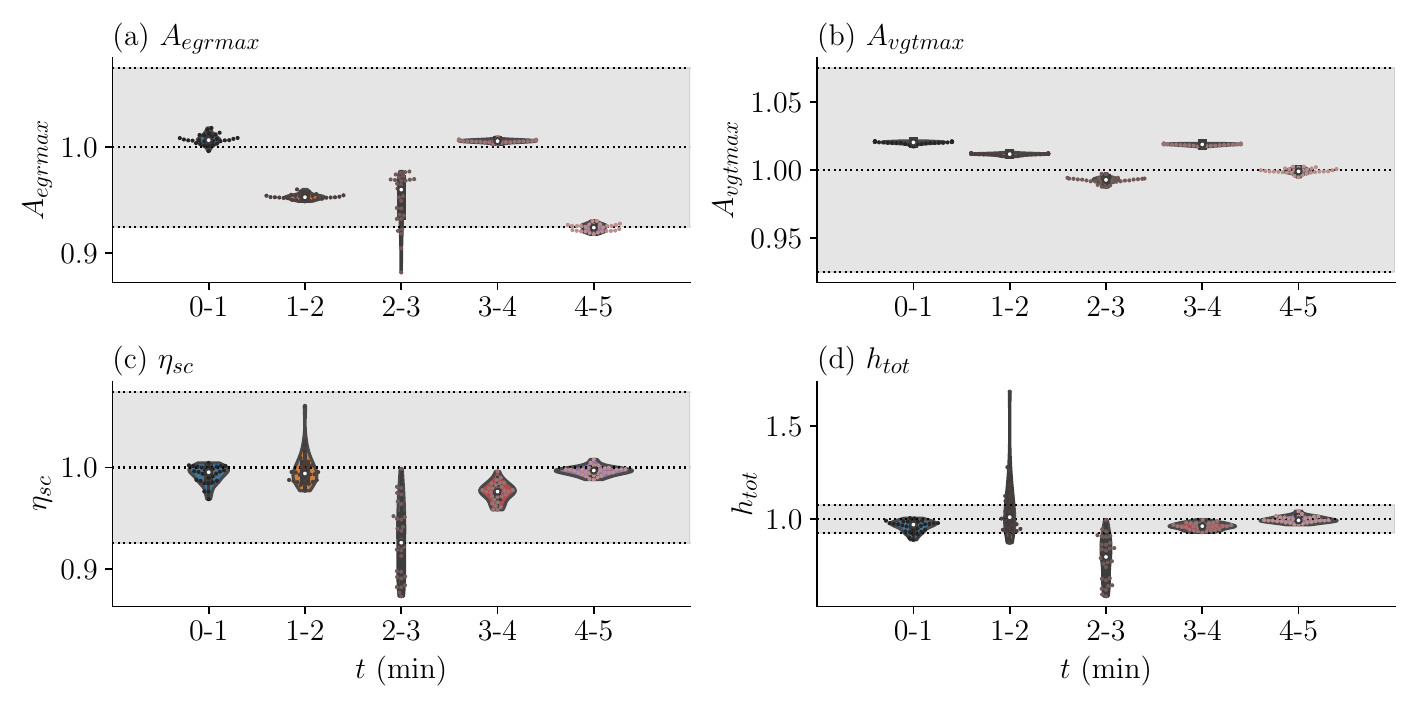}
    \caption{\textbf{Violin plot, few-shot learning, clean data, 151 data points:} of the predicted value of the unknown parameters. 151 data points are considered for each of the known parameters (for each one minute segment) and no noise is considered in this case. Similar to the previous violin plots, these results are shown in a normalized plot with the true value of the unknown parameters to be one.}
    \label{Figure:Few shot:Violin plot 151 clean data}
\end{figure}

\begin{figure}[H]
    \centering
    \includegraphics[width=0.9\textwidth]{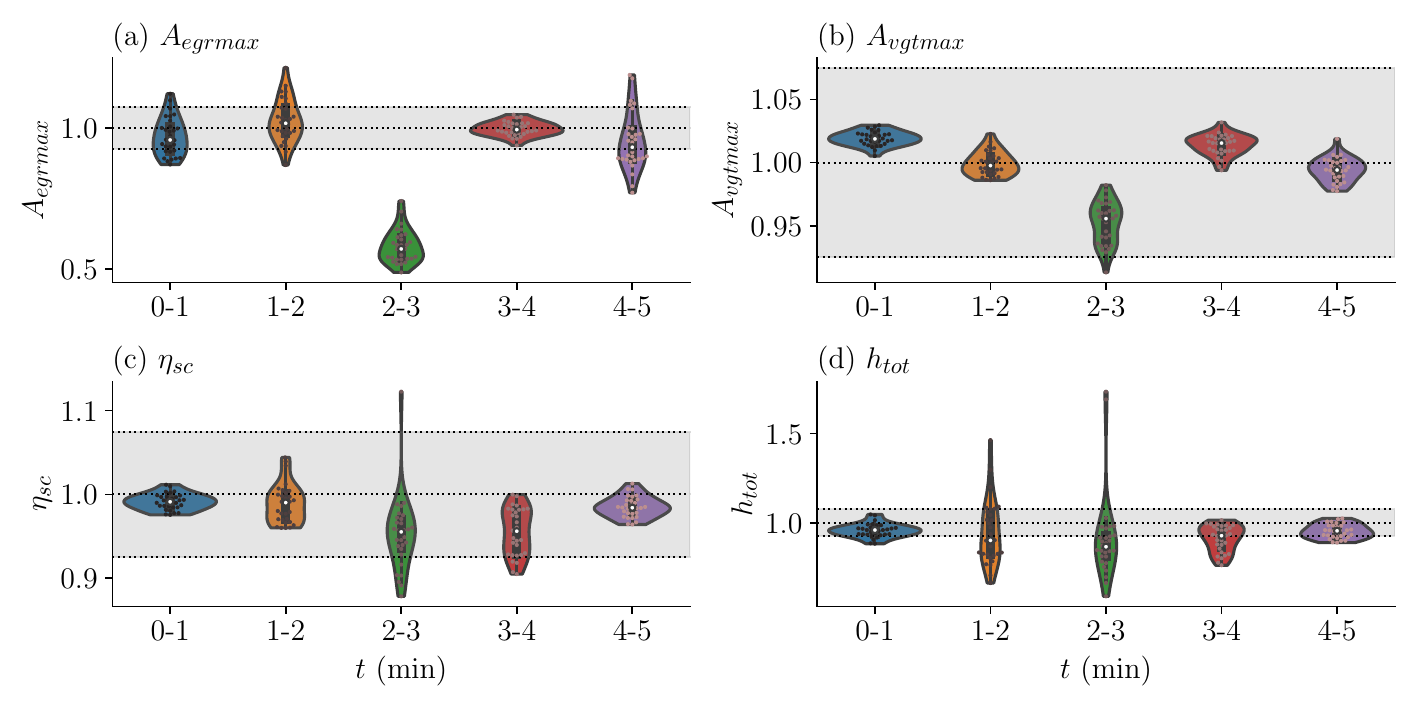}
    \caption{Violin plot of the predicted value of the unknown parameters, when predicted using few-shot TL with 301 noisy data points considered for each of the known variables (for each 1 minute). Similar to the previous violin plots, these results are shown in a normalized plot with the true value of the unknown parameters to be one.}
    \label{Figure:Few shot:Violin plot 301 noisy data}
\end{figure}

 We study the dynamics of the state variables when we add noise to the field variables. The dynamics of one of the realizations of the predicted state variables are shown in Fig. \ref{Figure: States: Few shot noisy data 301 points}. It can be observed that even with noisy data, the predicted state variables show a good agreement with the true value except $x_r$ and $T_1$. We note that even though predictions for $x_r$ and $T_1$ are not as good, the variable $T_e$ shows good accuracy ($T_e$ depends on $x_r$ and $T_1$). We also study the intermediate variables, which are shown in Fig. \ref{Figure: Other variable: Few shot Noisy data 301 points}. It is observed agreement in the predicted results except in $A_{egr}$ in 2$\sim$3 time zone. However, the dynamic pattern matches even in this time zone. This is because in the time zone, the predicted $A_{egrmax}$ deviates from the true value for this particular realization. This is expected as engine idling conditions are observed in this time zone. The little noise observed in the predicted results is expected as we are considering noisy data. The mean and standard deviation of the predicted unknown parameters for the three data cases are shown in Table \ref{Table:Param estimates, few shot learning}. The violin plots of the predicted unknown parameters for clean data of 151 point case and noisy data (301 data point) case are shown in Figs. \ref{Figure:Few shot:Violin plot 151 clean data} and \ref{Figure:Few shot:Violin plot 301 noisy data} respectively. It can be observed that predicted parameters show a good accuracy except during the 2-3 minutes duration. We observed that few-show TL show better accuracy compared to multi-stage TL. Additional results for the dynamics of the variable and ensemble study are shown in Appendix \ref{Appendix:Additional figure for few-shot transfer learning}

\begin{table}[!h]
\centering
\caption{\textbf{Predicted unknown parameters, Few-shot learning:} The mean and standard deviation of the predicted unknown parameters when predicted using Few-shot transfer learning with clean data and Gaussian noisy data. The first set of results are predicted mean $\pm$ standard deviation when 151 data points for the known variables (for each 60 sec segment) with clean data, the second set of results are when 301 data points for the known variables (for each 60 sec). The third set of results for noisy data with 301 data points for each known variables (for each 60 sec segment).}
\label{Table:Param estimates, few shot learning}
\begin{tabular}{c|c|c|c|c} \hline
\rowcolor{lightgray}
\textsc{Duration (min)} & $A_{egrmax}$ & $A_{vgtmax}$ & $\eta_{sc}$ & $h_{tot}$  \\ \hline

\multicolumn{5}{c}{\textbf{Clean data, 151 data points for each known variables (for each 60 sec.).}} \\ \hline
 $0\sim1$ 	&	$1.0073  \pm   0.0046$	&	$    1.0201 \pm   0.0008 $	&	$    0.9929 \pm   0.0086  $	&	$    0.9592 \pm   0.0296 $	\\
$1\sim2$	&	$ 0.9531 \pm   0.0026$	&	$    1.0115 \pm   0.0007 $	&	$    0.9980 \pm   0.0162 $	&	$    1.0559 \pm   0.1541 $		\\
$2\sim3$	&	$ 0.9512 \pm   0.0238$	&	$    0.9925 \pm   0.0022 $	&	$    0.9281 \pm   0.0357 $	&	$    0.7854 \pm   0.1144 $		\\[0.1cm] 
$3\sim4$ 	&	$1.0057 \pm   0.0013 $	&	$    1.0186 \pm   0.0006 $	&	$    0.9755 \pm   0.0091 $	&	$    0.9589 \pm   0.0173 $		\\
$4\sim5$	&	$0.9238  \pm  0.0032$	&	$    0.9993 \pm   0.0018 $	&	$    0.9965 \pm   0.0043 $	&	$    0.9951 \pm   0.0163 $		\\
 \hline \hline

\multicolumn{5}{c}{\textbf{Clean data, 301 data points for each known variables (for each 60 sec.).}} \\ \hline

$0\sim1$ 	&	$    1.0038  \pm  0.0044 $	&	$    1.0198 \pm   0.0007 $	&	$    0.9922 \pm   0.0056 $	&	$    0.9599 \pm   0.0207 $	\\
$1\sim2$	&	$    1.0312  \pm  0.0042 $	&	$    1.0043 \pm   0.0008 $	&	$    0.9958 \pm   0.0111 $	&	$    1.0084 \pm   0.0883 $		\\
$2\sim3$	&	$    0.9551  \pm  0.0263 $	&	$    0.9922 \pm   0.0014 $	&	$    0.9385 \pm   0.0315 $	&	$    0.8205 \pm   0.1001 $		\\
$3\sim4$	&	$    1.0057  \pm  0.0013 $	&	$    1.0190 \pm   0.0006 $	&	$    0.9688 \pm   0.0098 $	&	$    0.9532 \pm   0.0162 $		\\
$4\sim5$	&	$    0.9385  \pm  0.0041 $	&	$    1.0024 \pm   0.0006 $	&	$    0.9893 \pm   0.0047 $	&	$    0.9646 \pm   0.0156 $		\\
\hline \hline

\multicolumn{5}{c}{\textbf{Noisy data, 301 data points for each known variables (for each 60 sec.).}} \\ 
\multicolumn{5}{c}{ Noise level: $p_{em}=3\%$, $p_{im}=3\%$, $\omega_{t}=1\%$, $W_{egr}=5\%$, $T_{em}=3\%$}
\\[0.1cm] \hline
$0\sim1$ 	&	$    0.9699 \pm   0.0653 $	&	$    1.0190 \pm   0.0056 $	&	$    0.9913 \pm   0.0094 $	&	$    0.9568 \pm   0.0379 $ \\
$1\sim2$	&	$    1.0255 \pm   0.0748 $	&	$    0.9992 \pm   0.0092 $	&	$    0.9888 \pm   0.0230 $	&	$    0.9445 \pm   0.1749 $		\\
$2\sim3$	&	$    0.5800 \pm   0.0568 $	&	$    0.9520 \pm   0.0165 $	&	$    0.9524 \pm   0.0429 $	&	$    0.9087 \pm   0.2365 $		\\
$3\sim4$	&	$    0.9972 \pm   0.0246 $	&	$    1.0139 \pm   0.0081 $	&	$    0.9554 \pm   0.0276 $	&	$    0.9114 \pm   0.0721 $		\\ 
$4\sim5$	&	$    0.9525 \pm   0.0998 $	&	$    0.9942 \pm   0.0093 $	&	$    0.9849 \pm   0.0122 $	&	$    0.9519 \pm   0.0390 $		\\	\hline \hline
\rowcolor{Salmon}
\multicolumn{5}{c}{\textbf{True value (normalized)}} \\ \hline
\rowcolor{Salmon} & $1.000$ & $1.000$ & $1.000$ & $1.000$  \\ \hline

\end{tabular}
\end{table}


 \section{Comparison between Multi-stage transfer learning and Few-shot transfer learning}
\label{Section:Computational_cost}
We have presented two approaches for efficiently identifying unknown parameters of a diesel engine using a hybrid PINNs based approach. In this section, we compare the two methods based on empirical evidence from our numerical experiments. We note that both methods yield comparable results in terms of the mean values of unknown parameters. The multi-stage TL approach shows a higher variation in parameter estimation in general as compared to Few-shot TL. This is attributed to the fact that the few-shot TL network was trained on a much larger dataset that captures a wide variety of engine operating conditions. Consequently, generalization is better with the few-shot TL approach as compared to the multi-stage TL approach. However, statistical analyses of the parameter estimates did not yield conclusive evidence in support of few-shot approach in predicting target parameter values across all time segments. For instance, analysing the clean data for parameter $h_{tot}$, we observe a non-significant p-value at 95\%  confidence interval for the one-sample t-test between 240-300s which indicates enough evidence does not exist to reject the null $H_0 : \mu = 1.0$. Hence, for $h_{tot}$, one may conclude that the few-shot approach is more reliable in certain sections of the data. But the few-shot method shows departure from the target mean value of 1.0 when estimating parameter $A_{egrmax}$ in the same time segment.

For noisy data, we observe a much wider spread in the parameter estimation from the multi-stage TL approach as compared to the few-shot TL. This could be indicative of the need to retrain more layers of the multi-stage TL network instead of the last output layer to improve system dynamics approximation. For a particular time segment of $2-3$ minute, we observe that the estimation for $A_{egrmax}$ from the few-shot TL paradigm is less accurate as compared to the multi-stage TL approach. This might be because the flow through the EGR ($W_{egr}$) is very small (negligible) compared to the other time zone. Additionally, we do not observe a significant time difference in training the network with 151 or 301 data points. Any variations in computation time are transients due to network latency and the existing state of the computing node. Table \ref{tab:trnflear_vs_fewshot} shows the computational time comparison between the baseline PINN model (hybrid PINN+DeepONet), multi-stage TL, and few-shot TL paradigms. We observe that the proposed hybrid multi-stage TL strategy achieves nearly 50\% reduction in computational time while the incorporation of few-shot learning approach yields an additional 25\% improvement over the baseline PINN framework. Note that the multi-stage TL strategy still requires the full PINN framework to be trained for the initial 0-60s but learning across the remaining time segments is significantly faster. The few-shot paradigm, on the other hand requires additional training of the multi-head network but during evaluation phase, the gradient calculation for physics loss estimation is simplified leading to better computational efficiency as compared to the multi-stage TL approach.

\begin{table}[!ht]
\caption{\textbf{Computational time comparison, multi-stage vs few-shot TL}. Mean ($\mu$) and standard deviation ($\sigma$) of computational time are calculated from 30 independent runs for each time segment. While the hybrid multi-stage TL approach reduces computational time by approximately 50\%, the few-shot learning paradigm further improves the computational time by an additional 25\% in comparison to the baseline PINN model. Random variation in computational time results from system level variations and latency that exist in high-performance cluster systems which were used for this study. All computations were performed with 32-bit data precision on one CPU core with Tensorflow version 1.}
\label{tab:trnflear_vs_fewshot}
\centering
\begin{tabular}{C{3.cm}C{2cm}C{2cm}C{2.75cm}C{2.75cm}} 
\hline
\multirow{3}{*}{\textbf{Method}} & \multirow{3}{*}{\textbf{\begin{tabular}{C{1.5cm}}Training epochs \end{tabular}}} & \multirow{3}{*}{\textbf{\begin{tabular}{C{1.5cm}} Training segment\end{tabular}}} & \multicolumn{2}{c}{\textbf{Time in sec ($\mu$, $\sigma$)}} \\ \cline{4-5}
& & & \textbf{Clean data} & \textbf{Noisy data} \\ 
& & & \textbf{151 data points} & \textbf{301 data points} \\ 
\hline\hline
Baseline PINN (hybrid) & 80,000 & 0-60s & (487.12, 54.35) & (585.99, 79.87) \\ \hline
\multirow{4}{*}{\begin{tabular}{C{2.5cm}}Multi-stage Transfer learning\end{tabular}} & \multirow{4}{*}{35,000} & 60-120s & (245.50, 25.82) & (235.08, 42.45) \\
 &  & 120-180s & (257.16, 18.49) & (222.45, 51.76) \\
 &  & 180-240s & (219.47, 34.63) & (242.05, 31.99) \\
 &  & 240-300s & (237.10, 26.26) & (219.97, 45.19) \\
 \hline \hline
\multirow{4}{*}{\begin{tabular}{C{2.5cm}}Few-shot Transfer learning\end{tabular}} & \multirow{4}{*}{30,000} & 0-60s & (124.35, 11.00) & (140.28, 16.95) \\
 &  & 60-120s & (129.21, 18.20) & (137.13, 11.72) \\
 &  & 120-180s & (128.48, 14.63) & (133.41, 16.87) \\
 &  & 180-240s & (122.82, 12.28) & (141.81, 13.31) \\
 &  & 240-300s & (125.71, 11.74) & (135.28, 19.45) \\ \hline
\end{tabular}
\end{table}

\begin{figure}[H]
\centering
\centering
\includegraphics[width=1.0\textwidth]{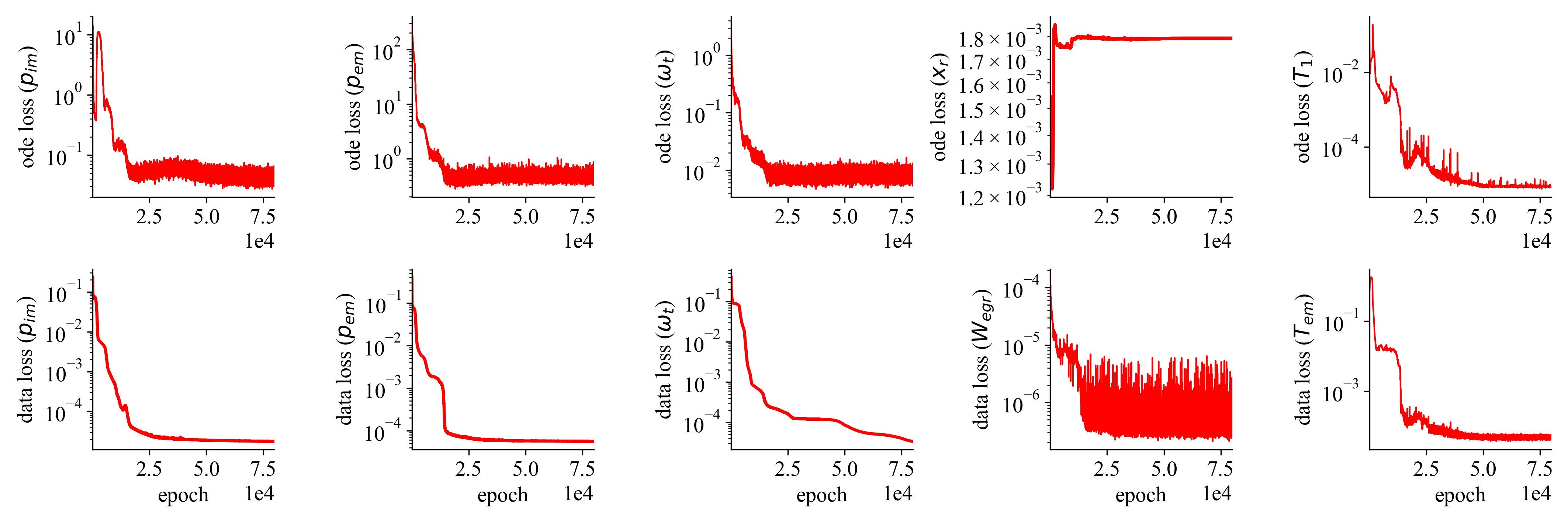}
(a) Baseline PINN
%
%
\includegraphics[width=1.0\textwidth]{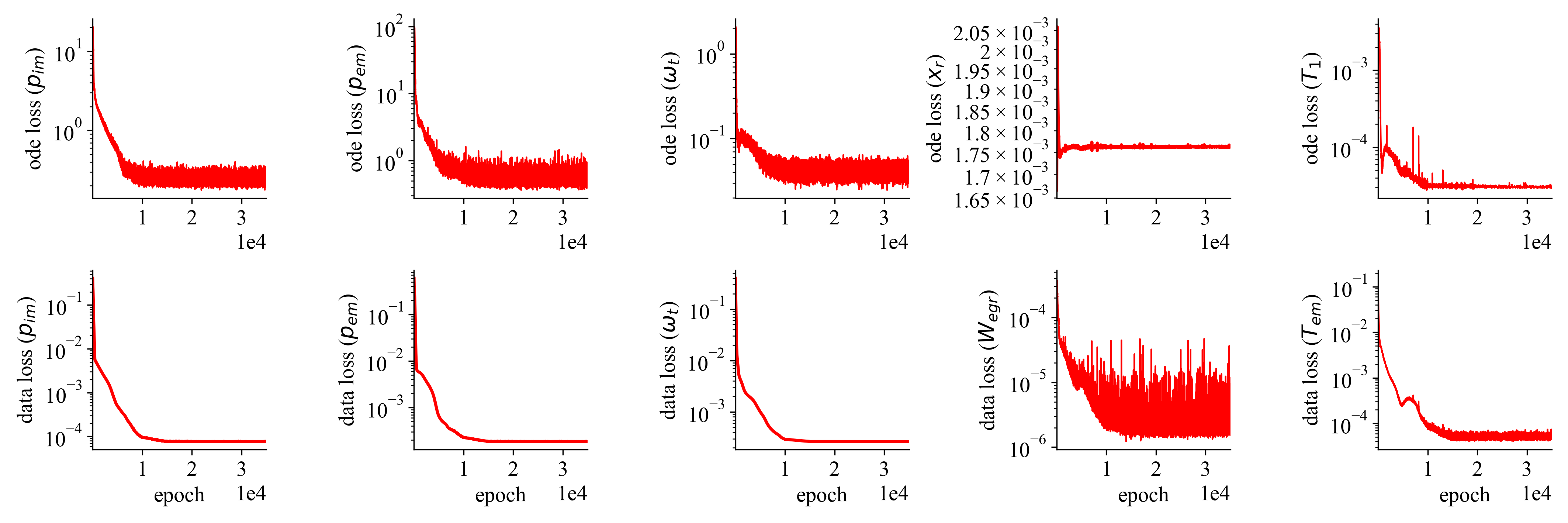}
(b) Multi-stage transfer learning
%
%
\includegraphics[width=1.0\textwidth]{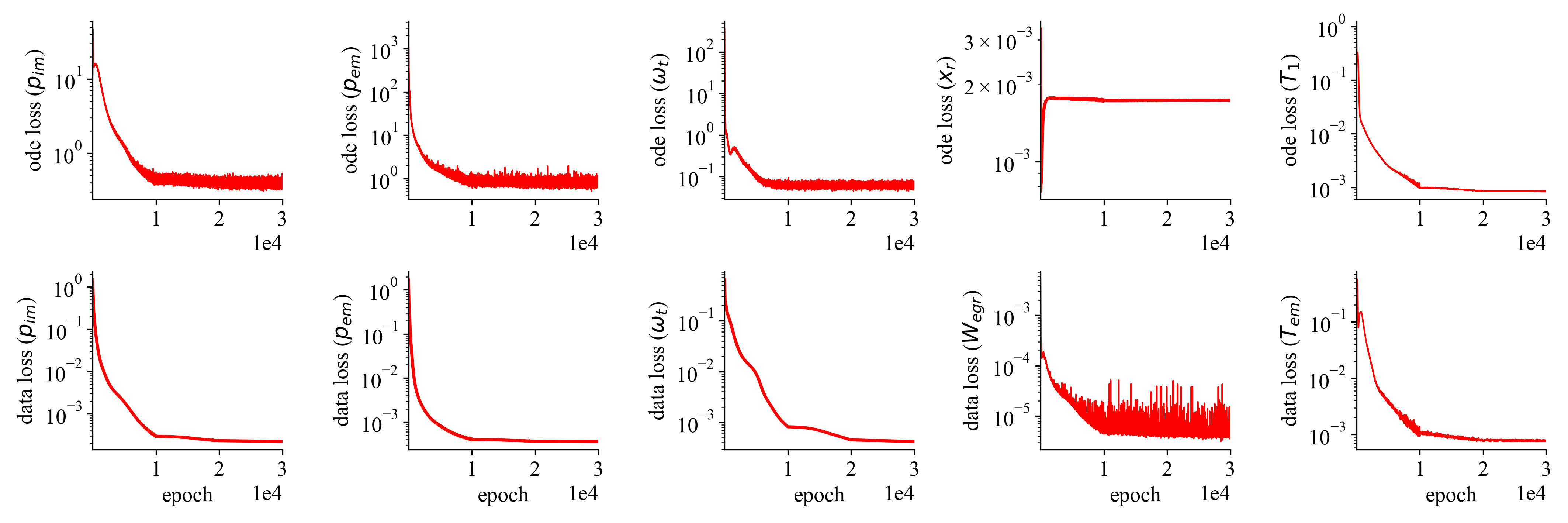}
(c) Few shot transfer learning
\caption{\textbf{Convergence of physics loss and data loss} vs epoch for different method: (a) Baseline PINN (hybrid PINN) model, (b) Multi-stage transfer learning model, and (c) Few-shot transfer learning model when 151 clean data is considered. The losses plotted are without the self-adaptive weights. Losses for both the transfer learning methods converge earlier than the baseline PINN. Initial condition losses are not shown. }
\label{fig:all_losses}
\end{figure}

\begin{figure}[H]
\centering
\includegraphics[width=1\textwidth,height=3.5cm]{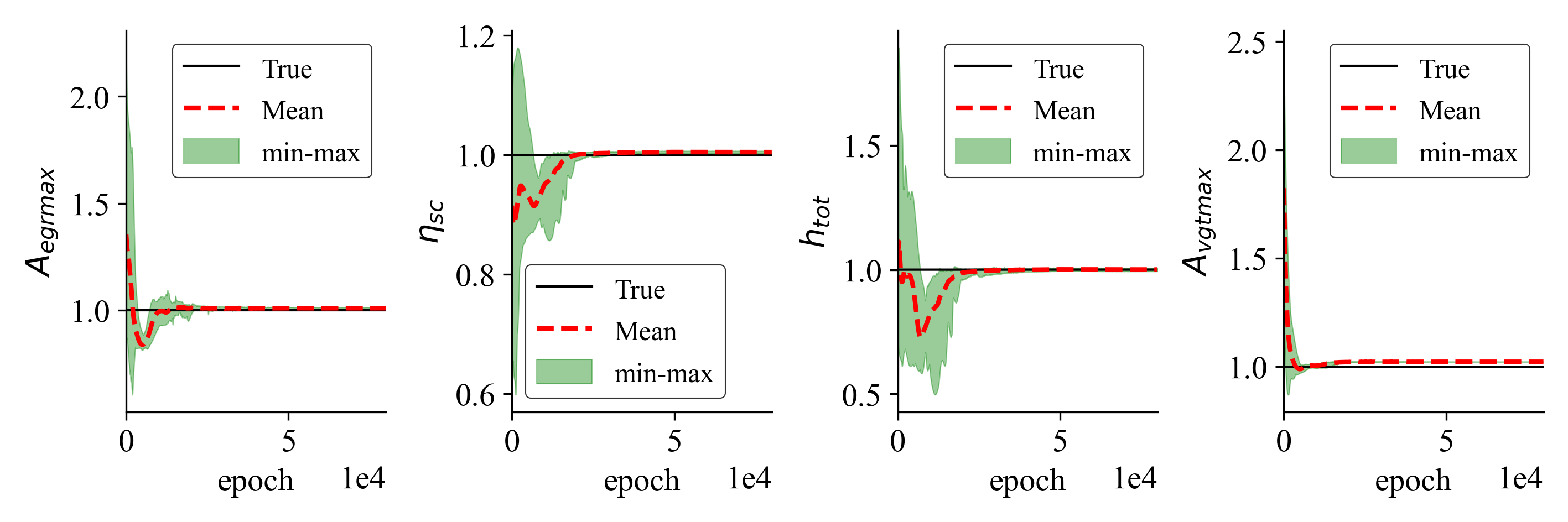}
(a) Baseline PINN (hybrid PINN)
\includegraphics[width=1.0\textwidth,height=3.5cm]{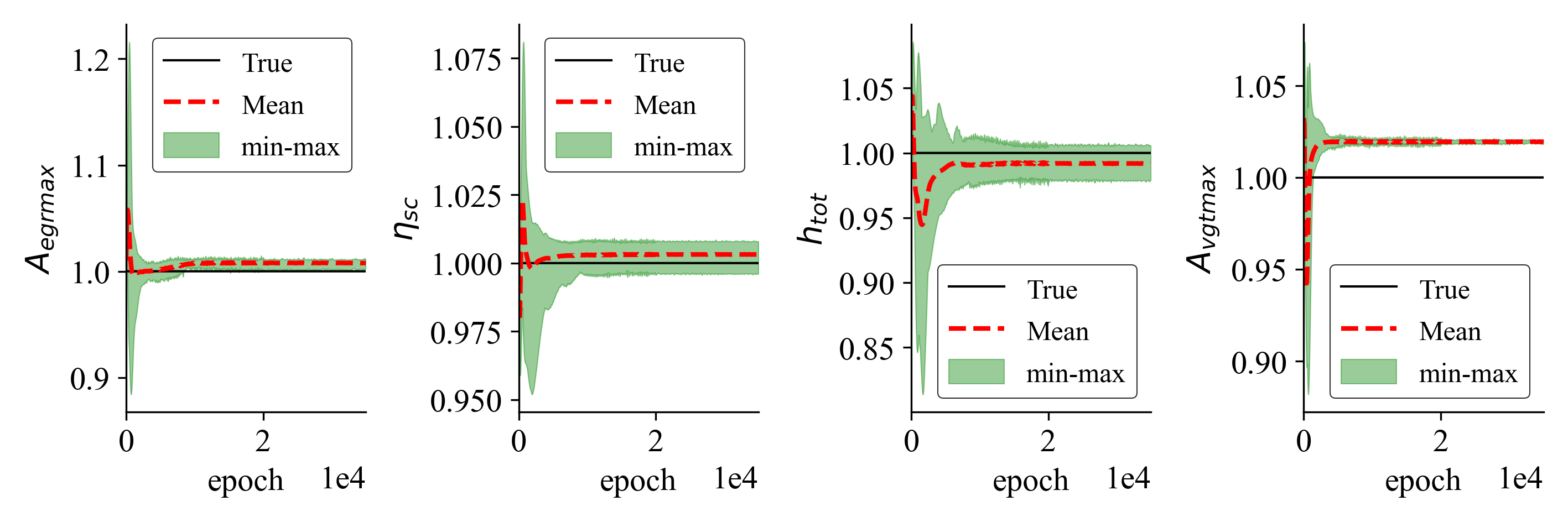}
(b) Multi-stage transfer learning
\includegraphics[width=1.0\textwidth,height=3.5cm]{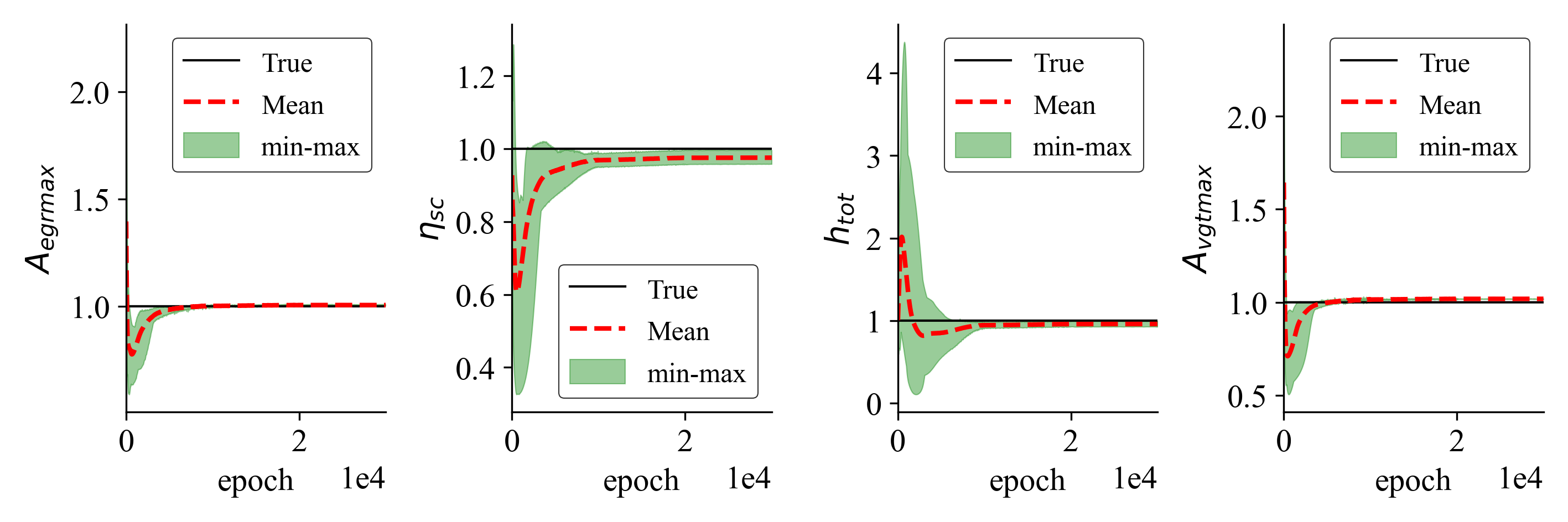}
(c) Few-shot transfer learning
\caption{\textbf{Parameter convergence, clean data, 151 data points, ensemble:} Convergence of the four unknown parameters of the engine showing the range of variation across the 30 independent runs for (a) Baseline PINN (hybrid PINN) model ($t = 0-1$ min), (b) Multi-stage transfer learning model ($t = 3-4$ min), and (c) Few-shot transfer learning model ($t = 3-4$ min) for clean data case with 151 data points in the loss function. The min-max band indicates the range over which the parameter estimation varies across the thirty independent runs that we used in this study. We observed that this variation is reduced significantly as training progresses, indicating a good repeatability of our PINN framework.}
\label{fig:param_conv_all}
\end{figure}

Fig. \ref{fig:all_losses} shows the loss convergence for the ODE and data loss terms for different engine states and gas flows. We note that all loss terms for the two TL paradigms converge faster than the baseline PINN model. Additionally, Fig. \ref{fig:param_conv_all} shows the convergence plot for the four unknown engine parameters generated through an ensemble across thirty independent runs for time segment $t = 3\sim4$ min (clean data with 151 data points) for base line PINN, multi-stage and few-shot TL. All unknown parameters for the TL methods converge at $\approx 20K$ epochs, while for the baseline PINN model, this takes $\approx 50K$ epochs. Also, compared to the multi-stage TL approach, the few-shot TL method uses significantly lower computational time. This is because the few-shot method eliminates the need for automatic differentiation for the output variable. However, this approach requires a larger corpus of labelled data for the baseline training of hidden layers (body) of the PINN networks, which are trained offline. Hence, trade-offs between data availability, computational and accuracy requirements are essential for choosing one method over the other.

\section{Limitations and future work}
\label{Section:Limitations}
Despite the demonstrated efficacy and advancements of our hybrid approach, there exist certain limitations of this method that we acknowledge. First, the systemic uncertainty has been modelled using fixed dropout rates in our empirical models that represent outputs from largely independent systems. The dropout rate in this study was chosen heuristically, and future work will explore learning data-driven dropout schedules for these systems. Second, while our few-shot TL strategy reduces fine-tuning costs, its performance is sensitive to the representativeness of the pre-training dataset and requires a copious amount of data for effective learning. Third, we note that the input data obtained from field measurements were collected under normal loading conditions for a range of engine speeds and actuator conditions. The current model may require fine-tuning of hyperparameters (such as self-adaptive weights) to deal with special loading conditions that may be encountered in the field. However, we believe that adapting to new scenarios will likely require marginal effort due to the physics-based regularization embedded in our framework. Lastly, our future research work will explore methods that can help reduce the aleatoric uncertainty leading to high variability in parameter estimation under noisy conditions. While the proposed methods show good accuracy for clean data, we observed a few instances (runs) of overfitting while using noisy data. To control this overfitting, one can consider early stopping and a good learning rate schedule. To further mitigate these issues, complementary robustness strategies can be considered in future work. For example, data augmentation using input perturbations or noise realizations can help improve generalization in our few-shot method. Adversarial training provides another approach to enhance model robustness by optimizing the model against worst-case perturbations. Additionally, robust regularization or loss formulations can also suppress overfitting and improve model's performance. Together, these methods provide a complementary pathway for improving the reliability and real-world resilience of the proposed few-shot transfer learning framework.

\section{Summary}
\label{Section:Summary}
In this study, we developed two computationally efficient frameworks for parameter estimation in diesel engine subsystems by integrating Physics-Informed Neural Networks (PINNs) with neural operator and transfer learning strategies. The proposed approach advances the state of the art in physics-based fault diagnostics and health monitoring of diesel engines by addressing two primary limitations of existing (conventional and neural network-based) methods. Firstly, the proposed method addresses the need to create a holistic framework for predicting engine dynamics at a system level, which encompasses all critical air flow sub-systems simultaneously. Secondly, our framework addresses a key challenge of incorporating models grounded in physical knowledge with measurable sensory data to estimate gas flow dynamics, along with the identification of unknown system parameters for use in continuous health monitoring during system operation.

\par To evaluate the applicability of this methodology in real scenarios, we introduced systematic variability through dropouts in pre-trained empirical networks for generating stochastic outputs from these networks. Additionally, we presented a hybrid framework that leverages Deep Operator Networks (DeepONet) for accelerated state predictions, with two transfer learning (TL) strategies designed to reduce computational time to meet requirements for field use. The first strategy, a \textbf{multi-stage TL} method relies on knowledge transfer between our PINN-based neural networks when learning system dynamics and parameter estimation across different temporal segments of the input signal. The second strategy, a \textbf{few-shot TL} approach leverages a large pre-trained neural network to generalize estimation of gas dynamics across multiple operating conditions simultaneously. Our results, based on simulated data with field-derived input signals and corrupted with Gaussian noise to mimic aleatoric uncertainty in sensory signals demonstrate that both methods effectively estimate unknown system parameters and dynamics states across diverse operating conditions. Notably, under noisy conditions, the parameter estimates exhibit variability in both methods which is expected since this challenge is shared by most existing approaches. Nonetheless, the proposed framework, demonstrates significant promise towards the use of physics-based methods in engine health monitoring and is a major step forwards as compared to prior works:
\begin{itemize}[leftmargin=*]
    \item Compared to pure data-driven methods such as \cite{Tosun_2016, pulpeiro2022_dieselmod, active_fault_tolerance}, our method offers improved interpretability by estimating physical parameters of engine sub-systems which are well defined by physical models. Compared to existing methods where a residual signal is used for fault estimation, our proposed method identifies scalar parameters from measured sensory data in an inverse problem paradigm. Drifts in these physical parameters can be monitored for triggering a fault as compared to a temporal residual signal, where momentary drifts over a short segment could lead to false positives. Further, existing methods for predicting engine dynamics are primarily focused on monitoring individual sub-components and their behaviour. On the contrary, our hybrid approach provides simultaneous estimates for gas flow dynamics and unknown system parameters for the complete engine model. This is a significant advancement for building comprehensive strategies for fault and health monitoring using system-level information in series rather than analysing individual component blocks for anomalies.
    \item Compared to a baseline PINNs implementation \cite{Nath_2023}, our hybrid approach advances it further since it reduces the computational time as independent state variables are approximated using offline trained neural operators, DeepONet in this case. While the hybrid \textbf{multi-stage TL} approach reduces computational time by approximately 50\%, the \textbf{few-shot TL} paradigm further improves the computational time by an additional 25\% in comparison to the hybrid PINN model.
    \item By fusing physics-based knowledge with measured sensory data, our proposed method provides a framework for extending use of neural network-based models across different operating conditions which otherwise is challenging. To test robustness of our framework, we use different sets of unknown parameters to generate data for pre-trained networks and DeepONet while inferring an unseen set of parameters with PINN learning. Existing neural network approaches that learn purely empirical mappings between inputs and outputs will likely produce erroneous estimates when evaluated at unseen conditions.
\end{itemize}

\par The key scientific and practical value added by this study lies in the fusion of physics-based neural networks with operator learning (data-driven) and transfer learning strategies as a step towards real-time, explainable, and unambiguous health monitoring for diesel engines. By transitioning from a residual signal-based fault detection paradigm, this framework lays the groundwork for system-wide fault detection and mitigation strategies that are rooted in physics-based knowledge as well as measured sensory data from the engine. Hence, we believe such a framework is well-suited for developing digital twin strategies for modelling diesel engines in the future.

\section*{Input data and data generation}
The input data are collected from actual engine running conditions. These data are used to generate simulated data with different ambient conditions using the Simulink file accompanied \cite{Wahlstrom} and available at \cite{Simulink_file}.

\section*{Author Contributions Statement}
\textbf{Kamaljyoti Nath:} Conceptualization, Methodology, Software, Validation, Investigation, Visualization, Result generation and analysis, Writing- original draft; \textbf{Varun Kumar:} Validation, Investigation, Visualization, Result generation and analysis, Writing- original draft; \textbf{Daniel J. Smith:} Conceptualization, Data curation, Project administration, Supervision, Writing - original draft; \textbf{George Em Karniadakis:} Conceptualization, Methodology, Funding acquisition, Project administration, Resources, Supervision, Writing - original draft. All authors reviewed the manuscript. 

\section*{Acknowledgement}
This research was conducted using computational resources and services at the Center for Computation and Visualization, Brown University. This study was funded by Cummins Inc. USA.

\bibliography{references}  
\bibliographystyle{unsrtnat}
\vspace{5cm}
\renewcommand{\thesection}{A. \Roman{section}} 
\renewcommand{\thesubsection}{\thesection.\arabic{subsection}}
\renewcommand{\theequation}{A.\arabic{equation}}
\renewcommand{\thetable}{A-\arabic{table}} 
\renewcommand{\thefigure}{A-\arabic{figure}} 

\clearpage
\begin{appendices}
\begin{center}
    \Large \textbf{Appendices} 
\end{center}
\setcounter{equation}{0}
\setcounter{table}{0}
\setcounter{figure}{0}

\begin{itemize}[leftmargin=*]
    \item Appendix \ref{Appendix:Engine model}: A description of the engine model considered in this study. 
    \item Appendix \ref{Appendix:PINN and DeepONet}: A brief description of vanilla PINN model 
    \item Appendix \ref{Section:Appendix:Flowchart}: Detailed flow chart for the hybrid PINN model
    \item Appendix \ref{Suppl:Detail loss function}: Explanation of different loss terms used for training are discussed.    
    \item Appendix \ref{Subsection:Numerical:DeepONet}: Detail and results of DeepONet for approximation of EGR and VGT states.
    \item Appendix \ref{sec:additional_plots_MTL}: Additional results for multi-stage transfer learning.
    \item Appendix \ref{Appendix:Additional figure for few-shot transfer learning}: Additional results for few-shot transfer learning.
\end{itemize}
\section{Engine model}
\label{Appendix:Engine model}
In this section we present the complete numerical model for the mean value model considered in this study. We have considered the mean value engine model proposed by Wahlstr\"{o}m and Eriksson \cite{Wahlstrom} in our present study. We only present the equation here, more detail can be found in ref \cite{Wahlstrom}.

\subsubsection*{Manifold pressures}
\begin{equation}
    \dfrac{d}{dt} p_{im} = \dfrac{R_a T_{im}}{V_{im}}(W_c+W_{egr} - W_{ei})
    \label{Eq:Append:p_im}
\end{equation}
\begin{equation}
    \dfrac{d}{dt} p_{em} = \dfrac{R_e T_{em}}{V_{em}}(W_{eo} - W_t - W_{egr})
    \label{Eq:Append:p_em}
\end{equation}
\subsubsection*{Cylinder}
\begin{align}
    W_{ei} & =  \dfrac{\eta_{vol}p_{im}n_eV_d}{120R_aT_{im}}
    \label{Eq:Append:W_ei} \\
    W_{eo} & = W_f + W_{ei}
    \label{Eq:Append:W_eo}
\end{align}
\begin{equation}
    W_f = \dfrac{10^{-6}}{120}u_\delta n_e n_{cyl}
    \label{Eq:Append:W_f}
\end{equation}
\begin{equation}
    \eta_{vol} = c_{vol1} \sqrt{p_{im}} +c_{vol2} \sqrt{n_e} + c_{vol3}
    \label{Eq:Append:eta_vol}
\end{equation}
\begin{equation}
    T_e = \eta_{sc}\Pi_e^{1-1/\gamma_a}r_c^{1-\gamma_a}x_p^{1/\gamma_a -1}\left[q_{in}\left(\dfrac{1-x_{cv}}{c_{pa}} + \dfrac{x_{cv}}{c_{Va}} \right) + T_1r_c^{\gamma_a-1}\right]
    \label{Eq:Append:T_e}
\end{equation}
\begin{equation}
    \Pi_e = \dfrac{p_{em}}{p_{im}}
    \label{Eq:Append:Pi_e}
\end{equation}
\begin{equation}
    T_1 = x_rT_e + (1-x_r)T_{im}
    \label{Eq:Append:T_1}
\end{equation}
\begin{equation}
    x_r = \dfrac{\Pi_e^{1/\gamma_a}x_p^{-1/\gamma_a}}{r_c x_v}
    \label{Eq:Append:x_r}
\end{equation}
\begin{align}
    x_p = & \dfrac{p_3}{p_2} = 1 + \dfrac{q_{in}x_{cv}}{c_{Va}T_1r_c^{\gamma_a-1}}
    \label{Eq:Append:x_p} \\
    x_v = & \dfrac{v_3}{v_2} = 1 + \dfrac{q_{in}(1-x_{cv})}{c_{pa}[(q_{in}x_{cv}/c_{va})+T_1r_c^{\gamma_a - 1}]}
    \label{Eq:Append:x_v}
\end{align}
\begin{equation}
    q_{in} = \dfrac{W_fq_{HV}}{W_{ei}+Wf}(1-x_r)
\end{equation}
\begin{equation}
    x_r = \dfrac{\Pi_e^{1/\gamma_a}x_p^{-1/\gamma_a}}{r_c x_v}
    \label{Eq:Append:x_r1}
\end{equation}
\begin{equation}
    T_{em} = T_{amb} + (T_e - T_{amb})\exp\left(\dfrac{-h_{tot}\pi d_{pipe}l_{pipe}n_{pipe}}{W_{eo}c_{pe}}\right)
    \label{Eq:Append:T_em}
\end{equation}

\subsubsection*{EGR valve}
\label{Appendix:EGR valve}
\begin{align}
    \dfrac{d\Tilde{u}_{egr1}}{dt} & = \dfrac{1}{\tau_{egr1}}\left[u_{egr}(t-\tau_{degr}) - \Tilde{u}_{egr1}\right]
    \label{Eq:Append:u_egr_1}\\
    \dfrac{d\Tilde{u}_{egr2}}{dt} & = \dfrac{1}{\tau_{egr2}}\left[u_{egr}(t-\tau_{degr}) - \Tilde{u}_{egr2}\right]
    \label{Eq:Append:u_egr_2} \\
    \Tilde{u}_{egr} & = K_{egr}\Tilde{u}_{egr1} - (K_{egr} - 1)\Tilde{u}_{egr2}
    \label{Eq:Append:u_egr}
\end{align}
\begin{equation}
    W_{egr} = \dfrac{A_{egr}p_{em}\Psi_{egr}}{\sqrt{T_{em}R_e}}
    \label{Eq:Append:W egr}
\end{equation}
\begin{equation}
    \Psi_{egr} = 1 - \left(\dfrac{1 - \Pi_{egr}}{1-\Pi_{egropt}} -1 \right)^2
    \label{Eq:Append:Psi egr}
\end{equation}
\begin{equation}
    A_{egr} = A_{egrmax}f_{egr}(\Tilde{u}_{egr})
    \label{Eq:Append:A egr}
\end{equation}
\begin{equation}
    \Pi_{egr} = \begin{cases}
    \Pi_{egropt} \;\;\;\;\;\; & \text{if}\;\; \dfrac{p_{im}}{p_{em}}< \Pi_{egropt} \\
    \dfrac{p_{im}}{p_{em}} \;\;\;\;\;\; & \text{if}\;\; \Pi_{egropt} \le \dfrac{p_{im}}{p_{em}} \le 1 \\
    1 \;\;\;\;\;\; & \text{if} \;\; 1<\dfrac{p_{im}}{p_{em}}
    \end{cases}
    \label{Eq:Append:Pi_egr}
\end{equation}
\begin{equation}
    f_{egr}(\Tilde{u}_{egr}) = \begin{cases}
    c_{egr1}\Tilde{u}_{egr}^2 + c_{egr2}\Tilde{u}_{egr} + c_{egr3} \;\;\; & \text{if}\;\; \Tilde{u}_{egr}\le -\dfrac{c_{egr2}}{2c_{egr1}} \\
    c_{egr3} - \dfrac{c_{egr2}^2}{4c_{egr1}} \;\;\; & \text{if}\;\; \Tilde{u}_{egr}> -\dfrac{c_{egr2}}{2c_{egr1}}
    \end{cases}
    \label{Eq:Append:f_egr}
\end{equation}

\subsubsection*{Turbocharger}
\label{Appendix:Turbocharger}
\begin{equation}
    \dfrac{d}{d t}\omega_t = \dfrac{P_t\eta_m - P_c}{J_t\omega_t}
    \label{Eq:Append:omega_t}
\end{equation}
\begin{equation}
    \dfrac{d\Tilde{u}_{vgt}}{dt} = \dfrac{1}{\tau_{vgt}}\left(u_{vgt}(t-\tau_{dvgt}) - \Tilde{u}_{vgt}\right)
    \label{Eq:Append:u_vgt}
\end{equation}
\begin{equation}
    W_t = \dfrac{A_{vgtmax}p_{em}f_{\Pi_t}(\Pi_t)f_{vgt}(\Tilde{u}_{vgt}))}{\sqrt{T_{em}R_e}}
    \label{Eq:Append:W_t}
\end{equation}
\begin{equation}
    f_{\Pi_t}(\Pi_t) = \sqrt{1-\Pi_t^{K_t}}
    \label{Eq:Append:f_Pi_t}
\end{equation}
\begin{eqnarray}
    \Pi_t =\dfrac{p_{amb}}{p_{em}}
\end{eqnarray}
\begin{equation}
    \left[\dfrac{f_{vgt}(\Tilde{u}_{vgt}) - c_{f2}}{c_{f1}}\right]^2 + \left[\dfrac{\Tilde{u}_{vgt} - c_{vgt2}}{c_{vgt1}}\right]^2= 1
\end{equation}
\begin{equation}
    f_{vgt}(\Tilde{u}_{vgt}) = c_{f2} + c_{f1}\sqrt{\text{max}\left(0, 1 - \left(\dfrac{\Tilde{u}_{vgt} - c_{vgt2}}{c_{vgt1}}\right)^2\right)}
    \label{Eq:Append:f vgt}
\end{equation}
Here we define a variable $F_{vgt,\Pi_t}$ as
\begin{equation}
    F_{vgt,\Pi_t} = f_{vgt}(\tilde{u}_{vgt})\times f_{\Pi_t} (\Pi_t)
    \label{Eq:Appendix:F_VGT_PI_T_Cal}
\end{equation}
\begin{equation}
    P_t\eta_m = \eta_{tm}W_tc_{pe}T_{em}\left(1 - \Pi_t^{1-1/\gamma_e}\right)
    \label{Eq:Append:P_t_eta_m}
\end{equation}
\begin{equation}
    \eta_{tm} = \eta_{tm,max} - c_m(BSR - BSR_{opt})^2
    \label{Eq:Append:eta_tm}
\end{equation}
\begin{equation}
    BSR = \dfrac{R_t\omega_t}{\sqrt{2c_{pe}T_{em}(1-\Pi_t^{1-1/\gamma_e})}}
\end{equation}
\begin{equation}
    c_m = c_{m1}[max(0, \omega_t - c_{m2}]^{c_{m3}}
\end{equation}
\subsubsection*{Compressor}
\label{Appendix:Compressor}
\begin{equation}
    \eta_c = \dfrac{P_{c,s}}{P_c} = \dfrac{T_{amb}(\Pi_c^{1-1/\gamma_a}-1)}{T_c - T_{amb}}
    \label{Eq:Append:eta c 1}
\end{equation}
\begin{equation}
    \Pi_c = \dfrac{p_{im}}{p_{amb}}
\end{equation}
\begin{equation}
    P_{c,s} = W_c c_{pa}T_{amb}(\Pi_c^{1-1/\gamma_a}-1)
    \label{Eq:Append:P_c,s}
\end{equation}
\begin{equation}
    P_c = \dfrac{P_{c,s}}{\eta_c} = \dfrac{W_cc_{pa}T_{amb}}{\eta_c} (\Pi_c^{1-1/\gamma_a - 1})
    \label{Eq:Append:P_c}
\end{equation}
\begin{equation}
    \eta_c = \eta_{cmax} - \bm{\mathcal{X}}^T\bm{Q}_c\bm{\mathcal{X}}
    \label{Eq:Append:eta c}
\end{equation}
\begin{equation}
    \bm{\mathcal{X}} = \begin{bmatrix}
        W_c - W_{copt} \\
        \pi_c - \pi_{copt}
    \end{bmatrix}
\end{equation}
\begin{equation}
    \pi_c = (\Pi_c -1)^{c_\pi}
\end{equation}
\begin{equation}
    \bm{Q}_c = \begin{bmatrix}
    a_1 & a_3 \\
    a_3 & a_2
    \end{bmatrix}
\end{equation}
\begin{equation}
    \Psi_c = \dfrac{2c_{pa}T_{amb}(\Pi_c^{1-1/\gamma_a}-1)}{R_c^2\omega_t^2}
    \label{Eq:Append:Psi c}
\end{equation}
\begin{equation}
    \Phi_c = \dfrac{W_c/\rho_{amb}}{\pi R_c^3\omega_t} = \dfrac{R_a T_{amb}}{p_{amb}\pi R_c^3\omega_t} W_c
    \label{Eq:Append:Phi_c}
\end{equation}
\begin{equation}
    c_{\Psi1}(\omega_t)(\Psi_c - c_{\Psi2} )^2 + c_{\Phi1}(\omega_t)(\Phi_c -c_{\Phi2})^2 = 1
    \label{Eq:Append:elliplse c_psi}
\end{equation}
\begin{equation}
    c_{\Psi1}(\omega_t) = c_{\omega_{\Psi1}}\omega_t^2 + c_{\omega_{\Psi2}}\omega_t + c_{\omega_{\Psi3}}
    \label{Eq:Append:1}
\end{equation}
\begin{equation}
    c_{\Phi1}(\omega_t) = c_{\omega_{\Phi1}}\omega_t^2 + c_{\omega_{\Phi2}}\omega_t + c_{\omega_{\Phi3}}
    \label{Eq:Append:2}
\end{equation}
\begin{equation}
    W_c = \dfrac{p_{amb} \pi R_c^3\omega_t}{R_aT_{amb}} \Phi_c
    \label{Eq:Append:W_c}
\end{equation}
\begin{equation}
    \Phi_c = \sqrt{\text{max}\left(0, \dfrac{1-c_{\psi1}(\Psi_c - c_{\Psi 2})^2}{c_{\Phi 1}} \right)} + c_{\Phi 2}
    \label{Eq:Append:3}
\end{equation}
For the value considered for the parameters in the equation interested readers may refer to  previous work \cite{Nath_2023}. We have shown detail flow chart for calculation of physics-informed losses in Appendix \ref{Section:Appendix:Flowchart}.

\section{Introduction to PINN}
\label{Appendix:PINN and DeepONet}
In the present study, we have considered Physics-Informed Neural Networks (PINNs). In this section, we briefly discuss the basic idea of PINN. PINN consists of two main parts: an approximation of a solution of the state variable of a PDE/ODE using a neural network, and the second part is the calculation of a physics-informed loss function. We will discuss it in terms of ODE, as in the present study, our problem is confined to ODEs only, and the concept can be translated to PDEs. PINNs have two different paradigms, the first one is the solution of a differential equation with known initial (and boundary) conditions, which we call the forward problem. The second one is the evaluation of unknown parameters ($\bm{\Lambda}$) (field) from measured data, which is known as an inverse problem. The neural network approximate the state variable $y$ whose input is time $t$. Thus,
\begin{equation}
    y = \mathcal{N}_y(\bm{\theta}, t)
\end{equation}
where $\mathcal{N}_y$ represent the neural network parametrized with $\bm{\theta}=\{\bm{W},\bm{b}\}$ are the weights ($\bm{W}$) and biases ($\bm{b}$) of the neural network. The derivative of the state variable, $\dfrac{dy}{dt}$ is generally calculated using automatic differentiation \citep{Baydin_2018} of $y$ with respect to input $t$. Once the derivative of the state variable is calculated, the next step is to calculate the residual of the differential equation. This is done by adding the derivative and other parts in the differential equation. A PINN model has two kinds of loss functions. The first one is the Physics-informed loss, which is a function of the residual (generally MSE loss of the residue). The second one is the data loss. In the case of the forward problem, the data loss is due to initial condition (and boundary condition, in case of PDEs), which is generally the MSE loss of the data. In case of the inverse problem, there is an additional data loss due to the known value of the state (or derived) variable. The total loss is given as the weighted sum of the physic loss and data loss, 
\begin{equation}
    \mathcal{L}(\bm{\theta}) = \lambda_1\mathcal{L}_{\text{data}} + \lambda_1\mathcal{L}_{\text{phy}}
\end{equation}
where $\lambda_1$ and $\lambda_2$ are weights which can be fixed or adaptive \citep{McClenny_2020}. The parameters $\bm{\theta}$ in the case of a forward problem or $\bm{\theta}$ and $\bm{\Lambda}\}$ in the case of an inverse problem and the weights ($\lambda$) may be optimized using a min-max problem. 
\begin{equation}
    \min\limits_{\bm{\theta}, \bm{\varLambda}}\;\max\limits_{\lambda} \mathcal{L}(\bm{\theta},\varLambda)
\end{equation}
If we consider a gradient descent-ascent approach, we can express the optimization process as
\begin{subequations}
    \begin{align}
        \bm{\theta},\bm{\varLambda} & = \bm{\theta},\bm{\varLambda} - lr_{\theta,\varLambda}\nabla_{\theta,\varLambda}\mathcal{L}(\bm{\theta},\varLambda) \\ 
        \lambda & = \lambda + lr_{\lambda}\nabla_{\lambda}\mathcal{L}(\bm{\theta},\varLambda),
    \end{align}
\end{subequations}
where $lr_\theta$ denotes the network optimiser learning rate for $\bm{\theta}$, $lr_{\varLambda}$ represents the learning rate for the optimizer used to identify unknown scalar parameters $\bm{\varLambda}$, and $lr_{\lambda}$ refers to the learning rate for the optimizer used for updating self-adaptive weights. Incorporating distinct learning rates and optimizers allows for fine tuning of each component for efficient convergence.

\section{Detailed flow chart}
\label{Section:Appendix:Flowchart}
\begin{figure}[H]
    \centering
    \includegraphics[width=1\textwidth]{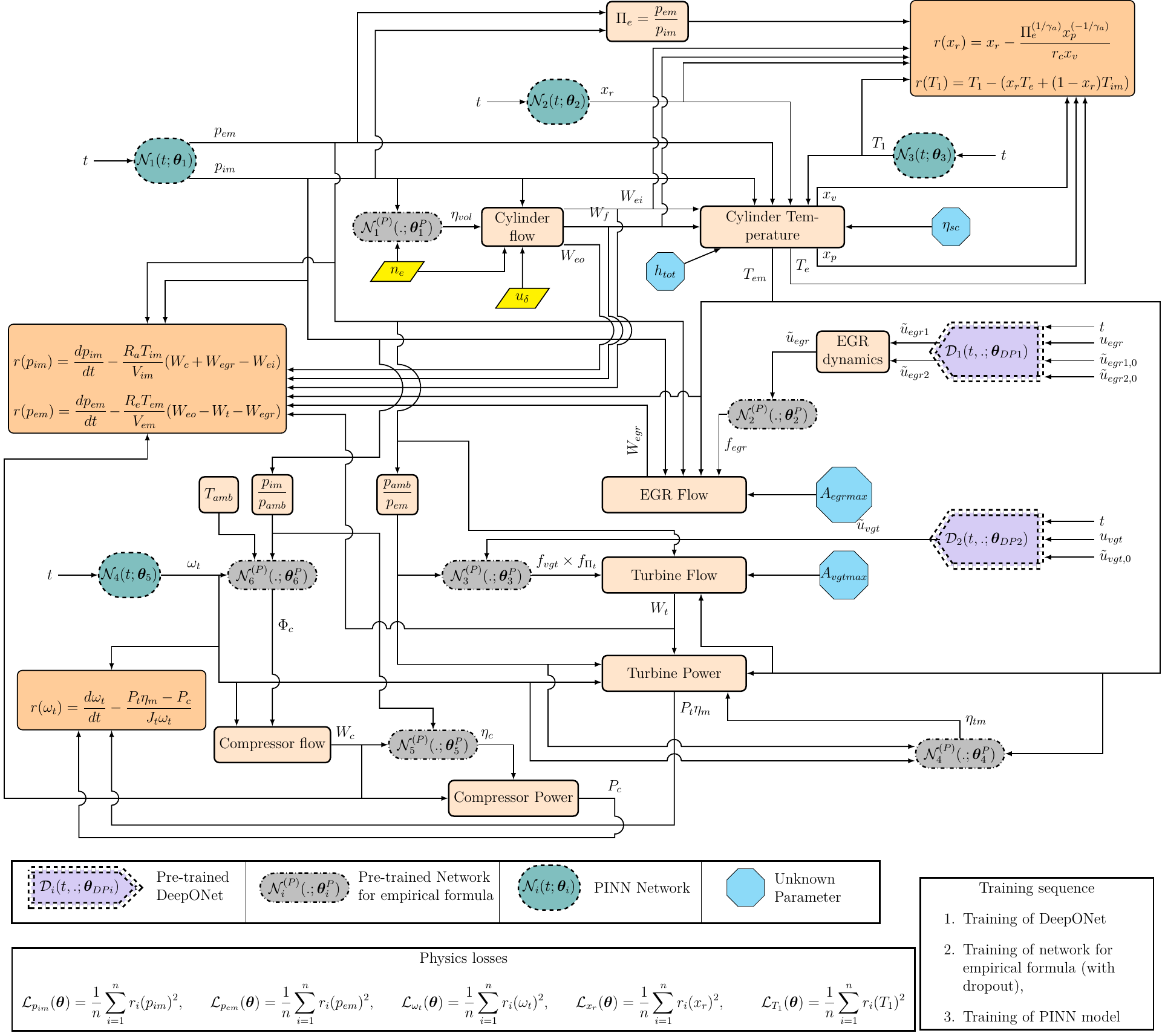}
    \caption{\textbf{Flow chart for proposed hybrid-PINNs model:} Flow chart for the proposed hybrid PINN model for the inverse problem for the engine for prediction of dynamics of the system variables and estimation of unknown parameters. The inputs are input control vector $\{u_\delta, u_{egr}, u_{vgt}\}$ and engine speed $n_e$. Four neural network $\mathcal{N}_i(t;\bm{\theta})$, $i=1,2,3,4$ indicated in dashed rectangular oval takes time $t$ as input and predict $p_{im}$, $p_{em}$, $x_r$ and $T_1$ and $\omega_{t}$. Two DeepONets $\mathcal{D}_i(.;\bm{\theta}_{DPi}$, $i=1,2$ indicated in double line rectangular pentagon takes time $t$ and others as input and predict $\tilde{u}_{egr1}$, $\tilde{u}_{egr2}$ and $\tilde{u}_{vgt}$ as shown in Table \ref{Table:DeepONet surrogates}. The parameters (weights and biases) of these pre-trained DeepONets are kept fixed to predict the three states. Four unknown parameters indicated in octagon are $\eta_{sc}$, $h_{tot}$, $A_{egrmax}$ and $A_{vgtmax}$. Six pre-trained neural networks $\mathcal{N}_i^{(P)}(.;\bm{\theta})$, $i=1,2,\dots,6$ indicated in dashed-dotted rectangular oval takes appropriate input and predict the empirical formulae as shown in Table \ref{Table:Surrogate empirical formulae}. The parameters (weights and biases) of these pre-trained DNNs are kept fixed to predict the empirical formulae.    There are eight main blocks calculating different variables. The equations for the calculation of each of the quantities are shown in Appendix \ref{Appendix:Engine model}. \textbf{Cylinder flow:} calculates $W_{ei}$, $W_f$ and $W_{eo}$ using Eqs. \ref{Eq:Append:W_ei}, \ref{Eq:Append:W_f} and \ref{Eq:Append:W_eo} respectively. \textbf{Cylinder Temperature:} calculates $x_v$, $x_p$, $T_e$ and $T_{em}$ using Eqs. \ref{Eq:Append:x_v}, \ref{Eq:Append:x_p}, \ref{Eq:Append:T_e} and \ref{Eq:Append:T_em} respectively. $h_{tot}$ and $\eta_{sc}$ are considered as learnable parameters in the calculation of $T_{em}$ and $T_e$ respectively. \textbf{EGR dynamics:} calculated $\Tilde{u}_{egr}$ using Eq. \ref{Eq:Append:u_egr}. \textbf{EGR Flow:} calculates EGR mass flow $W_{egr}$ using Eq. \ref{Eq:Append:W egr}. $A_{egrmax}$ is considered as learnable parameter. \textbf{Compressor flow:} calculates compressor mass flow $W_c$ using Eq. \ref{Eq:Append:W_c}. \textbf{Compressor Power:} calculates compressor power $P_c$ using Eq. \ref{Eq:Append:P_c}.\textbf{Turbine Flow:} calculates turbine mass flow $W_t$ using Eq. \ref{Eq:Append:W_t}. $A_{vgtmax}$ is considered as trainable parameter. textbf{Turbine Power:} calculates effective turbine power $P_t\eta_m$ using Eq. \ref{Eq:Append:P_t_eta_m} 
    There are three blocks, which calculate the residual of the equation. \textbullet\ The first block calculates the residual for state equations for $p_{im}$ and $p_{em}$, \textbullet\ The second one calculates the residual for the equations of $x_r$ and $T_1$, \textbullet\ The third block calculates the residual for the state equation for $\omega_t$. There is another block that calculates the physics losses. The calculation for data losses are not shown. The figure is adopted from \cite{Nath_2023}.}
    \label{Fig:flowchart}
\end{figure}

We have discussed the problem setup in section \ref{Subsection:Problem setup}, and in subsequent sections, we have discussed how we can approximate different variables using DNNs. Further, we discussed the basic philosophy and implementation of PINN along with DeepONet in the present problem context. We present a flowchart for calculating the physics-informed loss for the present problem in Fig. \ref{Fig:flowchart}. We also note that in the flowchart, we have not shown the data loss and self-adaptive weights. We briefly summarize below the sequence of training for various parts of the proposed workflow,
\begin{enumerate}[leftmargin=*]
    \item \textbf{DeepONet training:} We consider two DeepONets, the first one predicts the two EGR states $\tilde{u}_{egr1}$ and $\tilde{u}_{egr2}$, and the second DeepONet predicts the VGT state $\tilde{u}_{vgt}$. These are trained offline using laboratory data as discussed in Section \ref{Subsection:DeepONet} and Appendix \ref{Subsection:Numerical:DeepONet}. The DeepONets are considered pre-trained in the PINN formulation, and the parameters of the DeepONets are fixed during PINN training.
    \item \textbf{Training of DNN for empirical formulae:} We consider six DNN to approximate the empirical formula for $\eta_{vol}$, $f_{egr}$, $F_{vgt,\Pi_t}$, $\eta_{tm}$, $\eta_{c}$ and $\Phi_{c}$ as discussed in Section \ref{Subsection:Neural network surrogates for empirical formulae}. These DNNs are trained offline using laboratory data. Similar to the DeepONets, the parameters of these DNNs are also fixed during the PINN training; however, the dropout is considered.
    \item \textbf{PINN training for inverse problem:} We consider four networks to approximate the state variables $p_{em}$, $p_{im}$, $\omega_t$ and $x_r$ and $T_1$. We formulate the hybrid PINN model as discussed in section \ref{Subsection:PINN for engine problem}, combining the pre-trained DeepONets for the EGR and VGT states and the pre-trained DNNs for the empirical formulae. In the hybrid PINN, the trainable parameters are parameters of the PINN network, unknown parameters for the inverse problem, and the self-adaptive weights. The parameters of the DeepONet and DNN for the empirical formula are kept fixed in the hybrid PINN.
\end{enumerate}
We consider two different kinds of datasets, the laboratory data and the field data. The laboratory data are used to train the pre-trained networks, i.e., the DeepONet and the DNN for empirical formulas. The field data are used in the inverse problem using PINN. We also consider two transfer learning paradigms,
\begin{itemize}[leftmargin=*]
    \item \textbf{Multi-stage TL:} In this approach the hybrid PINN model is trained for the first segment of time ($0 \sim 60$ sec). In the subsequent time segments, we considered a multi-stage transfer learning as discussed in Section \ref{Subsection:Transfer learning}.
    \item \textbf{Few-shot TL:} Few-shot TL consists of two phases. In the first phase, we trained the multi-head networks for state variables $p_{em}$, $p_{im}$, $\omega_t$ and $x_r$ and $T_1$ offline using data loss function. In this phase, the multi-head networks are trained using laboratory data. In the second phase, we replace the head and train only the head (output layer) of these networks using hybrid PINN as discussed in Section \ref{Subsection:Few shot PINN}. We also like to mention that in Few-shot TL we calculate the derivative of the output only once and keep it outside the training loop as discussed in Section \ref{Subsection:Few shot PINN}. In the second phase, the PINN networks are trained using field data. The trainable parameters are the weights and biases of the output layers of the PINN network, the unknown parameters, and the self-adaptive weights.
\end{itemize}

\section{Detailed loss function for engine model}
\label{Suppl:Detail loss function}

\par In section \ref{Subsection:Problem setup}, we discussed the inverse problem of interest in this study. Our objectives  are to:  (1) learn the dynamics of the six gas flow dynamics state estimates, three of which are approximated using two DeepONets, and (2) infer the unknown parameters in the system, given field measurements of system variables $\{p_{im}, p_{em}, \omega_t, W_{egr}, T_{em}\}$ as well as Eqs. \eqref{Eq:p_im}-\eqref{Eq:x_r}, using PINNs. To achieve these objectives, we utilize four DNNs as surrogates for the solutions to different equations, as shown in Table \ref{Table:FNN for PINN}. We encode physics loss for Eqs \eqref{Eq:p_im}-\eqref{Eq:x_r} (EGR, VGT actuator equations are not included as these are approximated using DeepONets) using the automatic differentiation and formulate the physics-informed loss. The total loss function considered for training the PINNs model is given by:
\begin{equation}
    \begin{split}
    \mathcal{L}(\bm{\theta}, \bm{\mathcal{\varLambda}}, \bm{\lambda}_{p_{im}}, \bm{\lambda}_{p_{em}}, \bm{\lambda}_{\omega_t}, \bm{\lambda}_{W_{egr}}, \lambda_{T_{em}}, \lambda_{T_1}) = & \mathcal{L}_{p_{im}} + \mathcal{L}_{p_{em}} + \mathcal{L}_{\omega_{t}} + 100\times \mathcal{L}_{x_{r}} + \lambda_{T_1}\times\mathcal{L}_{T_{1}} + \\
    & \mathcal{L}^{ini}_{p_{im}} + \mathcal{L}^{ini}_{p_{em}} + \mathcal{L}^{ini}_{\omega_{t}} + 100\times \mathcal{L}^{ini}_{x_{r}} + 100\times \mathcal{L}^{ini}_{T_{1}} + \\
    & \mathcal{L}^{data}_{p_{im}}(\bm{\lambda}_{p_{im}}) + \mathcal{L}^{data}_{p_{em}}(\bm{\lambda}_{p_{em}}) +     \mathcal{L}^{data}_{\omega_t}(\bm{\lambda}_{\omega_t}) + \\ & \mathcal{L}^{data}_{W_{egr}}(\bm{\lambda}_{W_{egr}}) + \lambda_{Tm}\mathcal{L}^{data}_{T_{em}},
    \end{split}
\label{Eq:Suppl: Loss total loss}
\end{equation}

\par The individual physics losses are represented as:
\begin{subequations} \label{eqn:loss_terms_PINN}
    \begin{align}
    \mathcal{L}_{p_{im}} = & \dfrac{1}{n}\sum_{k=1}^{n}r(p_{im}^{(k)})^2 = \dfrac{1}{n}\sum_{k=1}^{n}\left(\dfrac{dp_{im}^{(k)}}{dt} - \dfrac{R_a T_{im}}{V_{im}}\left(W_c^{(k)}+W_{egr}^{(k)}-W_{ei}^{(k)}\right)\right)^2 \\
    \mathcal{L}_{p_{em}} = &  \dfrac{1}{n}\sum_{k=1}^{n}r(p_{em} ^{(k)})^2 = \dfrac{1}{n}\sum_{k=1}^{n}\left(\dfrac{dp_{em}^{(k)}}{dt} - \dfrac{R_e T_{em}^{(k)}}{V_{em}}\left(W_{eo}^{(k)} - W_t^{(k)}-W_{egr}^{(k)}\right) \right)^2 \\
    \mathcal{L}_{\omega_{t}} = &  \dfrac{1}{n}\sum_{k=1}^{n}r(\omega_{t}^{(k)})^2 = \dfrac{1}{n}\sum_{k=1}^{n}\left(\dfrac{d\omega_t^{(k)}}{dt} - \dfrac{P_t^{(k)} \eta_{m}-P_c^{(k)}}{J_t \omega_t ^{(k)}} \right)^2\\
    \mathcal{L}_{x_{r}} = & \dfrac{1}{n}\sum_{k=1}^{n}r(x_{r}^{(k)})^2 = \dfrac{1}{n}\sum_{k=1}^{n}\left(x_r^{(k)} - \dfrac{(\Pi_e^{(k)})^{1/\gamma_a} (x_p^{(k)})^{-1/\gamma_a}}{r_c x_v^{(k)}}\right)^2\\
    \mathcal{L}_{T_{1}} = & \dfrac{1}{n}\sum_{k=1}^{n}r(T_{1}^{(k)})^2 = \dfrac{1}{n}\sum_{k=1}^{n}\left(T_1^{(k)} - \left(x_r^{(k)} T_e^{(k)} + (1- x_r^{(k)})T_{im}\right)\right)^2
	\end{align}
\end{subequations}
where $n$ and $r(.)$ are the number of residual points and residual, respectively. The data losses for the known field data are given by:
\begin{subequations}
\begin{align}
    \mathcal{L}^{data}_{\phi}(\bm{\lambda}_{\phi}) = & \dfrac{1}{n_\phi}\sum_{j=1}^{n_\phi}\left[\left(\phi_{data}^{(j)} - \hat{\phi}_{data}^{(j)}\right)\lambda_{\phi}^{(j)}\right]^2, \quad \quad \phi = p_{im}, p_{em}, \omega_t, W_{egr} \\
    \mathcal{L}^{data}_{T_{em}} = & \dfrac{1}{n_{T_{em}}}\sum_{j=1}^{n_{T_{em}}}\left[\left(T_{em_{data}}^{(j)} - \hat{T}_{em_{data}}^{(j)}\right)\right]^2
\end{align}
\end{subequations}
where $\{\phi, T_{em}\}$ and $\{\hat{\phi}, \hat{T}_{em}\}$ are the true and predicted values of the variable, respectively, and $n_\phi$ is the number of data points for the variable. $\lambda$ is the self-adaptive weight associated with the variables.
\par The initial condition losses are given as:
\begin{equation}
    \mathcal{L}^{ini}_{\phi} = \dfrac{1}{1}\sum_{j=1}^1\left(\phi_{0}^{(j)} - \hat{\phi}_{0}^{(j)}\right)^2, \;\;\;\;\;\;\;\; \phi= p_{im}, p_{em}, \omega_t, x_r, T_1
\end{equation}

\section{DeepONet: Approximation of EGR states ($\Tilde{u}_{egr1}$ and $\Tilde{u}_{egr2}$) and VGT states ($\Tilde{u}_{vgt}$) using DeepONets}
\label{Subsection:Numerical:DeepONet}
\begin{figure}[H]
    \centering
    \includegraphics[width=0.9\textwidth]{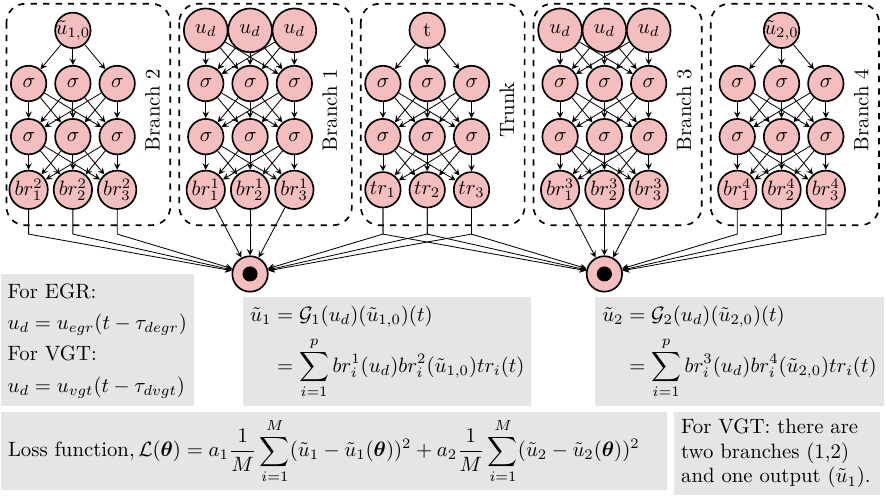}
    \caption{\textbf{Schematic diagram of DeepONet} for prediction of EGR states showing branches and trunk. The trunk network takes time as its input, at which the value of the output function needs to be evaluated. The four branches take $u_{d}$, $\tilde{u}_{1,0}$, $u_{d}$ and $\tilde{u}_{2,0}$ as input and approximate two outputs $\Tilde{u}_1 = \Tilde{u}_{egr1}$ and $\Tilde{u}_1 = \Tilde{u}_{egr2}$. Here, $\tilde{u}_{1,0} = \tilde{u}_{egr1,0}$ and $\tilde{u}_{2,0} = \tilde{u}_{egr2,0}$ are the initial conditions and $u_{d} = u_{degr} = u_{egr}(t-\tau_{degr})$, where $\tau_{degr}$ is the time delay coefficient. The loss function for individual output variables is the mean square error between the true value ($\Tilde{u}_1$) and the predicted one ($\Tilde{u}_1(\bm{\theta})$), and the total loss is the weighted sum ($a_1=1$, $a_2=100$) of loss for both outputs. In the case of prediction of VGT actuator states, there are only two branch networks, branch 1 and 2 and only one output $\tilde{u}_1 = \tilde{u}_{vgt}$. The inputs to these branch networks are $u_{d} = u_{dvgt} = u_{egr}(t-\tau_{dvgt})$ and the initial condition  $\tilde{u}_{1,0} = \tilde{u}_{vgt,0}$. Appropriate scaling of the input and output are considered for fast and better training. Details of the network sizes and input-output data structures are shown in Table \ref{Table:DeepONet input output}.}
    \label{Figure:DeepONet EGR}
\end{figure}

In section \ref{Subsection:DeepONet}, we discussed how the states describing the EGR actuator ($\tilde{u}{egr1}$ and $\tilde{u}{egr2}$) and the VGT actuator ($\tilde{u}{vgt}$) dynamics are approximated using two DeepONets. Fig. \ref{Figure:DeepONet EGR} illustrates a schematic diagram of the DeepONet architecture for the EGR actuator states. In this configuration, the DeepONet comprises of one trunk, which takes time as input, and four branch networks. The four branch networks receive $u_d = u_{degr} = u_{egr}(t-\tau_{degr})$, $\tilde{u}_{egr1,0}$, $u_d = u_{degr} = u_{egr}(t-\tau_{degr})$ and  $\tilde{u}_{egr2,0}$ as inputs, respectively. For the VGT actuator approximation, there are only two branch networks that take $u_d = u_{dvgt}=u_{vgt}(t-\tau_{dvgt})$ and $\tilde{u}_{vgt,0}$ as input. As discussed earlier, we have considered Causality-DeepONet \cite{Liu_2022_Causality} for these DeepONets. The input-output data structures for the DeepONet model are shown in Table \ref{Table:DeepONet input output}.
\par We train these two DeepONets using labelled data generated using Simulink model, representing laboratory test measurements in real systems. For training of the EGR actuator states, labelled dynamic data of $\tilde{u}_{egr1}$ and $\tilde{u}_{egr2}$ are used, corresponding to inputs $u_{egr}$, $\tilde{u}_{egr1, 0}$ and $\tilde{u}_{egr2, 0}$. Similarly, for the training of VGT actuator states, labelled dynamic data $\tilde{u}_{vgt}$ for inputs $u_{vgt}$ and $\tilde{u}_{vgt,0}$ are used. The loss function for DeepONet training is formulated as:
\begin{subequations}
\begin{align}
    \mathcal{L}_i(\bm{\theta}_{DP,i}) = &\dfrac{1}{n_i}\sum_{j=1}^{n_i}\left[y_i^{(j)} - \hat{y}_i^{(j)} \right]^2 , \;\;\;\;\; i =1, 2 \\
    = & \dfrac{1}{n_i}\sum_{j=1}^{n_i}\left[y_i^{(j)} - \mathcal{D}_i(t,\bm{x}_i;\bm{\theta}_{DP,i})^{(j)} \right]^2,  \;\;\; \bm{x}\rightarrow\text{Branch inputs}
\end{align}
\end{subequations}
where $i = 1,2$ denote the different DeepONets for approximating the EGR and VGT states; $\bm{x}_i$ represents the corresponding input to the $i^{th}$ DeepONet, $\hat{y}_i$ and $y_i$ are the output of the $i^{th}$ DeepONet, and the corresponding labelled data, respectively, and $n_i$ is the number of labelled dataset for the $i^{th}$ DeepONet. 

\par We also note that the DeepONet for the EGR states has two output quantities. Consequently, the loss function for this DeepONet is the weighted summation of the loss function for individual output quantities. These networks are trained using Adam optimizer. As discussed earlier, we utilize simulation data to train the DeepONets and subsequently replace the EGR and VGT dynamics with these pre-trained networks within the PINN model. Note that for DeepONet training here, we do not incorporate dropouts since the outputs of these DeepONets are used as inputs to the pre-trained empirical networks that include dropouts to inherently addresses the variability, as discussed in section \ref{Subsection:Neural network surrogates for empirical formulae}. In the following section, we present details about our hybrid PINN framework to tackle the parameter estimation problem.

The labelled data considered for the training and testing of these DeepONet are listed in Table \ref{Table:Pretrained data}. We divide the problem into a prediction of response of one minute. Thus, the training dataset consists of data of ``Case-I" divided into 120 samples of ${0-60}, {60-120},\dots,$ sec and ``Case-VI", divided into 24 samples of ${0-60}, {60-120},\dots,$ sec. We also consider data augmentation in the training dataset by considering a sliding window of the starting point of the time series. The testing data set ``Case-V" is also divided into 20 samples of ${0-60}, {60-120},\dots,$ sec. By doing this, we convert the problem to an initial value problem of one-minute duration. We consider the input signal $u_{egr}$ to be recorded at equal $dt = 0.2$ sec. The output $\tilde{u}_{egr1}$ and $\tilde{u}_{egr2}$ are also recorded at same $dt = 0.2$ sec. The trunk takes time input $t=0.0,\; 0.20,\; 0.4,\; \dots,\; 60.0$ sec. We know the states are bounded; thus we consider output transformation and scaling so that the outputs are always within the limit. The network sizes and other details are shown in Table \ref{Table:DeepONet input output}.

\par The loss function for individual output variables is the mean square error between the true value ($\Tilde{u}_1$) and the predicted one ($\Tilde{u}_1(\bm{\theta})$). The total loss is the weighted sum of losses for both outputs. For EGR actuator dynamics, we use weights $a_1=1$ and $a_2=100$ (ref. Fig. \ref{Figure:DeepONet EGR}). The DeepONets are trained using the Adam optimizer, with $3\times 10^5$ epochs for DeepONet $\mathcal{D}_1$ and $4\times 10^5$ epochs for DeepONet $\mathcal{D}_2$. Relative $L_2$ errors are as shown in Table \ref{Table:DeepONet results}. The mean and standard deviation of the error are calculated from six independent training sessions and show good accuracy. To further validate the model's robustness, we added a $3\%$ Gaussian noise to the labelled output data and trained the networks with this noisy data. The relative $L_2$ error for these cases is also shown in Table \ref{Table:DeepONet results}, and we note only a marginal change in error, demonstrating the model's robustness to noise in training data.

\begin{landscape}
\begin{table}[H]
\centering
\caption{\textbf{Input output structure of DeepONets for EGR ($\tilde{u}_{egr1}$, $\tilde{u}_{egr1}$) and VGT ($\tilde{u}_{vgt}$) states.} As shown in Fig. \ref{Figure:DeepONet EGR}, the DeepONet ($\mathcal{D}_1$) for EGR states has four branches and two outputs. The DeepONet ($\mathcal{D}_1$) for VGT state has two branches and one output. The input to the trunk is $t=0-60$ sec and scaled between $[-1,1]$. A schematic diagram of the DeepONet is shown in Fig. \ref{Figure:DeepONet EGR}.}
\label{Table:DeepONet input output}
\begin{tabular}{c|c|c|c|c|c|c} \hline
 Trunk & Branch-1 & Branch-2 & Branch-3 & Branch-4 & \multicolumn{2}{c}{Output} \\ \hline
 \multicolumn{7}{c}{\textbf{For EGR states $\tilde{u}_{egr1}$ and $\tilde{u}_{egr2}$}} \\ \hline 
$\begin{matrix} 
    t=t_0=0.00\\
    t=t_1=0.02\\
    \cdot \\
    \cdot \\
    t=t_n = 60.0
\end{matrix}$ & 
$\begin{matrix} 
    u_{degr,0} & 0 & 0 & \dots & 0 \\
    u_{degr,1} & u_{degr,0} & 0 & \dots & 0 \\
    \cdot & \cdot & \cdot & \cdots & \cdot \\
    \cdot & \cdot & \cdot & \cdots & \cdot \\
    u_{degr,n} & u_{degr,n-1} & u_{degr,n-2} & \dots & u_{degr,0} \\
\end{matrix}$ &
$\begin{matrix} 
    \tilde{u}_{egr1,0}\\
    \tilde{u}_{egr1,0}\\
    \cdot \\
    \cdot \\
    \tilde{u}_{egr1,0}
\end{matrix}$ &
$\begin{matrix} 
    u_{degr,0} & 0 & 0 & \dots & 0 \\
    u_{degr,1} & u_{degr,0} & 0 & \dots & 0 \\
    \cdot & \cdot & \cdot & \cdots & \cdot \\
    \cdot & \cdot & \cdot & \cdots & \cdot \\
    u_{degr,n} & u_{degr,n-1} & u_{degr,n-2} & \dots & u_{degr,0} \\
\end{matrix}$ &
$\begin{matrix} 
     \tilde{u}_{egr2,0}\\
    \tilde{u}_{egr2,0}\\
    \cdot \\
    \cdot \\
    \tilde{u}_{egr2,0}
\end{matrix}$ &
$\begin{matrix} 
    \tilde{u}_{egr1,0}\\
    \tilde{u}_{egr1,1}\\
    \cdot \\
    \cdot \\
    \tilde{u}_{egr1,n}
\end{matrix}$ & 
$\begin{matrix} 
    \tilde{u}_{egr2,0}\\
    \tilde{u}_{egr2,1}\\
    \cdot \\
    \cdot \\
    \tilde{u}_{egr2,n}
\end{matrix}$ \\ \hline 
\multicolumn{7}{c}{\textbf{For VGT states $\tilde{u}_{vgt}$}} \\ \hline
$\begin{matrix} 
    t=t_0=0.00\\
    t=t_1=0.02\\
    \cdot \\
    \cdot \\
    t=t_n = 60.0
\end{matrix}$ & 
$\begin{matrix} 
    u_{dvgt,0} & 0 & 0 & \dots & 0 \\
    u_{dvgt,1} & u_{dvgt,0} & 0 & \dots & 0 \\
    \cdot & \cdot & \cdot & \cdots & \cdot \\
    \cdot & \cdot & \cdot & \cdots & \cdot \\
    u_{dvgt,n} & u_{dvgt,n-1} & u_{dvgt,n-2} & \dots & u_{dvgt,0} \\
\end{matrix}$ &
$\begin{matrix} 
    \tilde{u}_{vgt,0}\\
    \tilde{u}_{vgt,0}\\
    \cdot \\
    \cdot \\
    \tilde{u}_{vgt,0}
\end{matrix}$ &
NA &
NA &
\multicolumn{2}{c}{$\begin{matrix} 
    \tilde{u}_{vgt,0}\\
    \tilde{u}_{vgt,1}\\
    \cdot \\
    \cdot \\
    \tilde{u}_{vgt,n}
\end{matrix}$}  
\\ \hline 
\end{tabular}
\end{table}

\begin{table}[H]
\centering
\caption{\textbf{DeepONet results for independent variables:} The training and testing errors for DeepONet for EGR and VGT states. The two DeepONets are trained using clean data and with added Gaussian noise. The relative $L_2$ error (\%) for training and testing are shown in the fourth and sixth columns, respectively. The bottom part of the table gives the $L_2$ error (\%) in training and testing when Gaussian noise is added to the training labelled data. We also note that the error in the case of noisy data is calculated with respect to the clean data. The mean and standard deviation of the error are calculated based on six different independent runs.}
\label{Table:DeepONet results}
\begin{tabular}{c|c|c|c|c|c} \hline
\multirow{2}{*}{DeepONet} & Output &  \multicolumn{2}{c|}{Training}  &  \multicolumn{2}{c}{Testing} \\ \cline{3-6}
& Variable & Data type & Rel. $L_2$ error (\%) & Data type & Rel. $L_2$ error (\%) \\ \hline

\multirow{2}{*}{$\mathcal{D}_1$} & $\tilde{u}_{egr1}$ & Clean data & $0.64\pm 0.04$ & Clean data & $2.69\pm 0.49$  \\ \cline{2-6}

& $\tilde{u}_{egr2}$ & Clean data & $0.65\pm 0.04$ & Clean data & $3.64\pm 0.18$  \\ \hline

$\mathcal{D}_2$&  $\tilde{u}_{vgt}$ & Clean data & $0.86\pm0.10$ & Clean data & $2.52\pm0.08$  \\ \hline

\multicolumn{6}{c}{with added noise in output during training} \\ \hline
\multirow{2}{*}{$\mathcal{D}_1$} & $\tilde{u}_{egr1}$ & 3\% noise added & $0.95\pm0.08$ & Clear data & $3.11\pm0.75$ \\ \cline{2-6}

& $\tilde{u}_{egr2}$ & 3\% noise added & $0.97\pm0.02$ & Clear data & $3.48\pm0.17$  \\ \hline

$\mathcal{D}_2$ & $\tilde{u}_{vgt}$ & 3\% noise added & $1.14\pm 0.11$ & Clear data & $2.62\pm 0.13$  \\ \hline

\end{tabular}
\end{table}
\end{landscape}

\section{Additional results for Multi-stage transfer learning prediction} 
\label{sec:additional_plots_MTL}
\begin{figure}[H]
    \centering
    \includegraphics[width=0.9\textwidth]{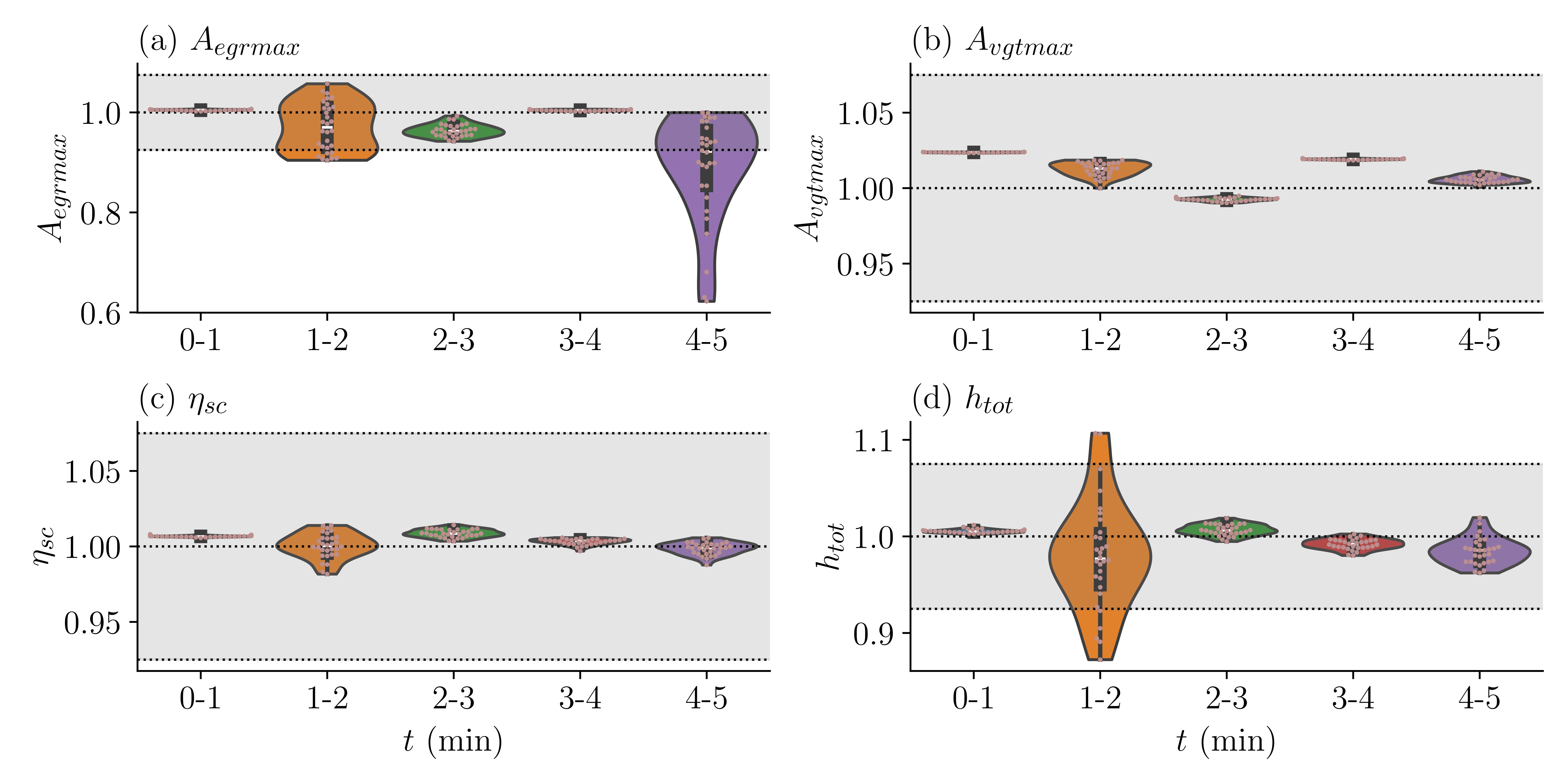}
    \caption{\textbf{Violin plot, multi-stage TL, clean data, 301 data points:} variation in engine parameter predictions over 30 runs with clean data and 301 data points. More data points help reduce the variation in estimation across all data segments as compared to the model trained with 151 data points} 
    \label{fig:TF_violin_clean301}
\end{figure}

\begin{figure}[H]
    \centering
    \includegraphics[width=0.8\textwidth]{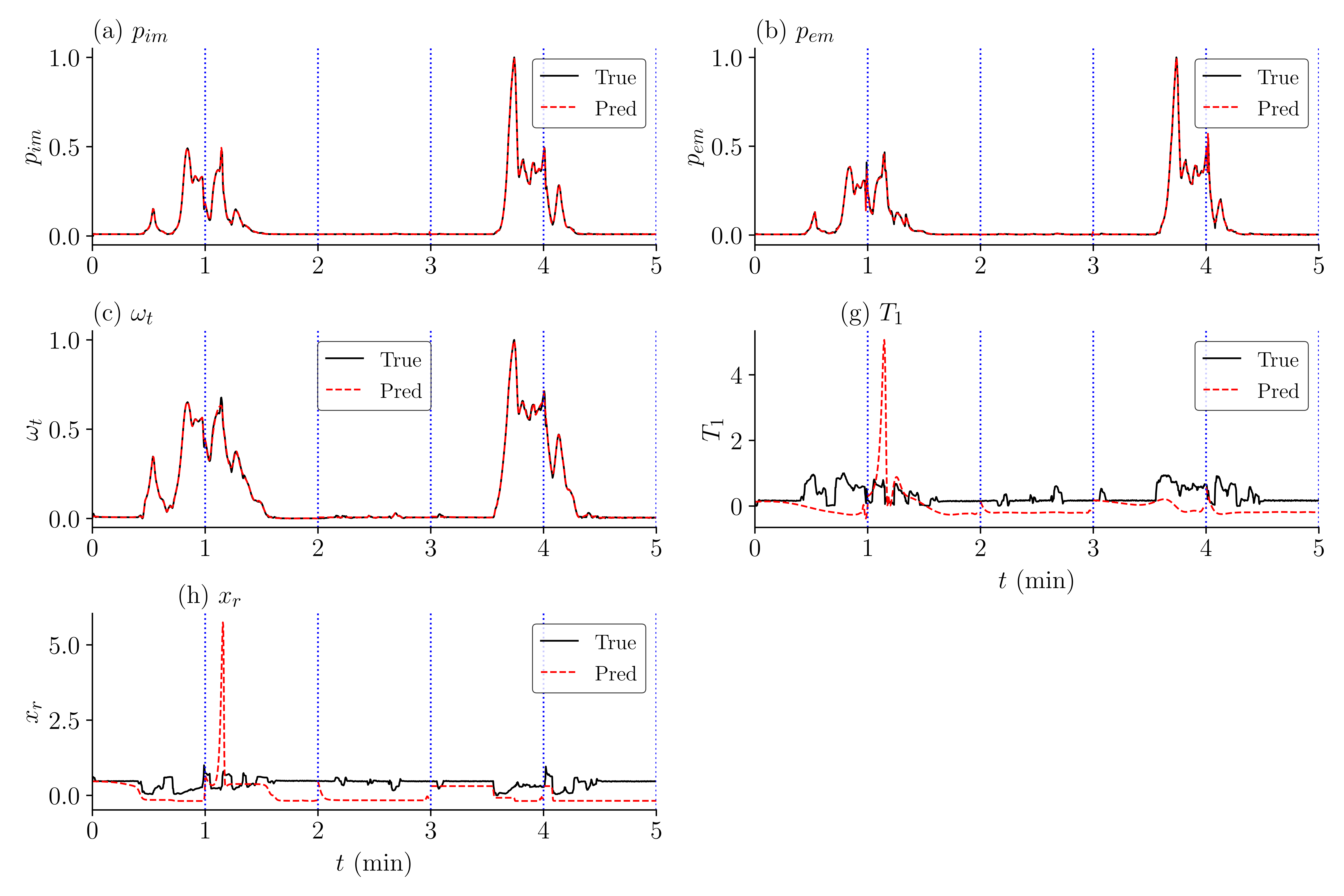}
    \caption{\textbf{Predicted states, multi-stage TL, clean data, 151 data points:} Primary gas flow dynamics states from multi-stage transfer learning model (data scale normalized due to confidentiality). The plot shows a representative case where clean data is used with 151 data points for online training. Note that states $\tilde u_{egr1}, \tilde u_{egr2},$ and $\tilde u_{vgt}$ are outputs obtained from pre-trained DeepONet and are not trained during the transfer learning process.}
    \label{fig:TF_states_S151}
\end{figure}

\begin{figure}[H]
    \centering
    \includegraphics[width=0.9\textwidth]{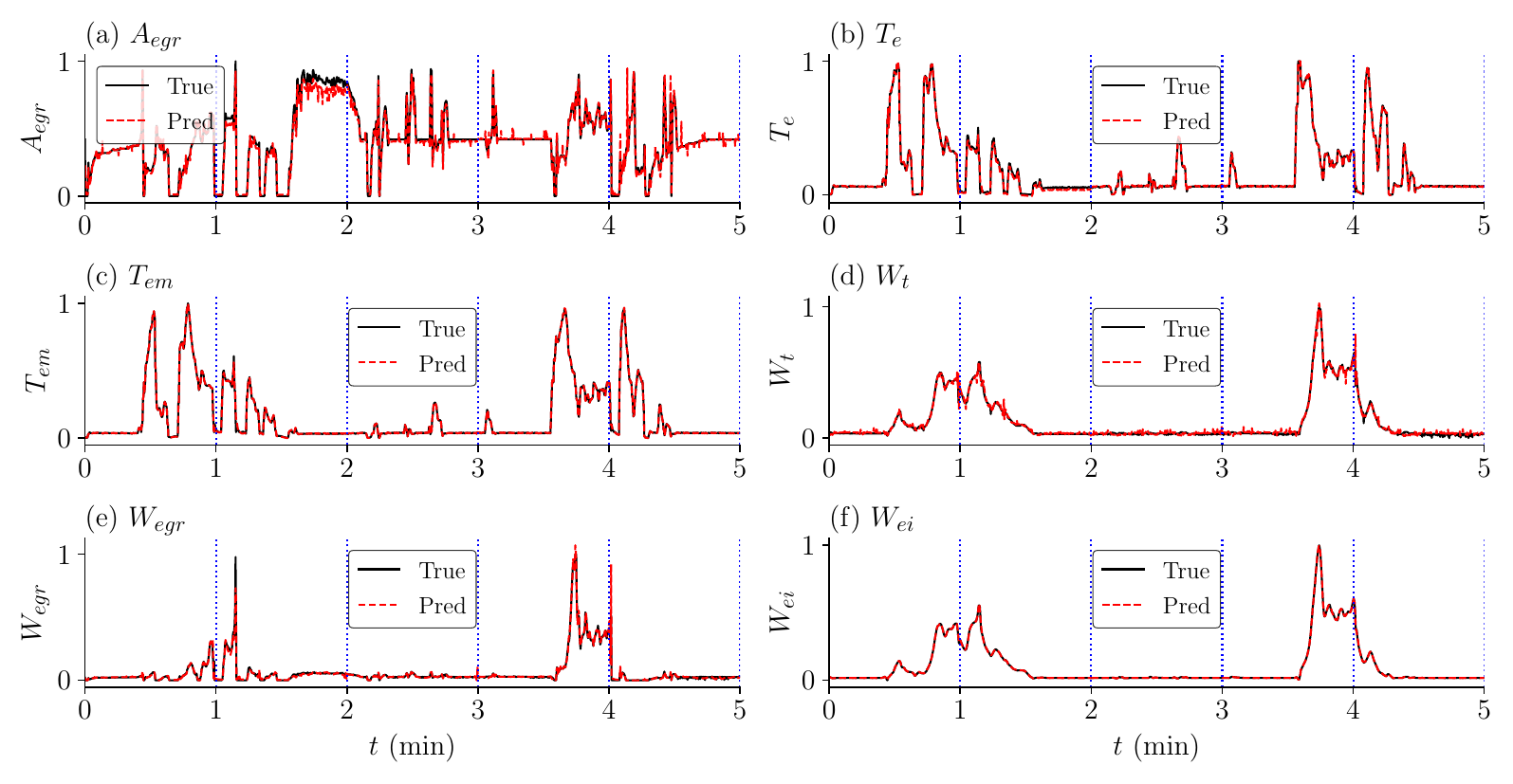}
    \caption{\textbf{Gas flow rates, multi-stage transfer learning, clean data, 151 data points:} Gas flow rates from transfer learning model. The plot shows a representative case where clean data is used with 151 data points for online training.}
    \label{fig:TF_flows_S151}
\end{figure}

\begin{figure}[H]
    \centering
    \includegraphics[width=0.9\textwidth]{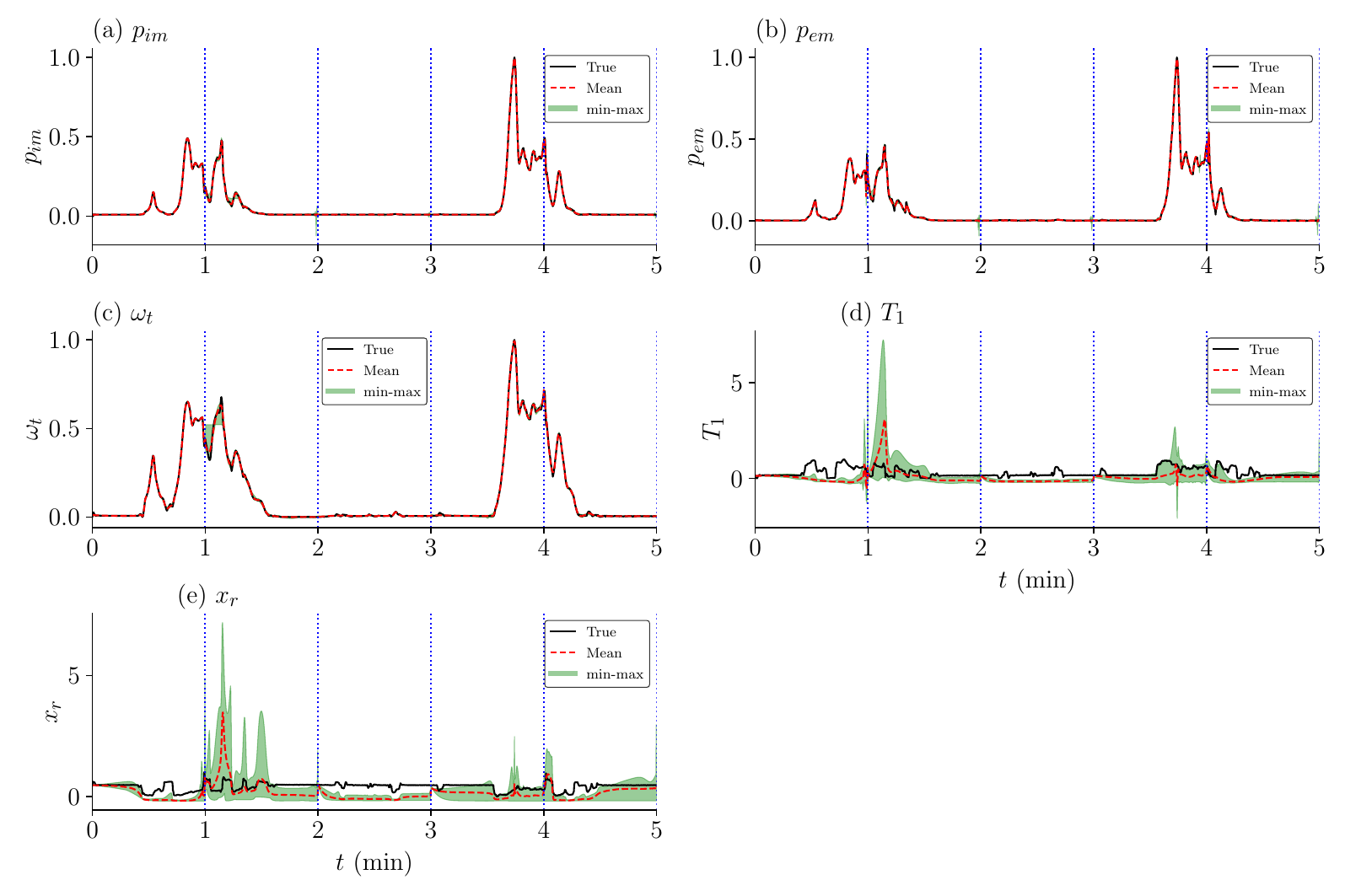}
    \caption{\textbf{Predicted dynamic states, multi-stage TL, clean data, 151 data points, ensemble:} Ensemble prediction results for primary gas flow dynamics states from transfer learning model (data scale normalized due to confidentiality). The plot shown is for scenario where clean data is used with 151 data points used for online training. The ensemble mean (red) for the 30 runs is plotted against ground truth (black) for the eight dynamic states defined by ODE's describing the gas flow dynamics. Error bands (shown in light green) indicate the spread of data between point-wise maximum and minimum values across all 30 runs. }    
    \label{fig:TF_ensmstate_S151}
\end{figure}

\begin{figure}[H]
    \centering
    \includegraphics[width=0.9\textwidth]{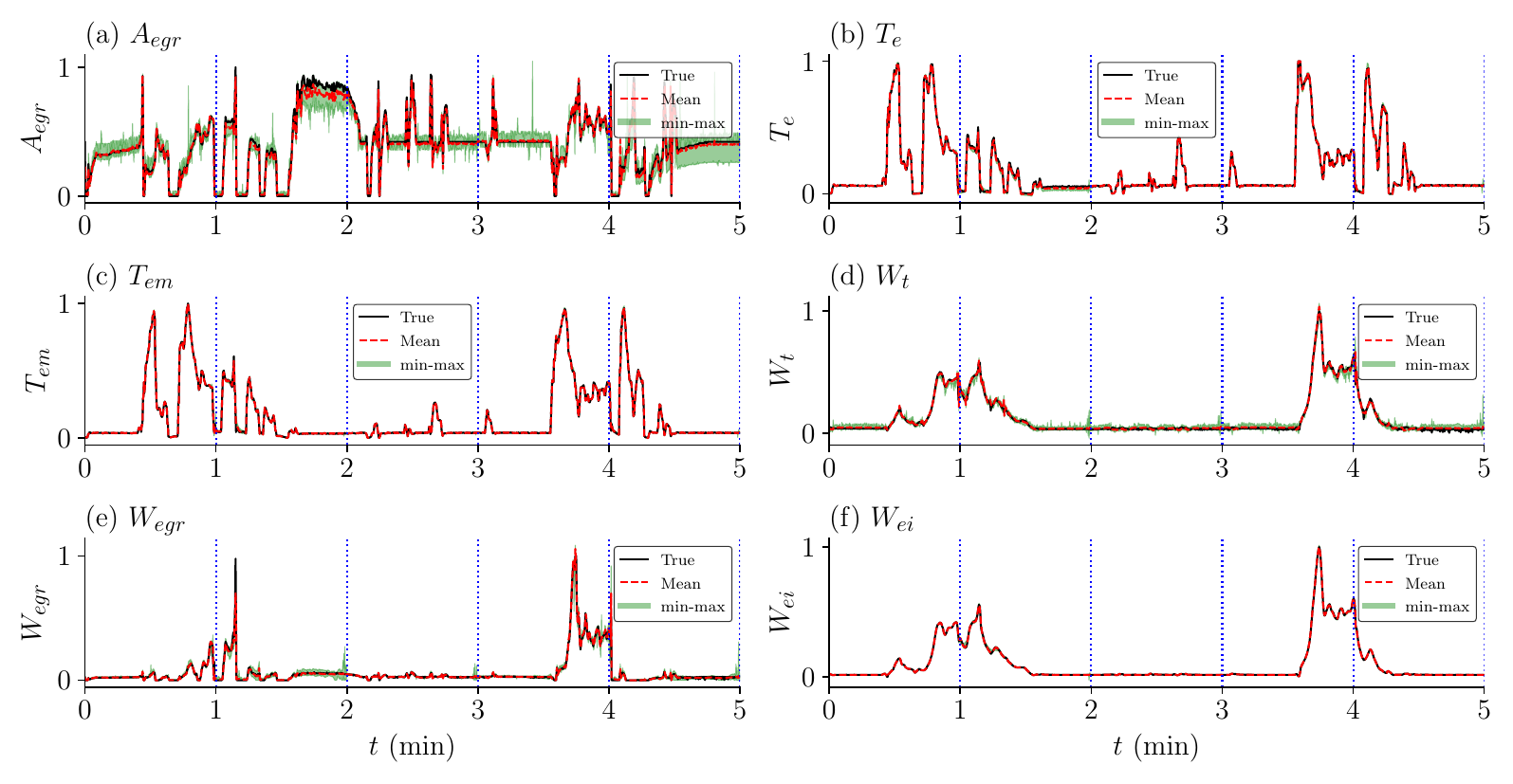}
    \caption{\textbf{Predicted gas flow, multi-stage TL, clean data, 151 data points, ensemble:} Ensemble prediction results for gas flow rates from the Transfer learning model with clean data over 30 independent runs across all segments. 151 data points were used during online parameter estimation task.}
    \label{fig:TF_ensmflow_S151}
\end{figure}

\begin{figure}[H]
    \centering
    \includegraphics[width=0.9\textwidth]{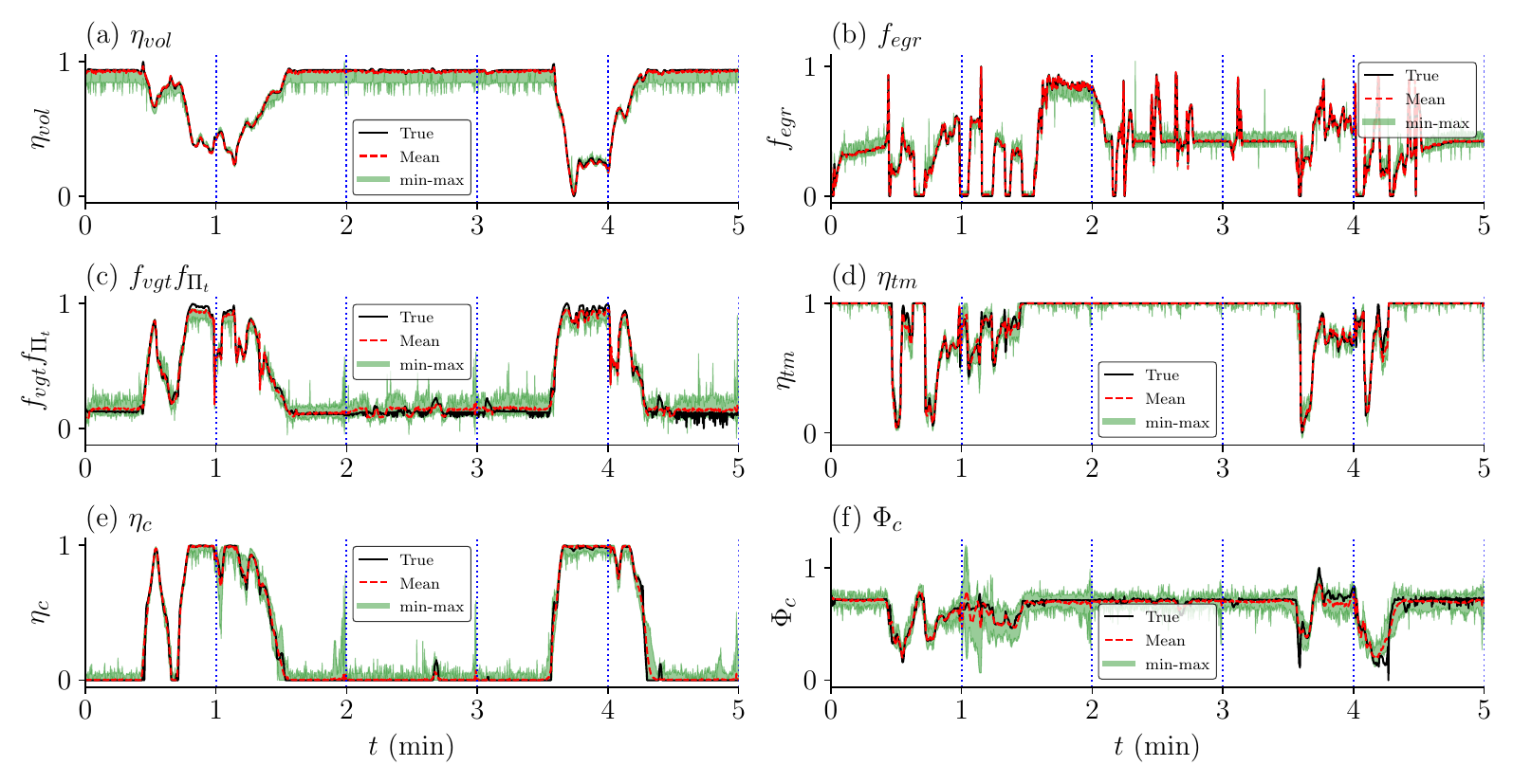}
    \caption{\textbf{Predicted empirical variables, multi-stage TL, clean data, 151 data points, ensemble:} Ensemble prediction estimates for empirical relationships for clean data with Transfer learning model. Note that in region $\approx 1.5 - 3.5 \, min$, the model's prediction indicates a spectral bias in learning the high frequency signals that typically result from engine idling. Elsewhere, the model predictions show good match with target values for these empirical relationships.}
    \label{fig:TF_ensmemp_S151}
\end{figure}

\begin{figure}[H]
    \centering
    \includegraphics[width=0.8\textwidth]{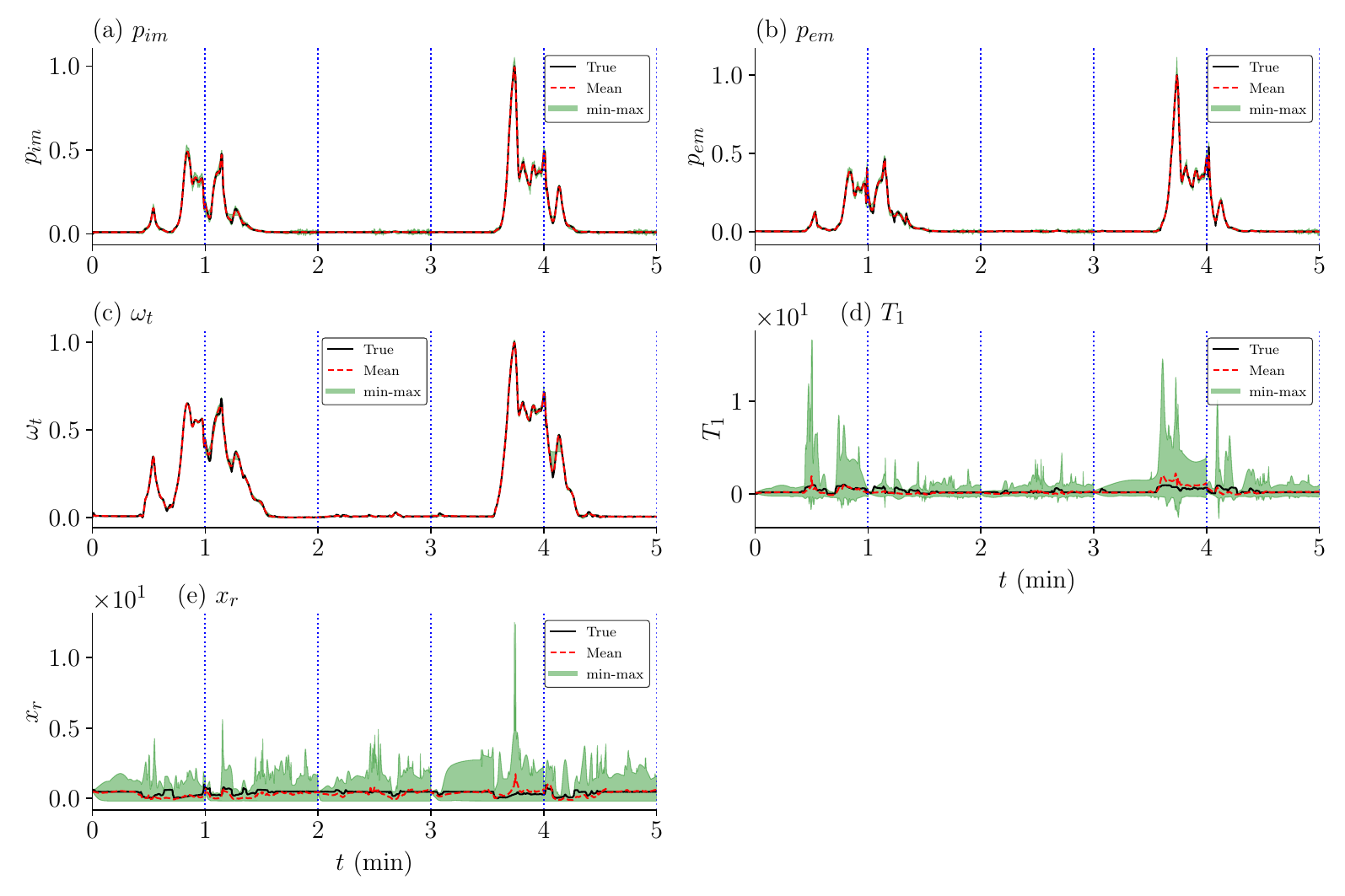}
    \caption{\textbf{Predicted dynamic states, multi-stage TL, noisy data, 301 data points, ensemble:} Ensemble prediction results for primary gas flow dynamics states from Transfer learning model (data scale normalized due to confidentiality). The plot shown is for scenario where noisy data is used with 301 data points used for online training. The ensemble mean (red) for the 30 runs is plotted against ground truth (black) for the eight dynamic states defined by ODE's describing the gas flow dynamics.}       
    \label{fig:TF_ensmstate_S301_noisy}
\end{figure}

\begin{figure}[H]
    \centering
    \includegraphics[width=0.9\textwidth]{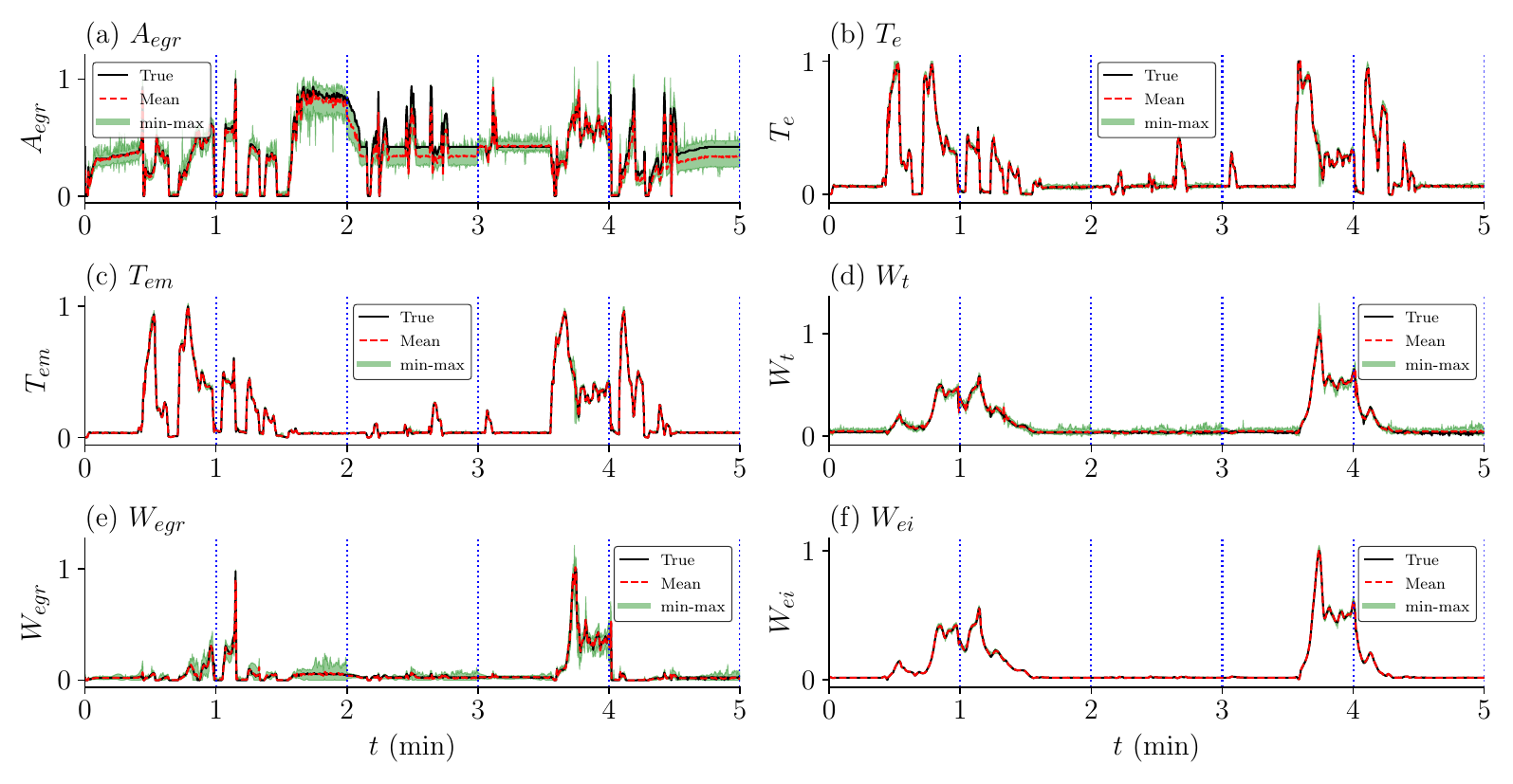}
    \caption{\textbf{Predicted gas flow rate, multi-stage TL, noisy data, 301 data points, ensemble:} Ensemble prediction results for gas flow rates from the Transfer learning model with noisy data. 301 data points were used during online parameter estimation task.}
    \label{fig:TF_ensmflow_S301_noisy}
\end{figure}

\begin{figure}[H]
    \centering
    \includegraphics[width=0.9\textwidth]{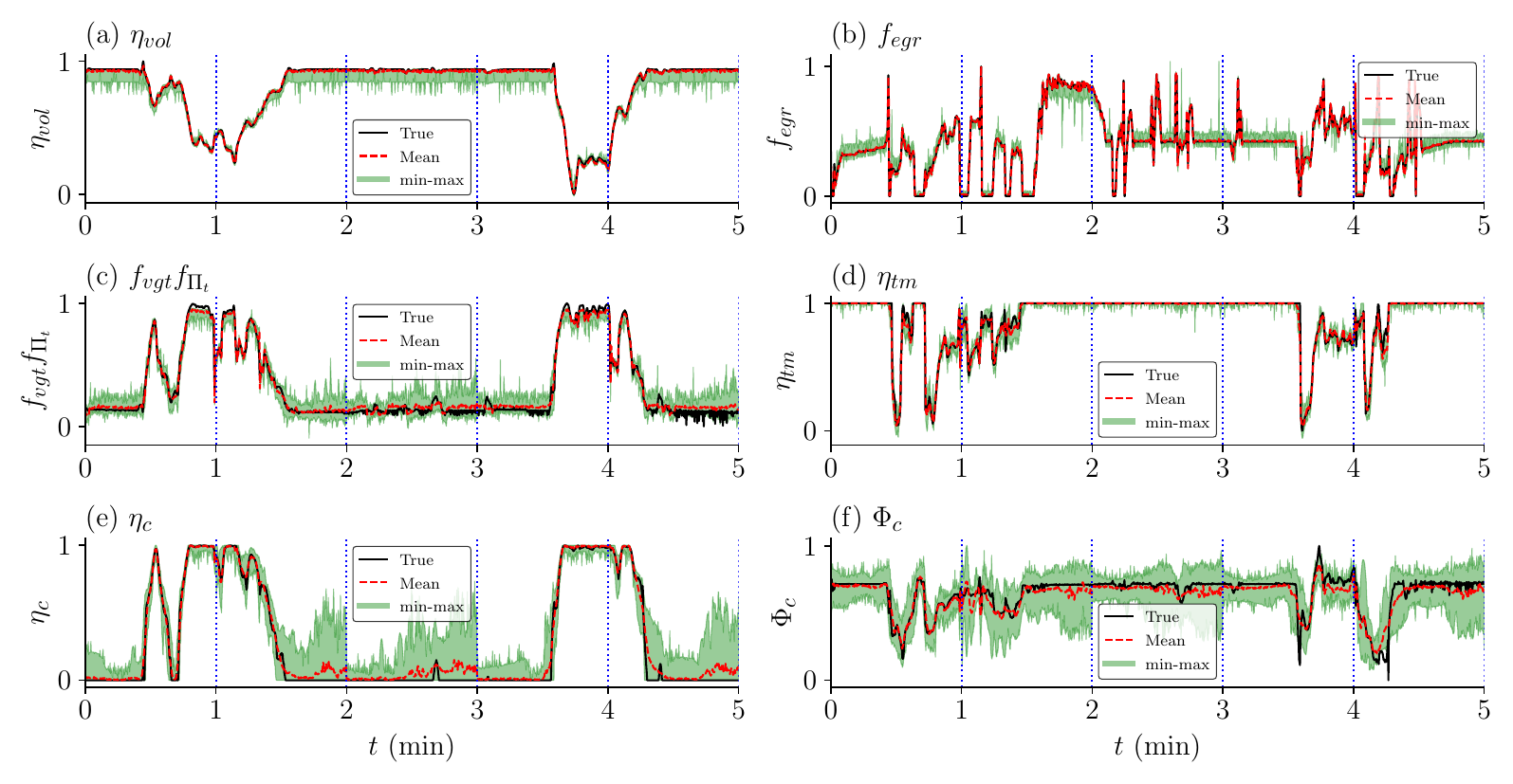}
    \caption{\textbf{Predicted empirical variable, multi-stage TL, noisy data, 301 data points, ensemble:} Ensemble prediction estimates for empirical relationships for noisy data with transfer learning model with 301 data points. Note that in region $\approx 1.5 - 3.5 \, min$, the model's prediction indicate a spectral bias in learning the high frequency signals that typically result from engine idling. Also, there is a significant increase in expected variability in prediction of these empirical states under noisy conditions as compared to clean conditions shown in figure \ref{fig:TF_ensmemp_S151}, correlating with increase in variability in prediction of other states and parameters as well.}
    \label{fig:TF_ensmemp_S301_noisy}
\end{figure}

In this section, we provide additional results showing the predictions for different engine related Quantities of Interest (QOIs). Fig. \ref{fig:TF_violin_clean301} show the variation in parameter estimation over 30 independent runs with 301 data points under clean conditions. Adding residual points help in reducing the variation in estimates in some scenarios. This, however comes at an additional computational cost. Higher variation is observed in time segment 1-2 minute for all parameters likely due the engine entering idling condition from a dynamical state which induces bias in PINN training. Figure \ref{fig:TF_states_S151} and \ref{fig:TF_flows_S151} show prediction results for main state estimates and flow rates obtained from one instance of the model run. Additional Figures \ref{fig:TF_ensmstate_S151} - \ref{fig:TF_ensmemp_S301_noisy} show the ensemble plot for different predictions over the 30 independent runs of the model for both clean and noisy test conditions. We notice higher variation in predictions with noisy data for all dynamic estimates. Overall, the mean estimated values are reasonably close to the ground truth values.

\section{Additional results for Few shot transfer learning} 
\label{Appendix:Additional figure for few-shot transfer learning}
\begin{figure}[H]
    \centering
    \includegraphics[width=0.9\textwidth]{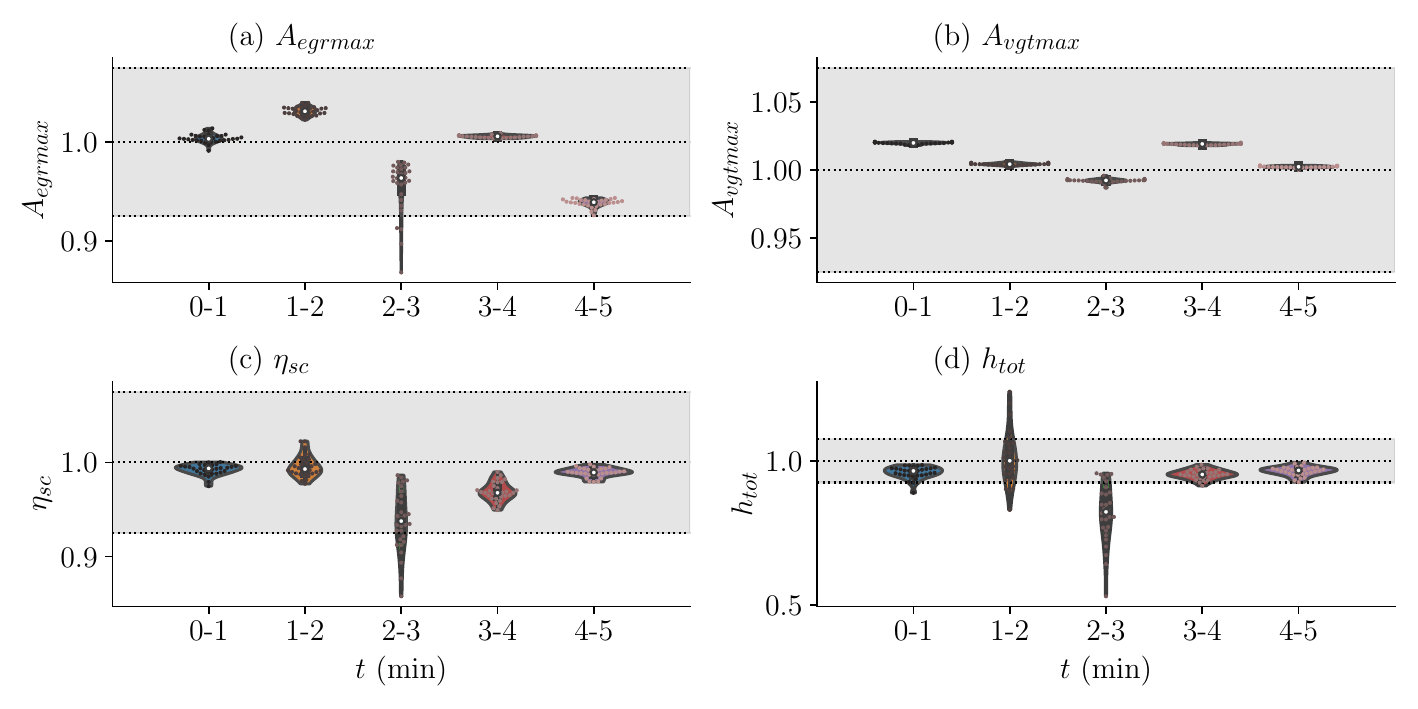}
    \caption{\textbf{Violin plot, few-shot learning, clean data, 301 data points:} of the predicted value of the unknown parameters. 301 data points are considered for each of the known parameters (for each one minute segment) and no noise is considered here.}
    \label{Figure:Few shot:Violin plot 301 clean data}
\end{figure}

\begin{figure}[H]
    \centering
    \includegraphics[width=0.9\textwidth]{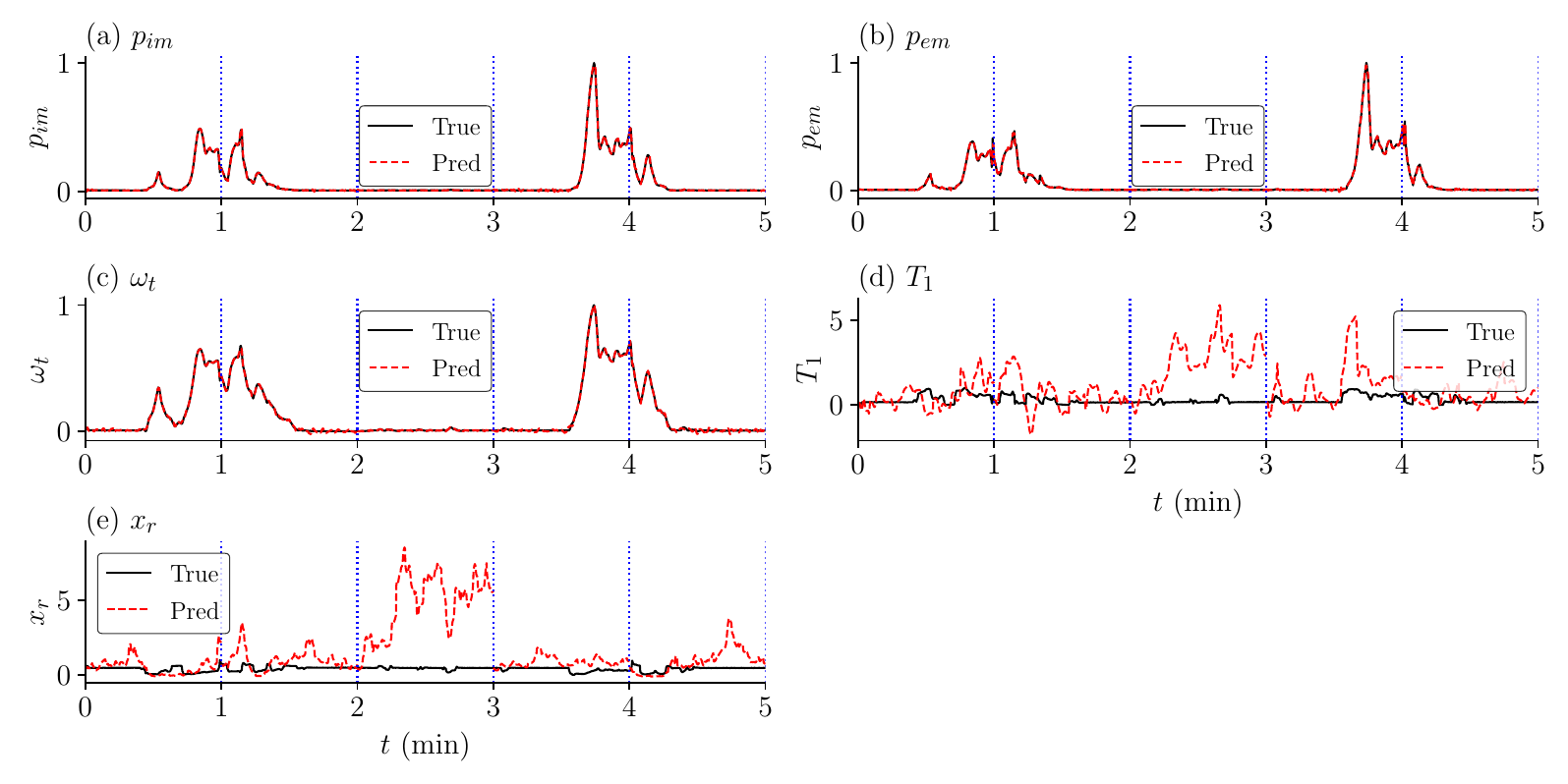}
    \caption{\textbf{Predicted states, Few-shot learning, Clean data, 151 data points:} One of the predicted realization of the state variables and $x_r$ and $T_1$, for the case where clean data and 151 data points are considered for each known variables (for each one minute duration). Each 1 minute is trained independently and combined to get the total 5 minutes results. The vertical dotted blue lines separate each minute result. The results are normalized using Eq. \eqref{Eq:Scale for results} with maximum and minimum true value calculated over the 5 minute.}
    \label{Figure: States: Few shot clean data 151 points}
\end{figure}

\begin{figure}[H]
    \centering
    \includegraphics[width=0.9\textwidth]{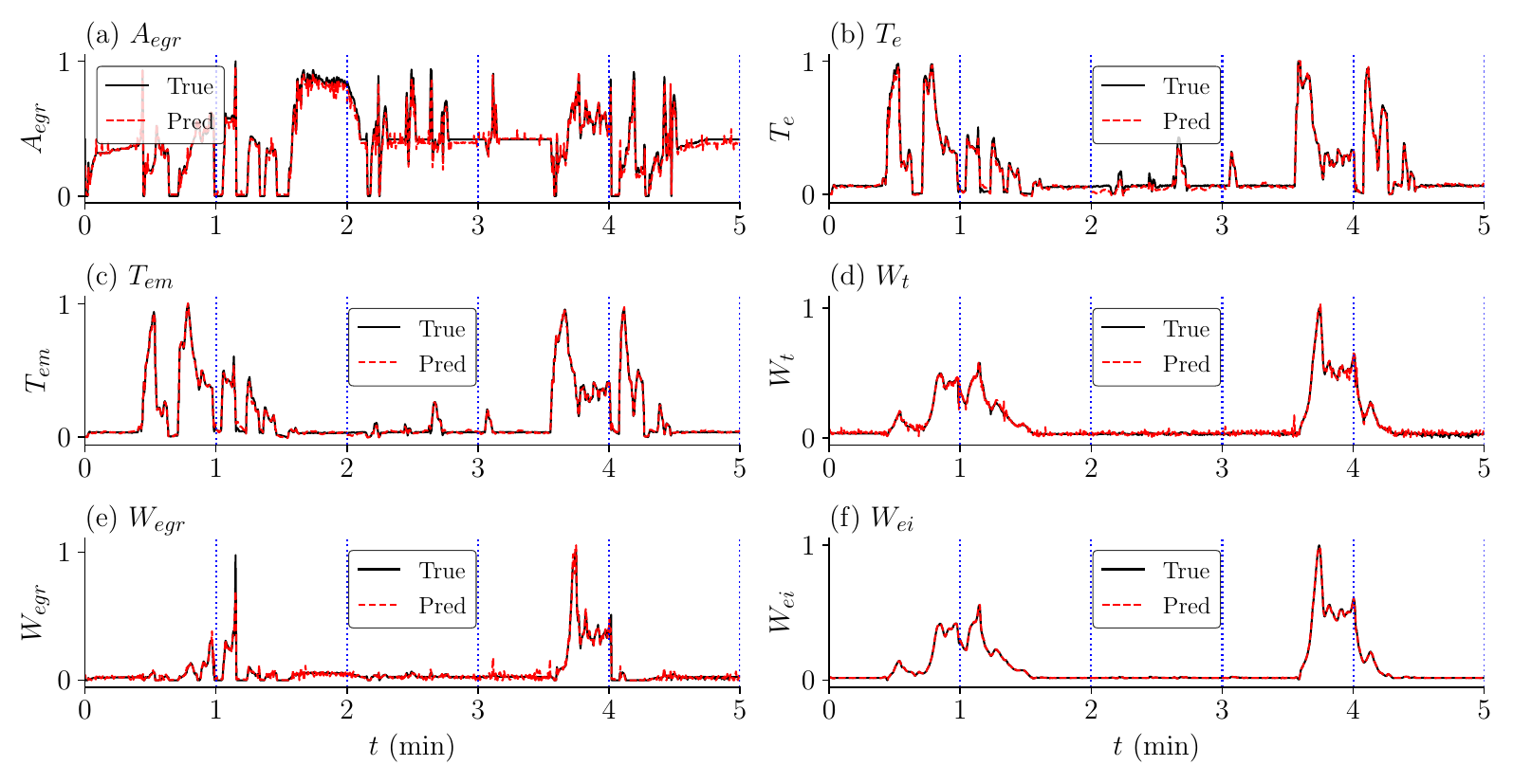}
    \caption{\textbf{Intermediate variables, Few-shot learning, Clean data, 151 data points:} One of the predicted realization of different variables of the diesel engine for the case where clean data and 151 data points are considered for each known variable (for each one minute duration). The variable $A_{egr}$, $T_{e}$, $T_{em}$ and $W_t$ are the variable which are directly depends on the unknown parameters $A_{egrmax}$, $\eta_{sc}$, $h_{tot}$ and $A_{vgtmax}$. The variable $T_e$ also depends on the variable $T_1$ and $x_r$. }
    \label{Figure: Other variable: Few shot clean data 151 points}
\end{figure}

\begin{figure}[H]
    \centering
    \includegraphics[width=0.9\textwidth]{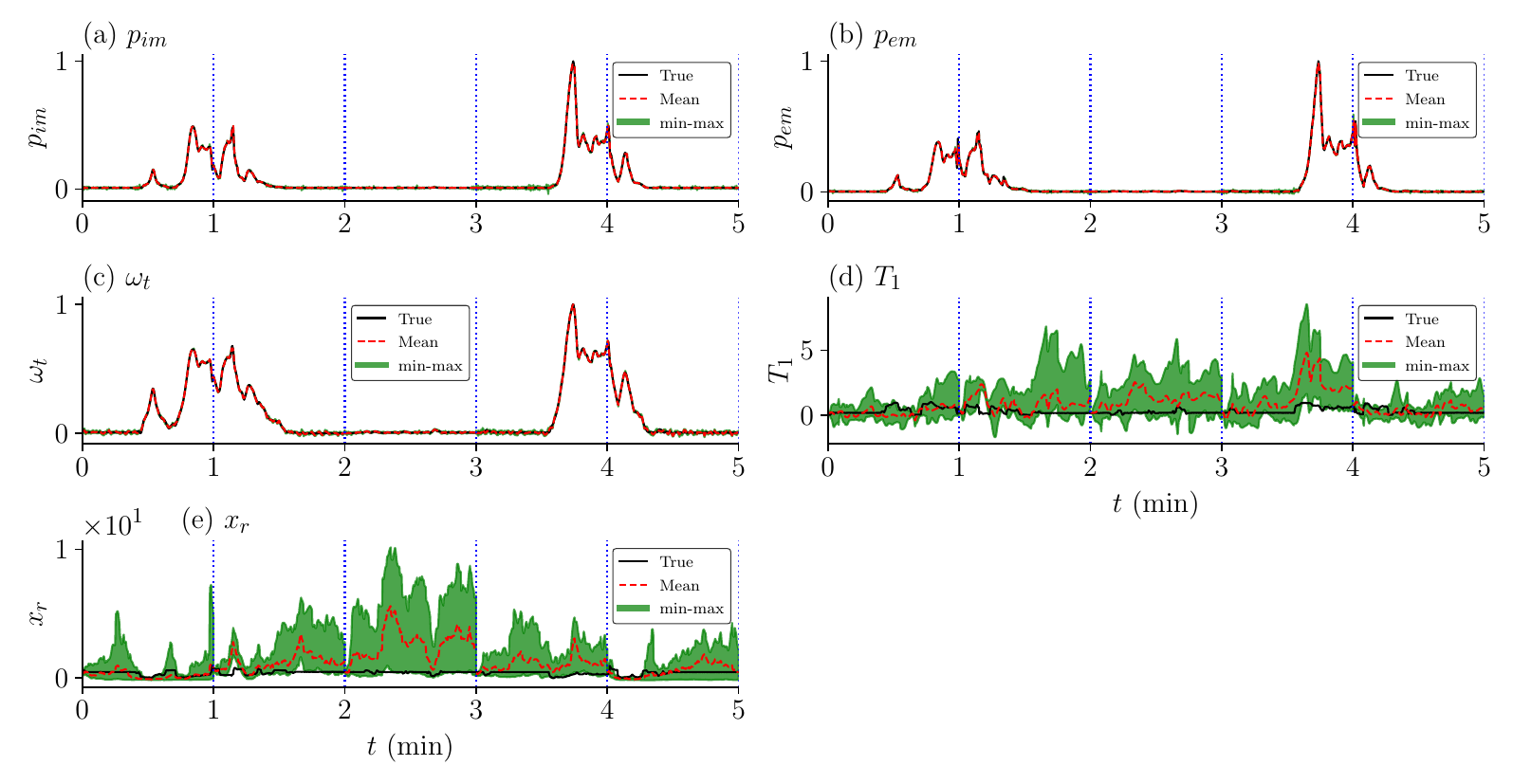}
    \caption{\textbf{Predicted states, Few-shot learning, Clean data, 151 data points, ensemble:} Ensemble prediction for state variable and $x_r$ and $T_1$ for clean data with 151 data points in the data loss predicted using few-shot transfer learning. The ensembles are calculated over 30 independent runs for each time segment.}
    \label{Figure: States: Few shot clean data 151 points: ensemble}
\end{figure}

\begin{figure}[H]
    \centering
    \includegraphics[width=0.9\textwidth]{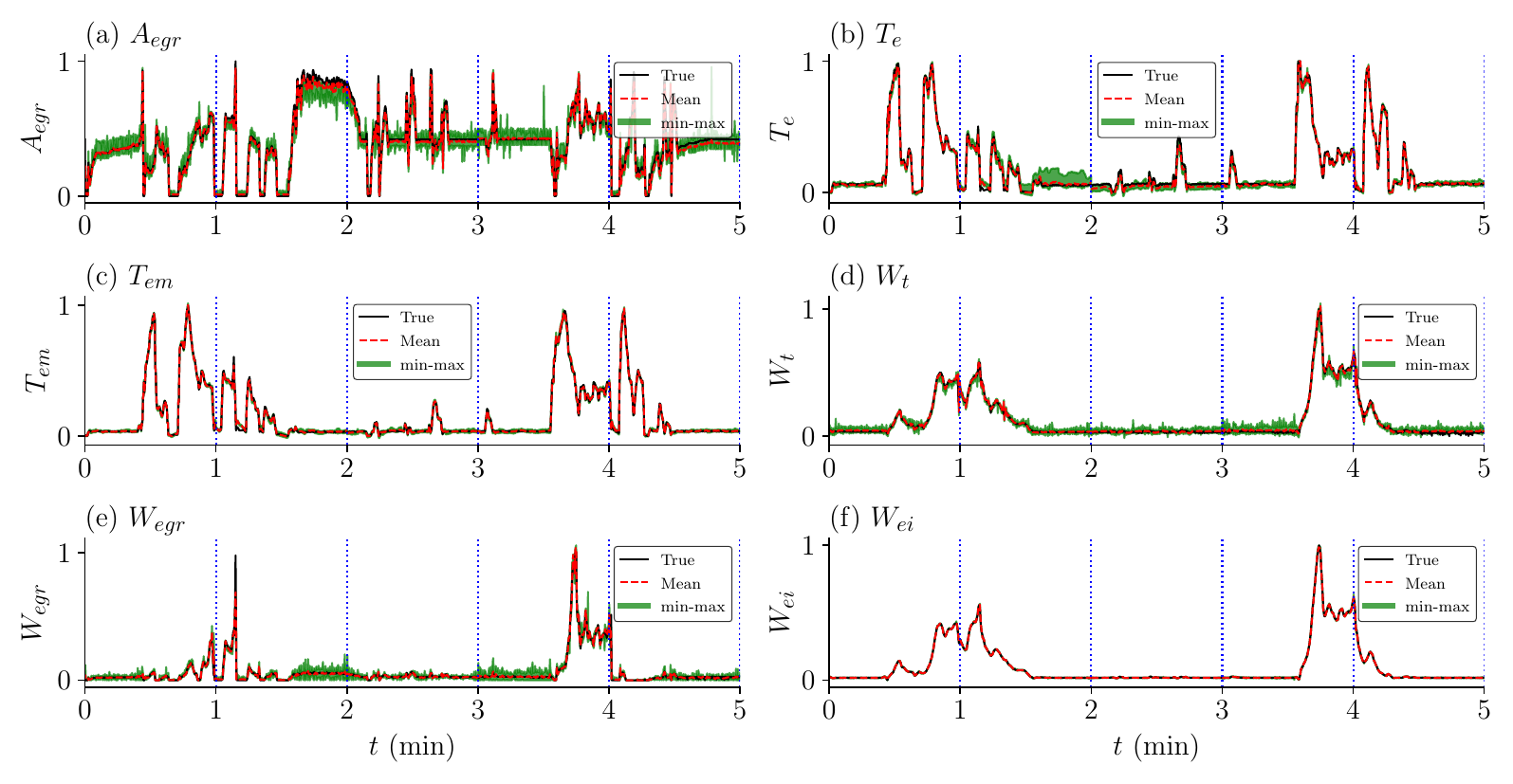}
    \caption{\textbf{Predicted states, Few-shot learning, Clean data, 151 data points, ensemble:} Ensemble prediction for different variables for clean data with 151 data points in the data loss predicted using few-shot transfer learning. The ensembles are calculated over 30 independent runs for each time segment.}
    \label{Figure: flow: Few shot clean data 151 points: ensemble}
\end{figure}

\begin{figure}[H]
    \centering
    \includegraphics[width=0.9\textwidth]{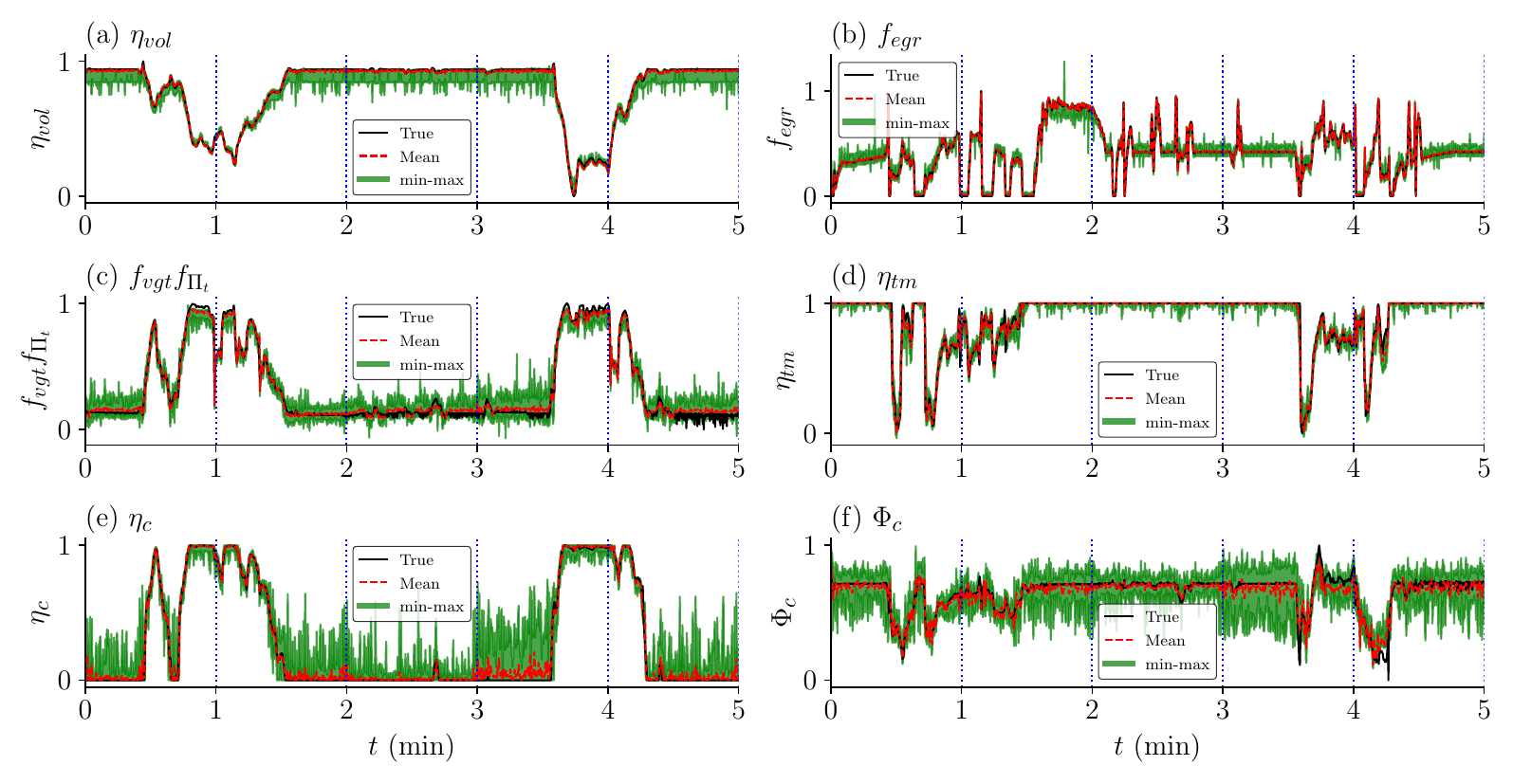}
    \caption{\textbf{Predicted empirical formulae, Few-shot learning, Clean data, 151 data points, ensemble:} Ensemble prediction for the empirical formulas for clean data with 151 data points in the data loss predicted using few-shot transfer learning. The ensembles are calculated over 30 independent runs for each time segment.}
    \label{Figure: empirical: Few shot clean data 151 points: ensemble}
\end{figure}

\begin{figure}[H]
    \centering
    \includegraphics[width=0.9\textwidth]{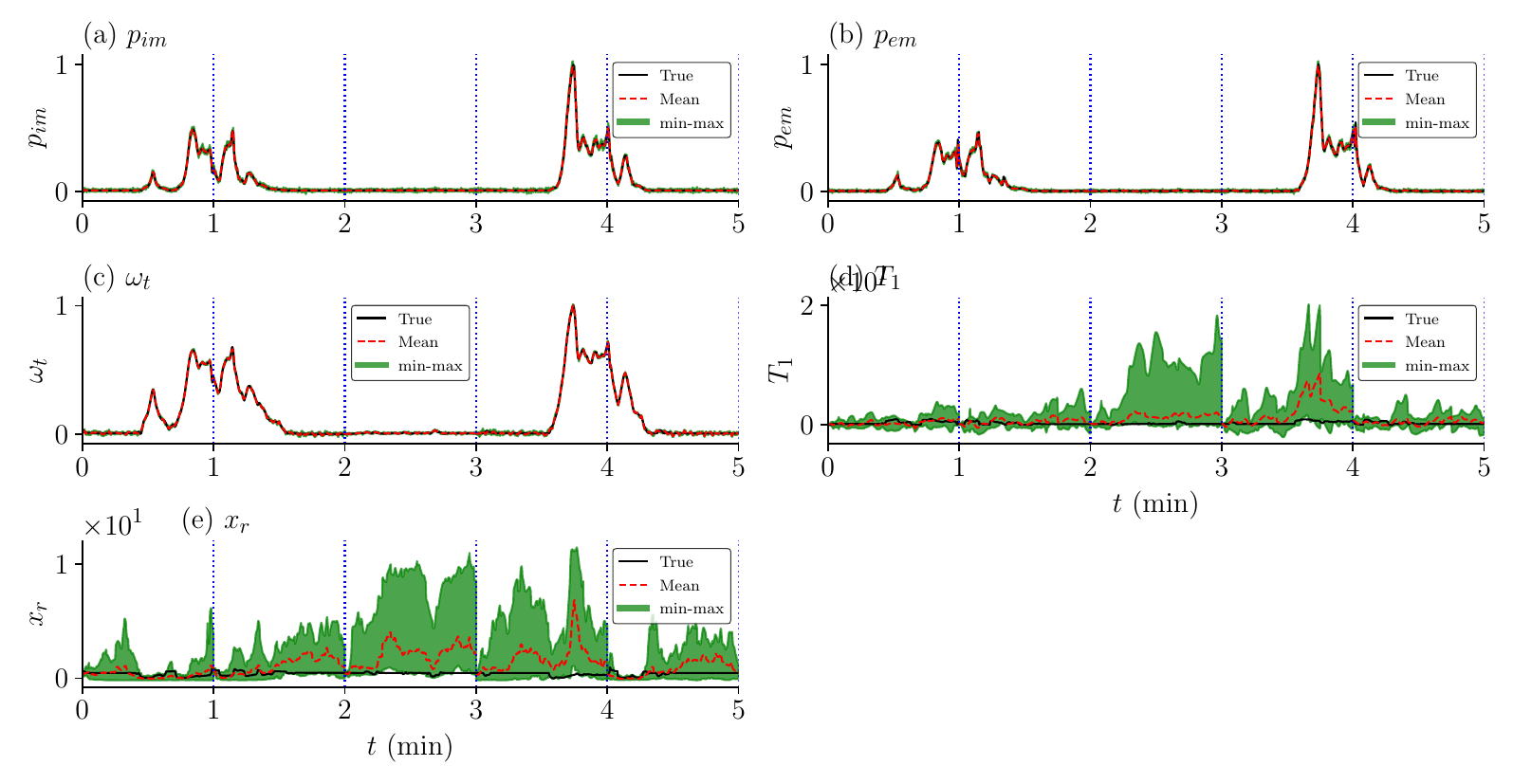}
    \caption{\textbf{Predicted states, Few-shot learning, Noisy data, 301 data points, ensemble:} Ensemble prediction for state variable and $x_r$ and $T_1$ for noisy data with 301 data points in the data loss predicted using few-shot transfer learning. The ensembles are calculated over 30 independent runs for each time segment.}
    \label{Figure: States: Few shot Noisy data 301 points: ensemble}
\end{figure}

\begin{figure}[H]
    \centering
    \includegraphics[width=0.9\textwidth]{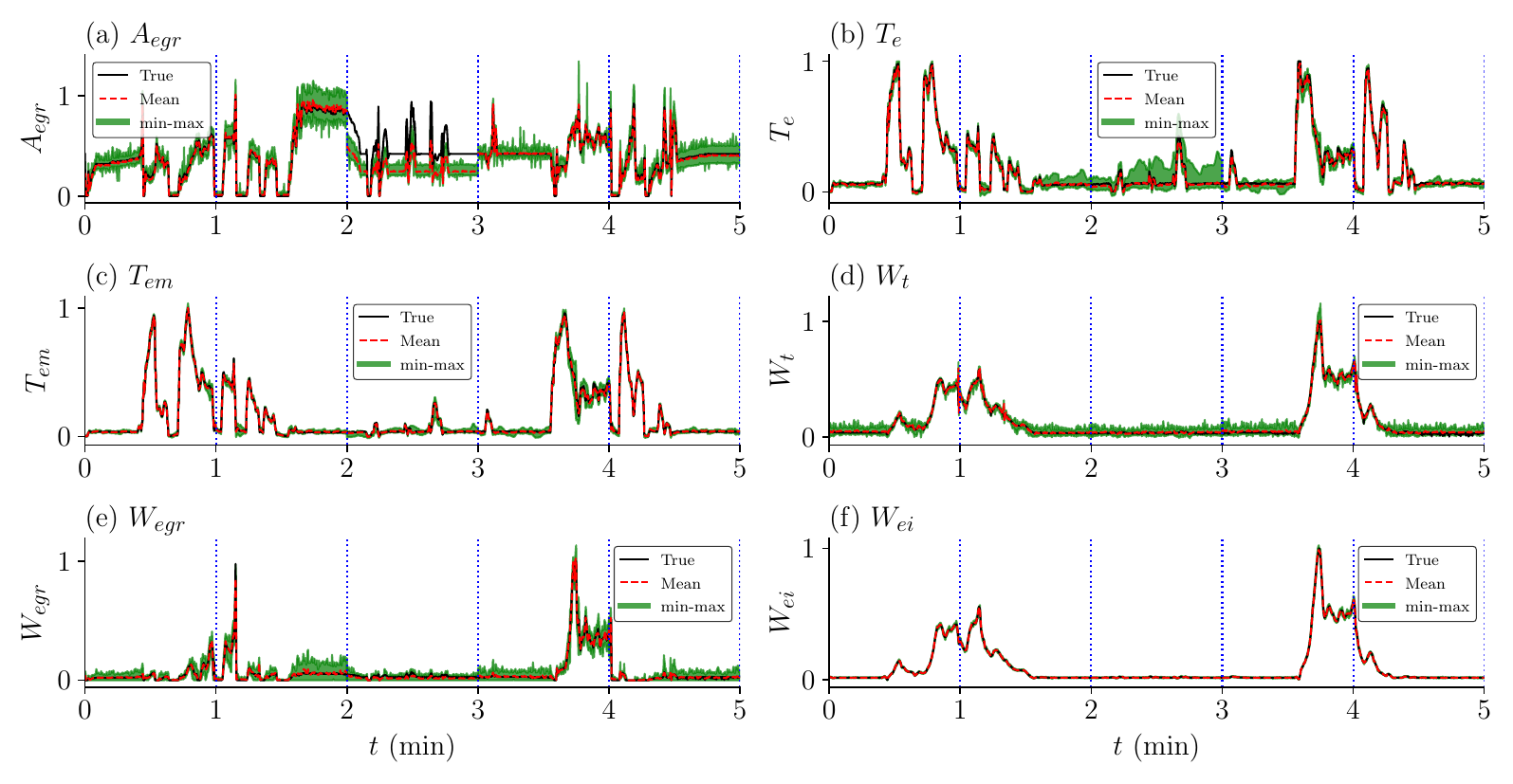}
    \caption{\textbf{Intermediate variables, Few-shot learning, Noisy data, 301 data points, ensemble:} Ensemble prediction for different variables for noisy data with 301 data points in the data loss predicted using few-shot transfer learning. The ensembles are calculated over 30 independent runs for each time segment.}
    \label{Figure: Flow: Few shot Noisy data 301 points: ensemble}
\end{figure}

\begin{figure}[H]
    \centering
    \includegraphics[width=0.9\textwidth]{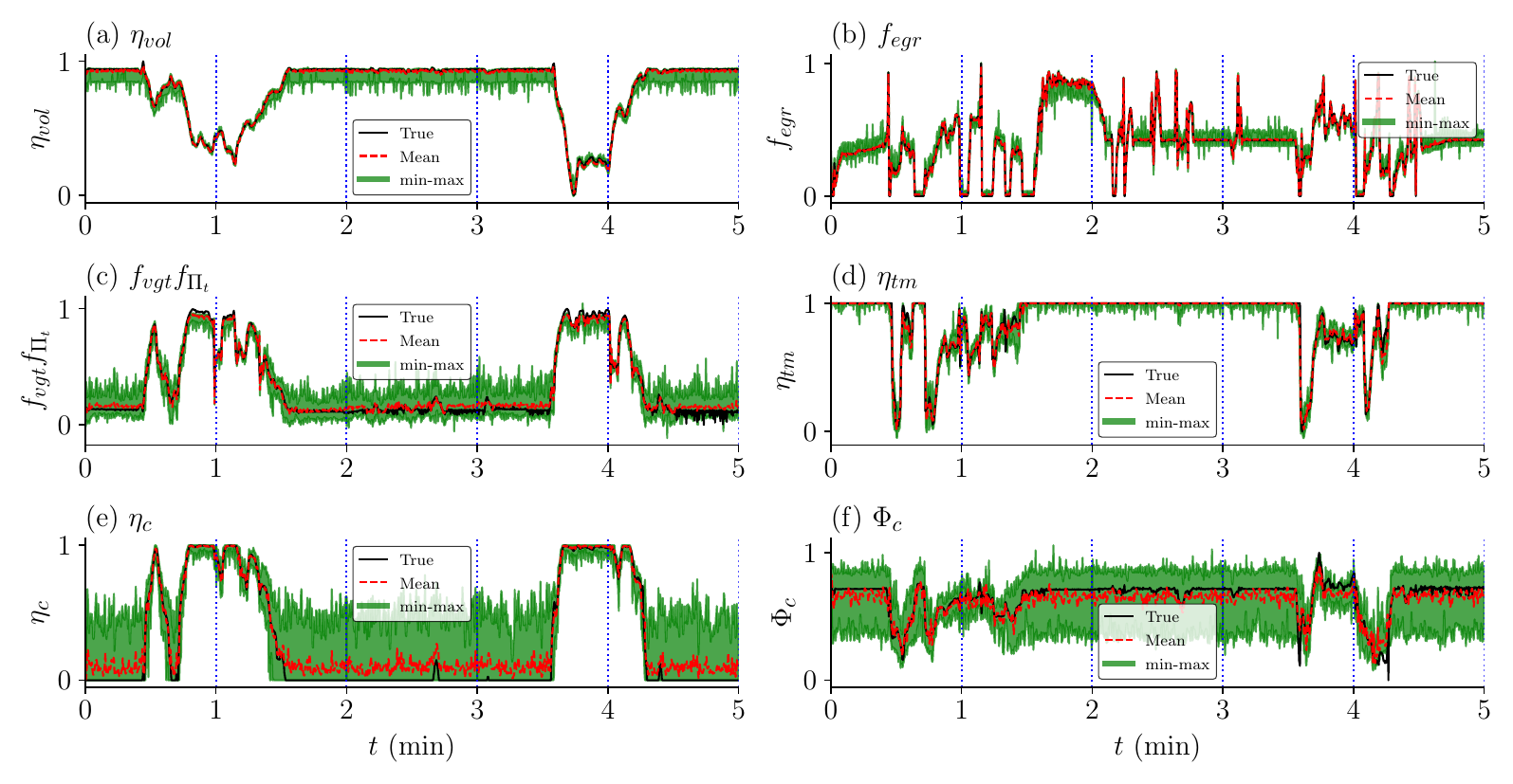}
    \caption{\textbf{Predicted empirical variables, Few-shot learning, Noisy data, 301 data points, ensemble:} Ensemble prediction for the empirical formulas for noisy data with 301 data points in the data loss predicted using few-shot transfer learning. The ensembles are calculated over 30 independent runs for each time segment.}
    \label{Figure: Empirical: Few shot Noisy data 301 points: ensemble}
\end{figure}

In this section, we provide additional results showing the predictions of different quantity for few-shot transfer learning method. In Table \ref{Table:Multihead_training}, we have shown the training of multi-head neural network for few-shot TL. As discussed in the main text, these networks are trained offline using data and Adam \cite{Kingma_2014adam} optimizer. One of the realizations of the state variable and $x_r$ and $T_1$ are shown in Fig. \ref{Figure: States: Few shot clean data 151 points} and intermediate variables are shown in Fig. \ref{Figure: Other variable: Few shot clean data 151 points} for the clean data case with 151 data points. As discussed in the main text, we also consider a case with 301 clean data. The violin plots of the predicted unknowns are shown in Fig. \ref{Figure:Few shot:Violin plot 301 clean data}. The ensemble plots for the different variables corresponding to 151 clean data case are shown in Figs. \ref{Figure: States: Few shot clean data 151 points: ensemble}, \ref{Figure: flow: Few shot clean data 151 points: ensemble} and \ref{Figure: empirical: Few shot clean data 151 points: ensemble}. The corresponding ensemble plots for different variables in the case of 301 noisy data points are shown in Figs. \ref{Figure: States: Few shot Noisy data 301 points: ensemble}, \ref{Figure: Flow: Few shot Noisy data 301 points: ensemble} and \ref{Figure: Empirical: Few shot Noisy data 301 points: ensemble}
\begin{table}[H]
\centering
\caption{\textbf{Training of multi-head networks for few-shot transfer learning (Phase-I):} Multi-head neural networks considered to approximate the state variables and $x_r$ and $T_1$.  The input to the network is time ($t$). The input time is 0 to 60 sec with $\delta t = 0.2$ sec, and the input is scaled between $[-1,1]$. Each head approximates a minute of the variable. We considered 240 heads for each output approximating 240 minutes of dynamics. Thus, network $\mathcal{N}_2(t, \bm{\theta}_2)$, $\mathcal{N}_3(t, \bm{\theta}_3)$ and $\mathcal{N}_4(t, \bm{\theta}_4)$ has 240 output, however network $\mathcal{N}_1(t, \bm{\theta}_1)$ has 480 output as it approximate two ($p_{em}$ and $p_{im}$) variables. The "Output transformation"  and scaling are considered as per Table \ref{Table:FNN for PINN}. The $L_2$ error shows \% relative $L_2$ error in training for each variable. }
\label{Table:Multihead_training}

\begin{tabular}{c|C{1cm}|C{3.15cm}}
\hline
Network & Output & Rel. $L_2$ error (\%) \\ \hline
 $\mathcal{N}_1(t, \bm{\theta}_1)$    & $p_{em}$, $p_{im}$ &  $p_{em}: 1.478\pm0.135$, $p_{im}: 1.476\pm0.078$\\ \hline
 $\mathcal{N}_2(t, \bm{\theta}_2)$    & $x_r$ &  $5.512\pm0.136$ \\ \hline
 $\mathcal{N}_3(t, \bm{\theta}_3)$    & $T_1$ &  $0.913\pm0.008$ \\ \hline
 $\mathcal{N}_4(t, \bm{\theta}_4)$    & $\omega_t$ & $1.035\pm0.028$ \\ \hline
\end{tabular}
\end{table}

\end{appendices}

\end{document}